\documentclass{article}

\usepackage[english]{babel}
\usepackage[acronym]{glossaries}
\newacronym{ge}{GE}{Grammatical Evolution}
\newacronym{sr}{SR}{Symbolic Regression}
\newacronym{dbs}{DBS}{Distance-based Selection}
\newacronym{alu}{ALU}{Arithmetic Logic Unit}
\newacronym{gp}{GP}{Genetic Programming}
\newacronym{bnf}{BNF}{Backus–Naur Form}
\newacronym{cfg}{CFG}{Context-Free Grammar}
\newacronym{ga}{GA}{Genetic Algorithm}
\newacronym{hdl}{HDL}{Hardware Description Language}
\newacronym{ea}{EA}{Evolutionary Algorithm}
\newacronym{ec}{EC}{Evolutionary Computation}
\newacronym{cpu}{CPU}{Central Processing Unit}
\newacronym{ic}{IC}{Integrated Circuit}
\newacronym{ahc}{AHC}{Agglomerative Hierarchical Clustering}
\newacronym{rmse}{RMSE}{Root Mean Square Error}
\newacronym{vhdl}{VHDL}{VHSIC Hardware Description Language}
\newacronym{mcu}{MCU}{Microcontroller Unit}
\newacronym{ml}{ML}{Machine Learning}
\newacronym{ssd}{SSD}{Solid-State Drive}
\newacronym{knn}{k-NN}{k-Nearest Neighbor}
\newacronym{cnn}{CNN}{Condensed Nearest Neighbor}
\newacronym{enn}{ENN}{Edited Nearest Neighbor}
\newacronym{bist}{BIST}{Built In Self Test}

\usepackage{algorithm}
\usepackage{algpseudocode}

\usepackage[T1]{fontenc}

\usepackage{booktabs}
\usepackage{multirow}
\usepackage{caption}
\usepackage{lscape}

\usepackage{subfigure}

\usepackage{fancyhdr}

\fancypagestyle{firstpage}{
    \fancyhf{} 
    \fancyfoot[L]{\thepage}
    \fancyfoot[R]{\makebox[\textwidth][r]{Preprint}}
}

\usepackage{authblk}
\usepackage{xcolor}

\usepackage[letterpaper,top=2cm,bottom=2cm,left=3cm,right=3cm,marginparwidth=1.75cm]{geometry}

\usepackage{amsmath}
\usepackage{graphicx}

\title{A Novel ML-driven Test Case Selection Approach for Enhancing the Performance of Grammatical Evolution}

\author[1]{Krishn Kumar Gupt\thanks{krishn.gupt@tus.ie}}
\author[2]{Meghana Kshirsagar}
\author[2]{Douglas Mota Dias}
\author[1]{\\Joseph P. Sullivan}
\author[2]{Conor Ryan}
\affil[1]{Department of Electrical \& Electronic Engineering, Technological University of the Shannon: Midlands Midwest, Limerick, V94 EC5T, Ireland}
\affil[2]{CSIS, University of Limerick, Limerick, V94 T9PX, Ireland}
\date{}

\begin{document}

\maketitle
\thispagestyle{firstpage}
\newenvironment{keywords}{{\vspace*{0.1in}\noindent\small\bf
Keywords:}\begin{quote}\small}{\par\end{quote}}
\begin{abstract}
Computational cost in metaheuristics such as \glspl{ea} is often a major concern, particularly with their ability to scale. In data-based training, traditional \glspl{ea} typically use a significant portion, if not all, of the dataset for model training and fitness evaluation in each generation. This makes \glspl{ea} suffer from high computational costs incurred during the fitness evaluation of the population, particularly when working with large datasets. To mitigate this issue, we propose a \gls{ml}-driven \gls{dbs} algorithm that reduces the fitness evaluation time by optimizing test cases. We test our algorithm by applying it to 24 benchmark problems from \gls{sr} and digital circuit domains and then using \gls{ge} to train models using the reduced dataset. We use \gls{ge} to test \gls{dbs} on \gls{sr} and produce a system flexible enough to test it on digital circuit problems further. The quality of the solutions is tested and compared against the conventional training method to measure the {\it coverage} of training data selected using \gls{dbs}, i.e., how well the subset matches the statistical properties of the entire dataset. Moreover, the effect of optimized training data on run time and the effective size of the evolved solutions is analyzed. Experimental and statistical evaluations of the results show our method empowered \gls{ge} to yield superior or comparable solutions to the baseline (using the full datasets) with smaller sizes and demonstrates computational efficiency in terms of speed.

\end{abstract}

\section{Introduction}

\glspl{ea} are well-known for their applications in different problem domains. One of the main features of \glspl{ea} such as \gls{ge} is using fitness evaluations to achieve a solution iteratively. This is the only step that guides how the chromosomes will change (using evolutionary operators) over time. As \cite{AdvancesInGP:kinnear:1994} explains, ``your fitness function is the only chance that you have to communicate your intentions to the powerful process that Genetic Programming represents.'' A fitness function, therefore, needs to do more than measure the performance of an individual; it needs to score individuals in such a way that they can be compared, and a more successful individual can be distinguished from a less one. Fitness evaluation time depends on several factors, such as the number of generations, population size, and test case volume (number of variables/features and instances). It is one of the most time-consuming steps \cite{fitnessTimeComsum:yang:2003}. Usually, the evolutionary parameters are user-based and can be tuned as per convenience and experience, but the test case is something we have less control over. A test case is a specific set of inputs, conditions, and expected outcomes used to assess the functionality and behavior of a design.

In the modern world, due to the ever-growing volume of test cases in problems like \gls{sr} and Digital Circuits, etc., the use of \gls{ge} becomes increasingly computationally costly and sometimes even impractical. Since these problem models learn and adapt by using data samples, their evolution is dependent on the training data\footnote{for ease of understanding, test cases used to train the model are referred to as training data. Similarly, test cases used to test the model are referred to as testing data} up to a great extent. At the same time, the greater the training data volume, the higher the evaluation time -- the computational time required to evaluate the model on data -- required. We preferred using \gls{ge} over other data-based learning algorithms as it combines the power of \gls{gp} with the flexibility of context-free grammar (details in Section \ref{sec:GE}), enabling it to handle a wide range of problems domains and evaluate the proposed algorithm.

One approach to speed up the evolutionary process is to reduce the amount of data that needs to be processed each generation. This is feasible because, in general,  not all instances of the training data are equally important. Usually, it is not known in advance to a test engineer which data samples will be most useful, particularly in the case of a `black-box' model. For example, in domains such as digital circuit design, where stringent testing is crucial but exhaustive testing is not viable, test engineers play a vital role in ensuring that products perform as required.

In real-world \gls{ml} processes, the past few years have witnessed increasingly large datasets due to the combination of improved data collection techniques and cheaper, faster storage such as \glspl{ssd}. 
As the world of Big Data has experienced, however, simply increasing the amount of training data doesn't necessarily improve the model-building process for ML. Too much data can lead to slower and more expensive training, and noise and redundancy within the dataset can negatively impact overall model accuracy \cite{CurseofBigData:duffy-deno}.  For instance, consider a dataset with millions of data instances, at least some of which are redundant. If one could successfully train an accurate model on a subset, then costs would be lower, perhaps significantly so, depending on the size of the subset. However, too small of a subset can lead to the missing of important patterns and subsequently produce less accurate models. This means the selected subset should statistically represent the appropriate distribution of the total test cases.

When using a large amount of training data in problems like \gls{sr} or digital circuits, \gls{ge} can require extensive resource consumption in fitness evaluation while evaluating test cases against each individual being trained. This makes the process more expensive for problems running for a large number of generations and/or population size or performing multiple runs. An effective test case selection, in such scenarios, can potentially contribute not only to improving the efficiency but also the accuracy of the model. However, while evolving \gls{sr} models using \gls{ge}, test case selection before learning is rarely considered, although several studies employed feature selection \cite{ali:automatedFeatureSel:2022}. There are some existing works in the case of digital circuits \cite{icsoft21,gupt:predive:2022}, but test case selection at the black-box level is still challenging and has much potential. Clustering is a standard technique to identify data instances based on their latent relationships, specifically to identify which instances are the most alike \cite{kshirsagar:igi:2022,Bindra2021InsightsRevolution}. We use it as a pre-processing step to obtain a useful subset of the test cases to test the model.

A key challenge with data selection is to identify how small the subset can be without compromising eventual model quality. We performed test case selection on six different fractions of the total data, as described in Section \ref{sec:methodology} and Table \ref{table:trainSizes}. The proposed algorithm is problem-agnostic, and we demonstrate its efficacy on  24 sets of benchmarks from the domains of SR (10 synthetic datasets and 10 real-world datasets) and digital circuits (4 problems). 

The algorithm offers an evolutionary advantage of competitive solutions to the examined baseline (where a larger dataset is used) regarding the test score and time efficiency. The contributions from our research are summarised below. Briefly, we 
\begin{itemize}
    \item propose \gls{dbs}, a problem-agnostic test case selection algorithm;
    \item use the \gls{dbs} algorithm to select training data to decrease the number of fitness evaluations during model training and, hence, reduces \gls{ge} run time;
   \item demonstrate how \gls{dbs} performs at least as well as -- and sometimes significantly better -- in terms of achieving better test scores than the baseline considered;
    \item show that the solutions produced using the training data selected by \gls{dbs} sometimes have interesting results on effective solution size, a phenomenon that merits further investigation;
    \item demonstrate our results by validating \gls{dbs} algorithm on 24 problems, laying the foundation for future studies that investigate this approach in other problem domains and more complex problems.
\end{itemize}

This article is organized as follows. Section \ref{sec:Literature} describes the current research on test case selection in the domain of \gls{sr} and digital circuits. In Section \ref{sec:GE}, we present the evolutionary approach of \gls{ge} used in this paper along with the basic foundation. In Section \ref{sec:DBS}, a detailed description of the proposed \gls{dbs} algorithm is presented. The benchmark problems tested and the methodology used in the research are discussed in Section \ref{sec:ExpSetup}. We discuss our findings and compare them to the baseline results along with statistical validations in Section \ref{sec:results} and conclude the contributions in Section \ref{sec:conclusions}.

\section{Background}
\label{sec:Literature}
This section reviews related studies on test case selection along with a brief background of SR and digital circuits. We roughly divide the related studies into two categories: test case selection for \gls{sr} and digital circuits.

\subsection{Symbolic Regression}
\gls{sr} is a method that creates models in the form of analytic equations that can be constructed by using even a small amount of training data, as long as it is representative of the eventual training data \cite{kubalik2020symbolic}. Sometimes, information or certain characteristics of the modeled system are known in advance, so the training set can be reduced using this knowledge \cite{instanceSel:arnaiz:2016}. Instance selection, as known in \gls{sr}, sometimes referred to as {\bf test case selection} in digital circuits, in such cases can significantly impact the efficiency and accuracy of the model's learning process by strategically choosing representative data points for evaluation.

In recent years, different studies have used test case selection in the field of \gls{sr} and digital circuits. \cite{instanceSelectionRegression:kordos:2012} introduced instance selection for regression problems, leveraging solutions from \gls{cnn} and \gls{enn} typically used in classification tasks. Another work by the same researchers used the \gls{knn} algorithm for data selection in \gls{sr} problems and later evaluated them using multi-objective \gls{ea}  \cite{instanceSelectionRegression:kordos:2018}. 

\cite{instanceSelectionRegression:Son:2006} proposed an algorithm where the dataset was divided into several partitions. An entropy value is calculated for each attribute in each partition, and the attribute with the lowest entropy is used to segment the dataset. Next, Euclidean distance is used to locate the representative instances that represent the characteristics of each partition. \cite{trainingSetSelection:kajdanowicz:2011} used clustering to group the dataset and distance based on the entropy variance for comparing and selecting training datasets. Their instance selection method computes distances between training and testing data, creates a ranking based on these distances, and selects a few closest datasets. This raises concerns about biased training, as it predominantly utilizes data that closely resemble the testing data for training purposes.

Numerous instance selection algorithms have been developed so far. A cluster-based instance selection algorithm utilizing a population-based learning approach is introduced in \cite{clusterSelection:czarnowski:2010,clusterSelection:czarnowski:2012}. This method aims to enhance the efficiency of instance selection by grouping similar data instances into clusters and then employing a population-based learning algorithm to identify representative instances. While effective for classification problems, it's worth noting that the population-based learning algorithm can be computationally demanding, especially with large datasets, due to its inherent complexity and extensive computations involved in identifying suitable representatives. The instance reduction in \cite{ClusterApproach:czarnowski:2003} relies on a clustering algorithm, which, however, does not take advantage of additional information, such as data diversity within the cluster. As a result, it might inadvertently pick similar cases from within a cluster. The instance removal algorithm in \cite{ClusterReduction:wilson:2000} is based on the relationships between neighboring instances. Nonetheless, the approach may benefit from enhanced user control over the resulting subset's size. Further, the method struggles with noisy instances, acting as border points, that can disrupt the removal order.

While various techniques have been proposed for data selection or reduction in \gls{ml} tasks, most studies have concentrated on classification tasks \cite{instanceSelectionRegression:Son:2006,clusterSelection:czarnowski:2012,clusterSelection:czarnowski:2010,clusterSelection:czarnowski:2012,ClusterApproach:czarnowski:2003,ClusterReduction:wilson:2000}. In contrast, little attention has been paid to the challenges related to instance selection in \gls{sr} \cite{featureSR:chen:2017,instanceSelectionRegression:arnaiz:2016}.

\subsection{Digital Circuits}
Digital circuits are of utmost importance in our daily lives, powering everything from alarm clocks to smartphones. A clear and concise way to represent the features of a digital design is by using a high-level language or \gls{hdl}. \glspl{hdl} are specialized and high-level programming languages used to describe the behavior and structure of digital circuits. Designing a digital circuit begins with designing its specifications in HDLs and ends with manufacturing, with several testing processes involved throughout the different design phases. There are many \glspl{hdl} available, but Verilog and \gls{vhdl} are the most well-known \glspl{hdl} and are frequently used in digital design \cite{book:IntrotoDigSys}. In this study, we use \gls{vhdl} as the design language for the circuits we examine, although Verilog would have been a perfectly acceptable alternative.

Testing is hugely important when designing digital circuits, as errors during fabrication can be prohibitively expensive to fix. This has resulted in the creation of extremely powerful and accurate circuit simulators \cite{tan2014verilog}. Numerous \gls{hdl} simulators are available, such as ModelSim, Vivado, Icarus, and GHDL, to name a few. During simulation, test scenarios are carefully chosen to reflect the circuits' desired behaviors in actual applications. A typical simulation session has the circuit model encapsulated in a testbench that consists of a response analyzer and other important components \cite{gupt2021PrimeField}. Testbenches are essentially \gls{hdl} programs containing test cases, which are intended to assure good coverage through testing \cite{gupt2021GF}. Typically, hardware simulation is much slower than software simulation \cite{hsiung2018reconfigurable}, which is one more reason to adopt test case selection and reduce the design and testing time. With this in mind, it is reasonable to expect that the benefit experienced in the circuit design experiments will be more noticeable than in \gls{sr} experiments.

Many studies have used automatic and semi-automated test case selection methods. These methods' main objective is to provide high-quality tests, ideally through automated techniques, that provide good coverage while reducing the overall number of tests \cite{tcg:YeildRampChallanges}. The fundamental testing method, random testing, often involves randomly choosing test cases from the test case space (the set of all potential inputs) \cite{orso2014software}. Although it is cost-effective and easy to implement, it may result in several issues, such as selecting redundant test cases, missing failure causing inputs, or lower coverage \cite{mrozek2012antirandom}.
\cite{PseudoExhaustive:Kuhn:2006} proposed pseudo-exhaustive testing that involves exhaustively testing pieces of circuits instead of the complete circuits.
Based on the number of defects the test cases can identify, \cite{thamarai2010heuristic} suggested reducing the number of test cases in their work. They suggested heuristic techniques for determining the most crucial test cases for circuits with smaller test sets. \cite{muselli2000training} suggested clustering as a method for classifying problems involving binary inputs. In their research, they compared input patterns and their proximity to one another using Hamming distance to create clusters.

\cite{testing:faultDefType:timeCal:Bushnell2002EssentialsCircuits} provide insights on algorithms such as Automatic Test-Pattern Generation (ATPG) for generating test cases for structural testing of digital circuits, which can find redundant or unnecessary circuit logic and are useful to compare whether one circuit's implementation matches another's implementation. They highlight the impracticability of performing exhaustive testing of a 64-bit ripple-carry adder with 129 inputs and 65 outputs, suggesting that such testing can be suitable for only small circuits. In contrast, modern-day circuits tend to have an exponential volume of test cases. 

\cite{mrozek2012antirandom} proposed \textit{anti-random testing} for \gls{bist} and used Hamming distance as a measure to differentiate the input test cases from previously applied ones. In an extended work, \cite{OptimalRandomTest:Mrozek:2017} proposed an optimized controlled random test where they used information from the previous tests to maintain as much distance as possible among input test cases. However, the quality of the generated test cases depends on the initial random test case; if it does not cover a diverse set of conditions, the subsequent test cases may not effectively explore the full range of possible inputs.

Although some studies have proposed test case selection in \gls{sr} or digital circuits, they have not been explored and evaluated with metaheuristics like \gls{ge}. Moreover, the existing methods are problem-specific, i.e., their test case selection algorithms apply to either \gls{sr} or digital circuits. Our proposed \gls{dbs} algorithm is problem agnostic and, with a minor tweak (based on the type of data), can be easily applied to both problem domains. We decided to evaluate the proposed test case selection method using \gls{ge} where \gls{dbs} is first used to select a subset of test cases and use them as training data for the model. This choice has the advantage of obtaining solutions with low prediction error (typically expressed as \gls{rmse}, for \gls{sr} problems) or a high number of successful test cases (expressed at \textit{hit-count} in the case of digital circuit design) using substantially reduced data. The obtained solutions using \gls{ge} are tested against the testing data to calculate their accuracy or, alternatively, the quality of training data and strength of \gls{dbs} algorithm.

\section{Grammatical Evolution}
\label{sec:GE}

\gls{ge}, initially proposed by \cite{ge:ryan1998}, is a grammar-based \gls{gp} approach used to evolve programs in arbitrary languages. Each individual in the population is represented by a list of integers called genotype, which is generated from initially random values from the interval [0, 255]. The list element utilizes a \textit{mapping} process to map a genotype to a corresponding phenotype, which is a potential solution to a problem. All evolutionary operators are performed on the genotype. 

The mapping is achieved by employing a grammar, typically described in \gls{bnf}. This is a \gls{cfg} often used to specify the syntax rules of different languages used in computing \cite{handbookofGE}. The grammar is a tuple {\em G = (NT, T, S, P)} where {\em NT} represents a non-empty set of \textit{non-terminals} {\em (NT)} that expand into non-terminals and/or \textit{terminals} ({\em T}), governed by a set of \textit{production} {\em (P)} rules. Terminal and non-terminal symbols are the lexical elements used in specifying the production rules constituting a formal grammar. {\em S} is an element of {\em NT} in which the derivation sequences start, called the axiom. The rule in {\em P} is represented as $A ::= \alpha$, where $A \in NT$ and $\alpha \in (NT \sqcup T)$. Figure 1(a) shows an example of \gls{cfg}.

\gls{ge} typically employs an integer-coded genotype. A \gls{ga} performs the evolutionary process by initially creating a population of integer-coded genomes and then,  at each iteration (generation), they are evaluated, and various genetic operations are conducted. Evaluation involves mapping them into their respective phenotypes using the \gls{ge} mapper. The mapping is done starting from the axiom {\em (S)} of the grammar and expanding the leftmost {\em NT}. The codons are used to choose which rule in {\em P} to expand by applying the modulo (mod) operator between the codon and the number of derivation rules in the respective {\em NT}. 

Figure 1(b) gives an example of mapping using the example of grammar in Figure 1(a). The mapping process starts with: (a) the axiom of the grammar \texttt{<\textcolor{blue}{start}>}, which expands to three alternatives, and (b) the unused codon of genotype 33. Next, a mod operator between codon (33) and the number of production rules (3) provides an index of rules to be expanded. The rule to be expanded at index 0 (\(33\ mod\ 3 = 0\)) is \texttt{<\textcolor{blue}{expr}><\textcolor{blue}{op}><\textcolor{blue}{expr}>}. This mapping procedure is repeated until all the {\em NTs} are expanded, or there are no more codons left. If the genotype is exhausted, but some {\em NTs} still remain, a wrapping technique can be employed where codons are reused. The wrapping process stops if a valid individual is achieved or after a certain number of wraps have been performed. In the latter case, the individual is considered invalid if any {\em NTs} remain unmapped.

\begin{figure}[!h]
    \centering
    \includegraphics[width=.9\textwidth]{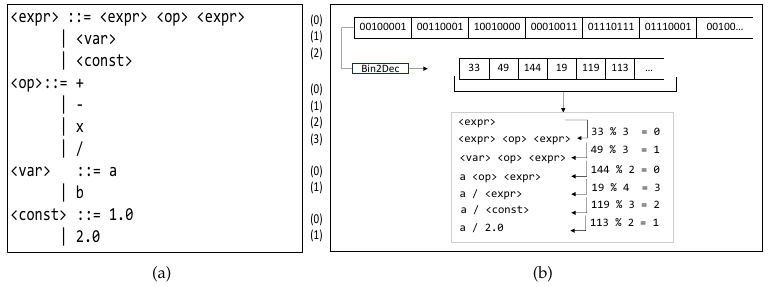}
    \caption{
    \textbf{(a)} Example of a \gls{cfg}. \textbf{(b)} Mapping of the integer-coded genome to a corresponding computer program using a grammar}
    \label{fig:GEBNFmapping}
\end{figure}

As with GP, GE evaluates each mapped program and assigns them a fitness score. Here, fitness is a measurement of the ability of an individual to attain the required objective. Candidate solutions are evaluated by comparing the output against the target function and taking the sum of the absolute errors over a certain number of trials. Programs that contain a single term or those that return an invalid or infinite result are penalized with an error value high enough that they have no opportunity to contribute to the next generation. Genetic selection, crossover, mutation, and replacement are probabilistically applied to the genotypic strings to produce the following generation. This continues for a specified number of generations, budgets, or until the sought solution is achieved. The search space in \gls{ge} comprises all the possible computer programs that can be produced using a particular grammar. In recent years, \gls{ge} has been widely used in many domains such as cybersecurity \cite{handbookofGE}, software testing \cite{anjum2021seeding}, \gls{sr} problems \cite{murphy2021time,youssef2021evolutionary}, digital circuit design \cite{gupt:predive:2022} to name a few.

\section{Proposed Approach}
\label{sec:DBS}
DBS has two objectives when creating a training set: maintain high coverage while minimizing the amount of training data. The selected test cases are used as training data to evolve the respective model using \gls{ge}. Since the test cases are spread over a vast space, it is important to categorize them into groups based on certain relationships. For this, we take advantage of clustering algorithms before test case selection is performed.

Based on the nature of data (integer and floating-point number) in \gls{sr}, we use K-Means clustering with Euclidean distance ($d_E$) as  
\begin{equation}
\label{eq:euclidean}
    d_E\left( p,q\right)   = \sqrt {\sum _{i=1}^{n}  \left( q_{i}-p_{i}\right)^2 }, 
\end{equation}
where $p$ and $q$ are two test case points in Euclidean n-space.
K-Means implicitly relies on pairwise Euclidean distances between data points to determine the proximity between observations. A major drawback in the K-Means algorithm is that if a centroid is initially introduced as being `far away', it may very likely end up having no data points associated with it. This is a known weakness with the K-Means algorithm, which can be mitigated by using K-Means++ as the initialization method. This ensures a more intelligent introduction of the centroids and enhances the clustering process. The {\em elbow method} (a heuristic to determine the number of clusters in a dataset) is used along with the \texttt{KneeLocator} function to get an optimal number of clusters. This completely automates the process and produces the optimal number of clusters.

When preparing data for an ML exercise, \gls{dbs} is applied before building models. We describe the process for the digital circuit domain, although it is broadly the same for \gls{sr}.

Since the test cases in digital circuit domains are binary (0,1) data and K-Means computes the mean, the standard `mean' operation is useless. Also, after the initial centers are chosen (which depends on the order of the cases), the centers are still binary data. There are chances of getting more frequent ties in the distance calculation, and the test cases will be assigned to a cluster arbitrarily \cite{ibmClusteringBinary}. Hence, to group the data, we use \gls{ahc}, a `bottom-up' approach that is the most common type of hierarchical clustering used to group objects in clusters based on their similarity \cite{AgHierarchClustring:Contreras:2015}. As Hamming distance $d_H$ is more appropriate for binary data strings, we use it as a distance measure with \gls{ahc}. To calculate the Hamming distance between two test cases, we perform their XOR operation, $(p \oplus\ q)$, where $p$ and $q$ are two test case (binary) points of n-bits each, and then count the total number of 1s in the resultant string as
\begin{equation}\label{eq:Hamming}
d_H (p, q)= \sum_{i=0}^{n} |p_i - q_i|.
\end{equation}

We calculate the distance between clusters using complete-linkage, where the similarity of two clusters is the similarity of their most dissimilar members. This has been assessed \cite{tamasauskas2012evaluation} as a highly effective hierarchical clustering technique for binary data. In practice, this is comparable to picking the cluster pair whose merge has the smallest diameter. This merge criterion is non-local, and decisions regarding merges can be influenced by the overall clustering structure. This causes sensitivity to outliers, which is important for test case selection and a preference for compact clusters \cite{hierClsu:linkage:manning:2008}.

\begin{figure}[!h]
    \centering
    \includegraphics[width=.8\textwidth]{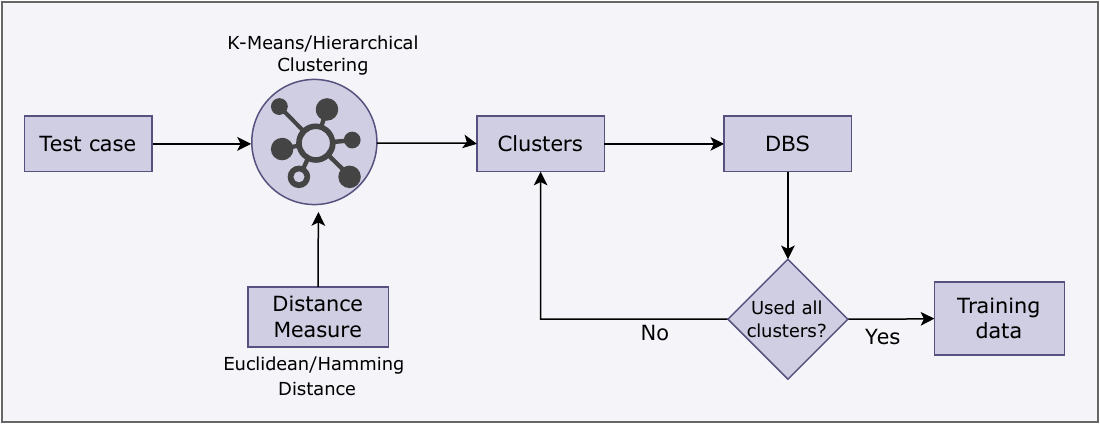}
    \caption{The selection process involving \gls{ml} and DBS algorithm}
    \label{fig:TestSelectionBlock}
\end{figure}

Once we have created the clusters, the test case selection is performed using DBS. The complete process is shown in Figure \ref{fig:TestSelectionBlock}. Note that clustering is a pre-processing step for \gls{dbs}. The DBS selection approach, given in Algorithm \ref{alg:dbs}, employs a greedy strategy to pick the test case pairs most dissimilar to all others in the given cluster. As a result, test cases are chosen to maintain diversity (to represent a large test case space) and maximize coverage. A diversity analysis of the data selected using the DBS algorithm is presented in \cite{icsoft21}. 

During test case selection, a distance matrix is first created by executing a proximity calculation on each test case of a given cluster. The choice of distance measurements is crucial, as was mentioned during the clustering procedure. We use Euclidean distance for \gls{sr} data and Hamming distance for digital circuit data because they are better appropriate for their respective datasets. The test case selection is performed based on the maximum distance between input test case pairs. It is worth mentioning that the entire clustering and test case selection process is applied to the input test cases. It is possible that two or more distinct input test cases might pose the same distance from a given input test case, and hence, duplicate data selections are avoided. The size of the test case subset to be chosen or the stopping criterion relies on the user-defined budget or $B$.

\begin{algorithm}[t]
\caption{\acrfull{dbs}}\label{alg:dbs}
\begin{algorithmic}[1]
\Require{$C$ clusters of test cases}
\Require{Selection budget $B\%$}
\Ensure{Training data}
\State $trainingData \gets []$ \Comment{Initialize an empty list}
\For{each cluster $C_i$ in $C$}
    \State $inputTestCases \gets$ Extract input values from test cases in $C_i$
    \State $N \gets$ Count of test cases in $C_i$

    \State Initialize a distance matrix $D$ with dimensions $N \times N$
    
    \For{each pair of $inputTestCases$ $(T_j, T_k)$ in $C_i$}
        \State $D \gets$ distance between $T_j$ and $T_k$ \Comment{use $d_E$ or $d_H$}
        \State Store the distance in $D[j][k]$
    \EndFor
    
    \State Remove the upper half of the distance matrix \Comment{$D[j][k]$ where $k \geq j$}

    \State $sortedDistances \gets$ Sort distances in $D$ in decreasing order

    \State $cnt \gets \frac{B \cdot N}{100}$ \Comment{Calculate the number of cases to select as $B\%$ of $N$}

    \State $desireIndex \gets$ Indices of the top $cnt$ test cases corresponding to $sortedDistances$
    
    \For{$index$ \textbf{in} $desireIndex$}
        \State $trainingData \gets$ Append the corresponding test case from $C_i$
    \EndFor
\EndFor

\State \textbf{Return} $trainingData$ (all selected test cases from all clusters)
\end{algorithmic}
\end{algorithm}

Once the required test suite has been obtained, we use it as the training set. The approach is effective and the same for SR and digital circuits but necessitates choosing the distance measure based on the data type. %
Sorting the distances is essential while choosing the test cases. This facilitates a simple test case selection based on the ascending/descending order of test case diversity. In our experimental setup, we select test cases in descending order of diversity. The value of the count may need to be rounded off depending on how many test cases are in a cluster. We round it up using the \texttt{ceil} function.

\section{Experimental Setup}
\label{sec:ExpSetup}
In this section, the benchmarks selected to evaluate the DBS method are discussed. Moreover, the methodology used for SR and digital circuit problems is discussed in detail, along with the GE setup and parameters.

\subsection{Benchmarks}
\label{sec:benchmakrs}

We selected benchmark suites with a high level of variance in terms of problem complexity and test case volume. These benchmarks are popular in their respective problem domains, i.e., \gls{sr} and digital circuit. 

\subsubsection{SR Benchmarks}
The SR benchmarks are selected from both synthetic and real-world problems. Recently, a subset of these benchmarks was also used to test different tools and algorithms, based on \gls{ge}, such as GELAB \cite{gelab:krishn:2022}, AutoGE \cite{ali:2021:autoge}, and GRAPE \cite{Allan:grape:2022}.

Table \ref{tab:bench_Synthetic} presents 10 benchmarks with different ranges of variables selected from \cite{mcdermott:2012:gpNeedBench}. The benchmarks are diverse, having 1 to 5 input features and sample sizes varying from 402 to 20,000. More details, such as mathematical expressions and training and test range, can be found in \cite{mcdermott:2012:gpNeedBench}. 
\begin{table}[!h]
    \centering
    \caption{Synthetic symbolic regression benchmark candidates. Benchmarks are in order by \# features.}
    \label{tab:bench_Synthetic}
    \begin{tabular}{l c r}
    \cmidrule(lr){1-3}
        \textbf{Dataset}  & \# \textbf{ Features}  & \# \textbf{ Instances}\\
\cmidrule(lr){1-3}
Keijzer-4 & 1 & 402 \\
Keijzer-9 & 1 & 1102 \\
Keijzer-10 & 2 & 10301 \\
Keijzer-14 & 2 & 3741 \\
Nguyen-9 & 2 & 1020 \\
Nguyen-10 & 2 & 1020 \\
Keijzer-5 & 3 & 11000 \\
Vladislavleva-5 & 3 & 3000 \\
Korns-11 & 5 & 20000 \\
Korns-12 & 5 & 20000 \\
\cmidrule(lr){1-3}
\end{tabular}
\end{table}

We also selected 10 real-world \gls{sr} problems that have been widely used in many research \cite{Oliveira:SRreal-word:2018,sarmaad:SRreal-world:ecta21}. These are listed in Table \ref{tab:bench_realWorld}. The dataset for Dow Chemical was obtained from the GP Benchmarks\footnote{http://gpbenchmarks.org/?page\_id=30}. The other datasets were obtained from the UCI Machine Learning Repository\footnote{https://archive.ics.uci.edu/ml/datasets.php} and StatLib-Datasets Archive\footnote{http://lib.stat.cmu.edu/datasets/}. The benchmarks are diverse in terms of problem difficulty, having 5 to 124 input features with varying numbers of instances from 506 to nearly 4,898. Each dataset is referred to with a short name for ease and will be used in the rest of the paper.

\begin{table}[!h]
    \centering
    \caption{Real-world symbolic regression benchmark candidates. Benchmarks are in order by \# features.}
    \label{tab:bench_realWorld}
    \begin{tabular}{l c c r}
    \cmidrule(lr){1-4}
        \textbf{Dataset}  & \textbf{Short name } & \textbf{\# Features}  & \textbf{\# Instances}\\
\cmidrule(lr){1-4}
Airfoil Self-Noise & \texttt{airfoil} & 5 & 1503 \\
Energy Efficiency - Heating & \texttt{heating} & 8 & 768 \\
Energy Efficiency - Cooling & \texttt{cooling} & 8 & 768 \\
Concrete Strength & \texttt{concrete} & 8 & 1030 \\
Wine Quality - Red Wine & \texttt{redwine} & 11 & 1599 \\
Wine Quality - White Wine & \texttt{whitewine} & 11 & 4898 \\
Boston Housing & \texttt{housing} & 13 & 506 \\
Pollution & \texttt{pollution} & 15 & 60 \\
Dow Chemical & \texttt{dowchem} & 57 & 1066 \\
Communities and Crime & \texttt{crime} & 124 & 1994 \\
\cmidrule(lr){1-4}
    \end{tabular}
\end{table}

\subsubsection{Digital Circuit Benchmarks}
 A brief description of the selected circuit problems is described below and is listed in Table \ref{tab:bench_circuit}.
\begin{table}[!h]
\centering
 \caption{Digital circuit benchmark candidates. Circuits are in alphabetical order.}
 \label{tab:bench_circuit}
 \begin{tabular}{l c c r}
 \cmidrule(lr){1-4}
 \textbf{Circuit} & \textbf{\# Input}  & \textbf{\# Output} & \textbf{\# Test cases} \\ 
 \cmidrule(lr){1-4}
5-bit Comparator & 10 & 3 & 1,024 \\
5-bit Parity & 5 & 1 & 32 \\
11-bit Multiplexer & 11 & 1 & 2,048 \\
ALU & 12 & 5 & 4,096 \\ 
\cmidrule(lr){1-4}
 \end{tabular}
\end{table}

A binary magnitude comparator is a combinational logic circuit that takes two numbers and determines if one is greater than, equal to, or less than the other. These circuits are often used in \glspl{cpu}, \glspl{mcu}, servo-motor control, etc. In this paper, a 5-bit comparator is used. A parity generator is a combinational logic circuit that generates the parity bit in a transmitter. A data bit's and a parity bit's sum can be even or odd. In the case of even parity, the additional parity bit will make the total number of 1s even. In contrast, in the case of odd parity, the additional parity bit will make the total number of 1s odd. A multiplexer, or MUX, is a logic circuit that acts as a switcher for a maximum of $2^s$ inputs and $s$ selection lines and synthesizes to a single common output line in a recoverable manner for each input signal. They are used in communication systems to increase transmission efficiency. In this paper, an 11-bit multiplexer is used. An \gls{alu} is a combinational circuit that performs arithmetic and logic operations according to the control bits. It is a fundamental building block of the \gls{cpu}. In this paper, we have used a 5-bit operand size and 2-bit control signal, and thus, it can target four operations (arithmetic \& logic): addition, subtraction, AND, and OR.

\subsection{Methodology}
\label{sec:methodology}
The experimental setup for both problem domains is relatively similar, except for their implementation of the fitness evaluator, and hence, both problems incorporate two separate fitness functions.

We performed 5,040 experiments in total, corresponding to 30 independent runs for each of the 24 problems using each training dataset (described next). Based on a small set of initial runs, the evolutionary parameters used for all the experiments are given in Table \ref{tab:gePara}.
\begin{table}[!h]
\centering
 \caption{Evolutionary parameters used in experimentation}
 \label{tab:gePara}
 \begin{tabular}{l r}\cmidrule(lr){1-2}
 \textbf{Parameter  } & \textbf{Value}\\  
 \cmidrule(lr){1-2}
 \# Runs & 30 \\ 
 Total Generations & 50  \\
 Population Size & 250 \\
 Selection Type & Tournament \\
 Crossover Type & Effective  \\ 
 Crossover Rate & 0.9  \\  
 Mutation Rate & 0.01  \\  
 Initialisation Type & Sensible  \\  \cmidrule(lr){1-2}
 \end{tabular}
\end{table}

We evaluate the \gls{dbs} algorithm for both domains at a range of test case selection budgets $B$. Using the baseline training data (see Table \ref{table:trainSizes} for details on this, created for each domain), we apply the \gls{dbs} algorithm for test case selection using different budgets from 70\% to 45\% at intervals of 5\%. The budget starts at 70\% of the baseline training dataset as we tried to have a significant reduction. We considered six different budgets to select training data using \gls{dbs}, i.e., 70\%, 65\%, 60\%, 55\%, 50\%, 45\%.

\begin{landscape}
\begin{table}[H]
\centering
 \caption{Training data and Testing data size used in different experiments of 24 benchmarks.}
 \label{table:trainSizes}
\begin{tabular}{c|lcccccccr|}
 \cmidrule(lr){2-10} &
\multirow{2}{*}{\textbf{Benchmarks}} & \multicolumn{7}{c}{\textbf{Training data size}} & \multirow{2}{*}{\textbf{Testing data size}} \\ \cmidrule(lr){3-9}
                           & & \textbf{Baseline} & \textbf{DBS (70\%)} & \textbf{DBS (65\%)} & \textbf{DBS (60\%)} & \textbf{DBS (55\%)} & \textbf{DBS (50\%)} & \textbf{DBS (45\%)} & \multicolumn{1}{c|}{} \\
\cmidrule(lr){2-10}
\multirow{10}{*}{\rotatebox[origin=c]{90}{\textbf{Synthetic SR}}}
    & Keijzer-4       & 282   & 199  & 185  & 170  & 157  & 141  & 128  & 120  \\
    & Keijzer-9       & 772   & 540  & 503  & 464  & 426  & 386  & 349  & 330  \\
    & Keijzer-10      & 7211  & 5049 & 4689 & 4328 & 3968 & 3606 & 3247 & 3090 \\
    & Keijzer-14      & 2619  & 1835 & 1704 & 1572 & 1443 & 1310 & 1180 & 1122 \\
    & Nguyen-9        & 714   & 501  & 466  & 430  & 395  & 357  & 322  & 306  \\
    & Nguyen-10       & 714   & 501  & 465  & 428  & 394  & 358  & 322  & 306  \\
    & Keijzer-5       & 7700  & 5391 & 5007 & 4621 & 4237 & 3851 & 3466 & 3300 \\
    & Vladislavleva-5 & 2100  & 1471 & 1366 & 1261 & 1157 & 1051 & 946  & 900  \\
    & Korns-11        & 14000 & 9802 & 9103 & 8401 & 7703 & 7002 & 6304 & 6000 \\
    & Korns-12        & 14000 & 9803 & 9104 & 8401 & 7703 & 7001 & 6304 & 6000 \\
\cmidrule(lr){2-10}
\multirow{10}{*}{\rotatebox[origin=c]{90}{\textbf{Real-world SR}}}
    & airfoil   & 1053 & 739  & 687  & 634  & 581  & 527  & 476  & 450  \\
    & heating   & 538  & 379  & 352  & 326  & 298  & 269  & 245  & 230  \\
    & cooling   & 538  & 379  & 352  & 326  & 298  & 269  & 245  & 230  \\
    & concrete  & 721  & 506  & 472  & 435  & 399  & 362  & 328  & 309  \\
    & redwine   & 1120 & 786  & 730  & 674  & 619  & 561  & 505  & 479  \\
    & whitewine & 3429 & 2402 & 2230 & 2058 & 1887 & 1715 & 1544 & 1469 \\
    & housing   & 355  & 250  & 233  & 215  & 198  & 178  & 161  & 151  \\
    & pollution & 42   & 32   & 30   & 26   & 25   & 22   & 21   & 18   \\
    & dowchem   & 747  & 524  & 487  & 450  & 412  & 375  & 338  & 319  \\
    & crime     & 1396 & 979  & 912  & 842  & 771  & 700  & 632  & 598  \\
\cmidrule(lr){2-10}
\multirow{5}{*}{\rotatebox[origin=c]{90}{\textbf{Circuit}}}
    & 5-bit Comparator & 1024  & 720   & 672   & 616  & 568  & 512  & 461  & 1024  \\
    & 5-bit parity & 32	& 24 & 22 & 20 & 18 & 16 & 16 & 32\\ 
    & 11-bit Multiplexer & 2048  & 1440  & 1344  & 1232 & 1136 & 1024 & 928  & 2048  \\
    & ALU & 4096  & 2877  & 2669  & 2461 & 2253 & 2045 & 1856 & 4096  \\
     
\cmidrule(lr){2-10}
\end{tabular}
\end{table}
\end{landscape}

\subsubsection{SR Methodology}
\label{sec:SRmethodology}
In the case of experiments involving \gls{sr} datasets, we used the conventional train-test split approach; in our case, 70\%-30\% train-test split on the datasets. We use \gls{rmse} as a fitness function in this experiment. This is one of the most commonly used measures for evaluating the quality of predictions and is represented in Equation \ref{eq:rmse}.
\begin{equation}
    \mathit{RMSE}  = \sqrt \frac{{\sum _{i=1}^{\eta_o}  \left( P_{i}-O_{i}\right)^2 }}{\eta_o}
    \label{eq:rmse}
\end{equation}
Here, $P_i$ and $O_i$ are the predicted and observed (actual) values for the $i^{th}$ observation, respectively, while $\eta_o$ is the total number of observations. Across all experiments, we allowed the model to train for 50 generations. 

The best-performing individuals using 70\% training data from each generation are tested against the 30\% testing data, and their \gls{rmse} scores are recorded. We consider this as a baseline in this experiment. 

A list of all the problems and their respective training and testing data sizes are given in Table \ref{table:trainSizes}.
These new training datasets selected using the \gls{dbs} method are used to train the \gls{sr} models. The best-performing individuals from each generation are tested against the testing data, and their \gls{rmse} scores are recorded. All experiments on a particular benchmark use the same testing data.

For ease of readability, 70\% of the total data used to train is referred to as {\em baseline} in Table \ref{table:trainSizes} and the remaining 30\% as {\em testing data}. The data selected from baseline training data using \gls{dbs} at budget $B$ is represented as \gls{dbs} (B\%) training data.

\subsubsection{Digital Circuit Methodology}
\label{sec:DCmethodology}
In the case of digital circuits, we use a slightly different approach. Unlike calculating predictions or \gls{rmse} as fitness in the case of \gls{sr}, here we calculate the number of successful events or test cases (where expected output and observed output are the same) of the evolved circuit. The objective is to achieve a design that should pass all (or maximum) instances in training data. We call this fitness function `hit-count,' as it counts the number of successful hits as 
\begin{equation}
    \mathit{hit\textnormal{-}count} =  {{\sum _{i=1}^{\eta_o}  (P_i=O_i \rightarrow 1)}},
    \label{eq:hitcount}
\end{equation}
where $P_i$ and $O_i$ are the predicted and observed outputs for $i^{th}$ observation, while $\eta_o$ is the total number of observations or instances in training data. Considering the above-mentioned objective and problem specification, the total data available for a benchmark is used as training data and exhaustive testing is performed using the same dataset. The best test scores are considered as baseline. 

\begin{figure}[!h]
    \centering
    \includegraphics[width=.8\textwidth]{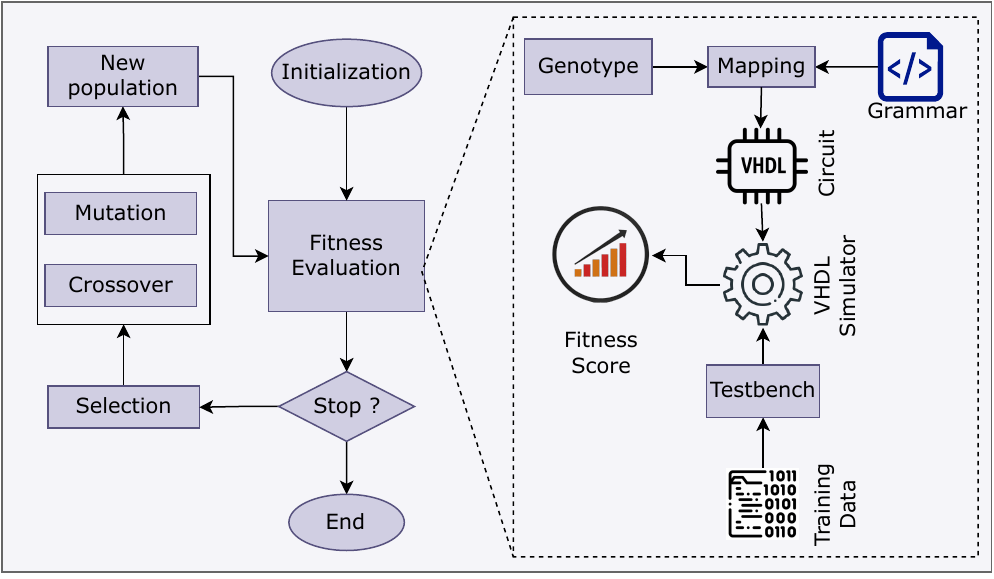}
    \caption{Circuit training process}
    \label{fig:SetupCircuit}
\end{figure}

The \gls{dbs} approach uses the same strategy of selecting a subset of baseline training data as is used in \gls{sr}. The testing data is kept the same for all experiments on a benchmark. A brief detail of training and testing data size is given in Table \ref{table:trainSizes}. The circuit's evolutionary process and architecture are shown in Figure \ref{fig:SetupCircuit}. A major difference in the implementation of \gls{sr} and digital circuit setup is the fitness evaluation module (shown in a separate block). In essence, this is the part where \gls{ge} implementation differs for both domains. Thus, the grammar, test cases, and fitness evaluation modules are problem-specific. For example, the solutions in \gls{sr} are mathematical models. These are evaluated using testing data, and \gls{rmse} score is calculated without any external tool, unlike in a digital circuit experiment where a \gls{hdl} simulator is required.

Since the solutions in the latter case are circuits made up of \gls{vhdl} code, they need a simulator to be compiled, executed, and evaluated. In this experiment, we use GHDL\footnote{https://github.com/ghdl/ghdl/releases}, an open-source simulator for \gls{vhdl} designs. 
In addition to the GHDL simulator, the evolved solutions (circuits) require a testbench to simulate and subsequently assess the design. This extra step extends the time required for the fitness evaluation process. In this case, a substantial influence on the fitness evaluation process is caused by the speed of the circuit simulator tool and the clock rate offered to evaluate training data instances in a synchronous order. These additional steps of \gls{hdl} simulation make circuit evaluation much slower than software evaluation \cite{hsiung2018reconfigurable}.

The \gls{ge} circuit training procedure, as shown in Figure \ref{fig:SetupCircuit}, involves the usage of training data. We employ yet another testbench to test the evolved circuit against testing data. In all cases, we save the best individuals across each generation and test them after a run is completed. As a result, there is a less frequent requirement to initialize the testing module. 

The design and testing process of benchmarks from both problem domains is similar at an abstract level. As outlined in Section \ref{sec:GE}, GE individuals are mapped from binary strings to executable structures using grammars.

In this experiment, we used a machine with an Intel i5 CPU @ 1.6GHz with 4 cores, 6 MB cache, and 8GB of RAM. The test case selection algorithm is coded in Python. The \gls{ge} experiments are run using libGE \cite{Nicolau:libGE}. LibGE is a C++ library for \gls{ge} that is implemented and maintained by the Bio-computing and Developmental Systems (BDS) research group at the University of Limerick.

\section{Results and Discussion}
\label{sec:results}
The results from the benchmarks are shown in Table \ref{tab:bench_Synthetic}-\ref{tab:bench_realWorld} (SR) and 
Table \ref{tab:bench_circuit} (digital circuits) and are analyzed in this section. We also perform analysis on GE run time and solution size.

\subsection{Test Scores for SR}
\begin{figure}[!h]
    \centering
    \subfigure[Keijzer-4\label{fig:Test_Keijzer4}]{\includegraphics[width=0.32\textwidth]{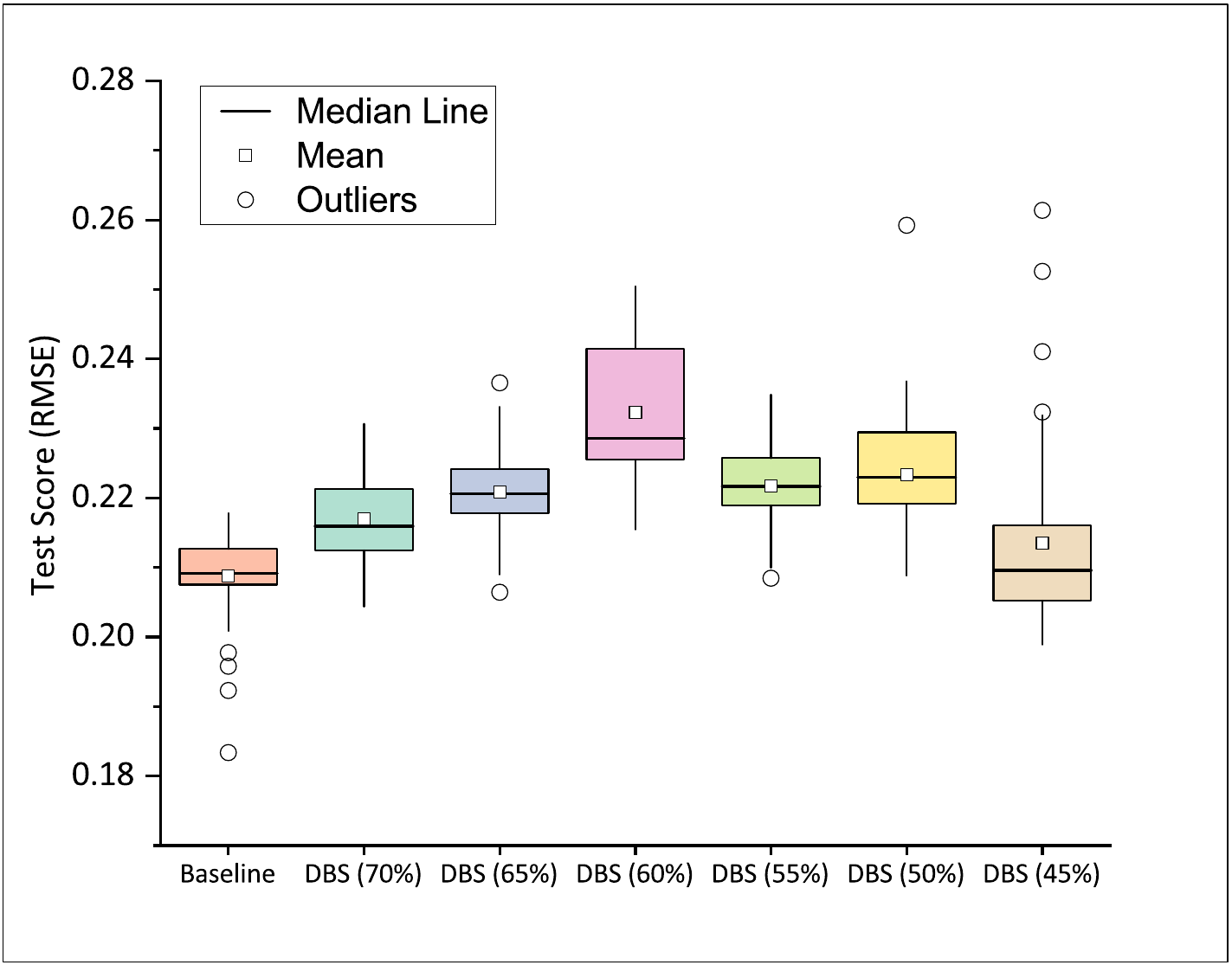}}
    \subfigure[Keijzer-9\label{fig:Test_Keijzer9}]{\includegraphics[width=0.32\textwidth]{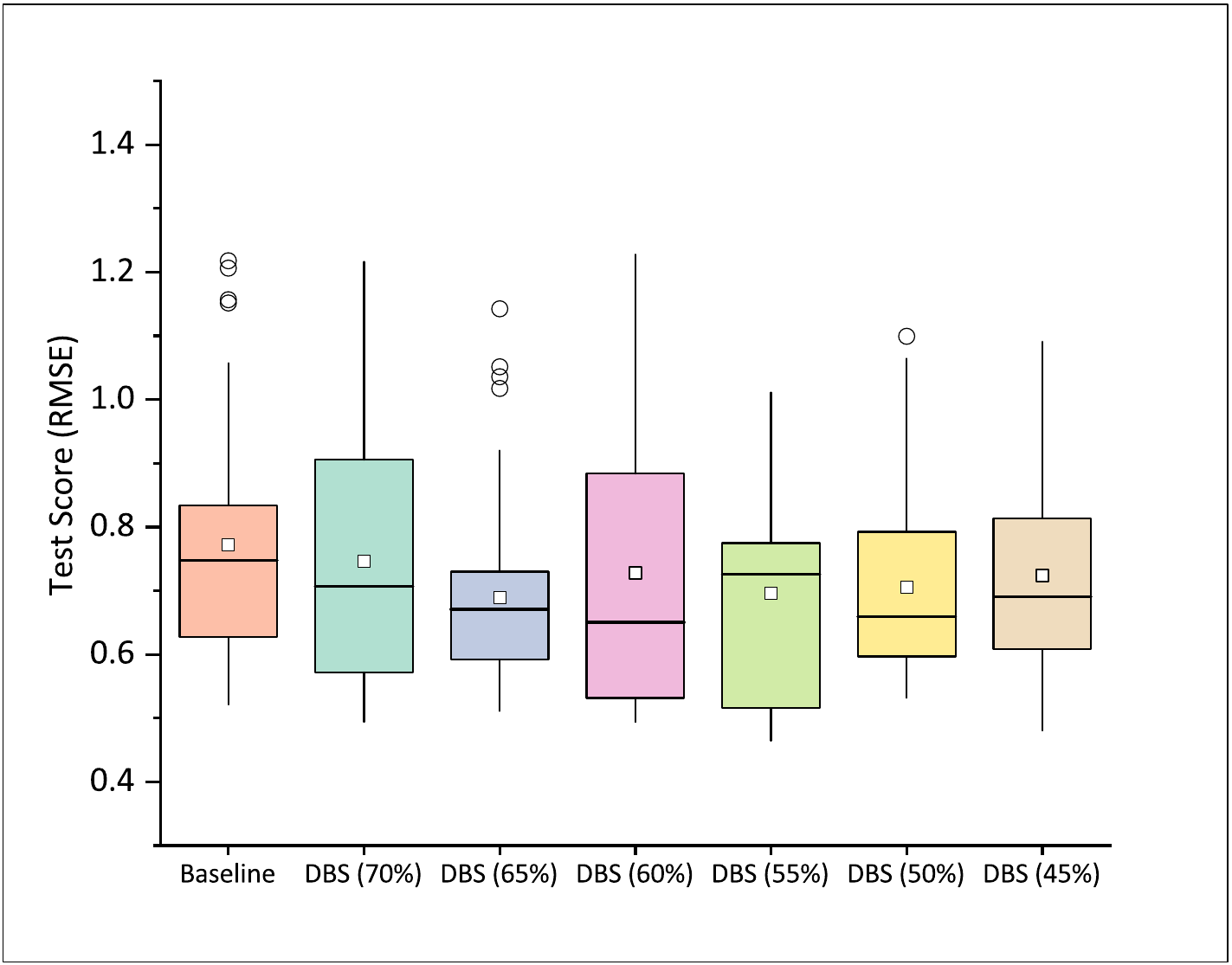}}
   \subfigure[Keijzer-10\label{fig:Test_Keijzer10}]{\includegraphics[width=0.32\textwidth]{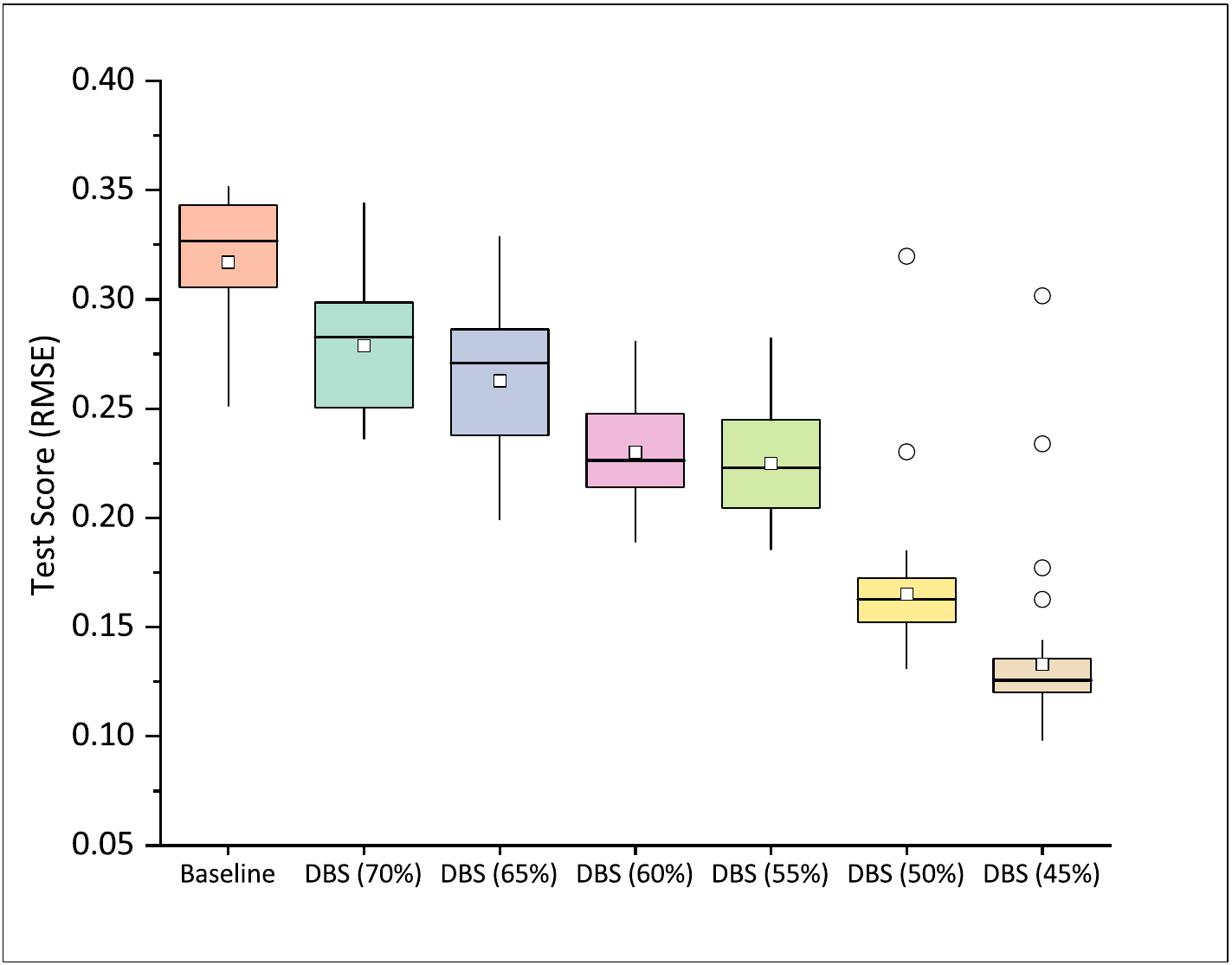}}\\
    \subfigure[Keijzer-14\label{fig:Test_Keijzer14}]{\includegraphics[width=0.32\textwidth]{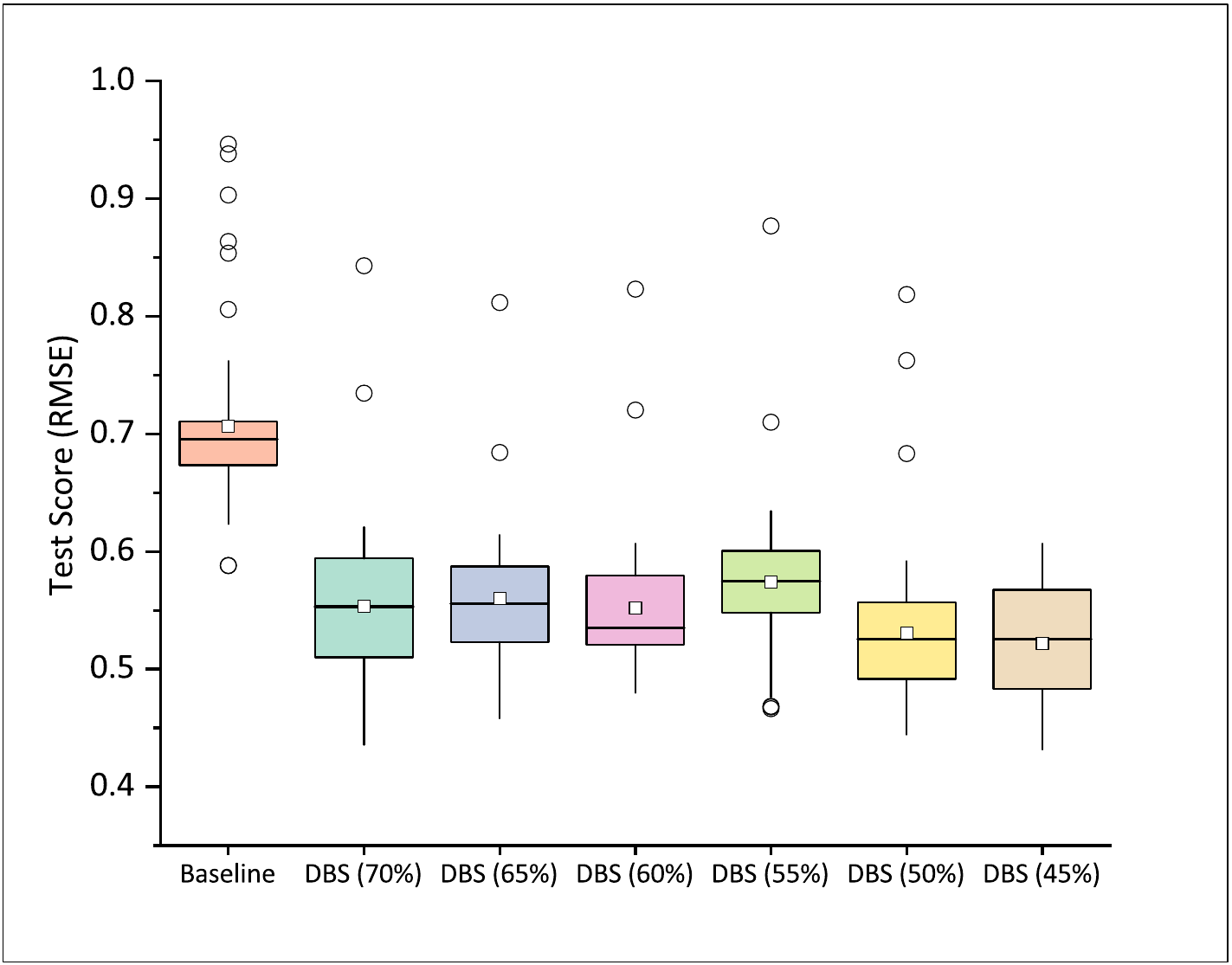}}
    \subfigure[Nguyen-9\label{fig:Test_Nguyen9}]{\includegraphics[width=0.32\textwidth]{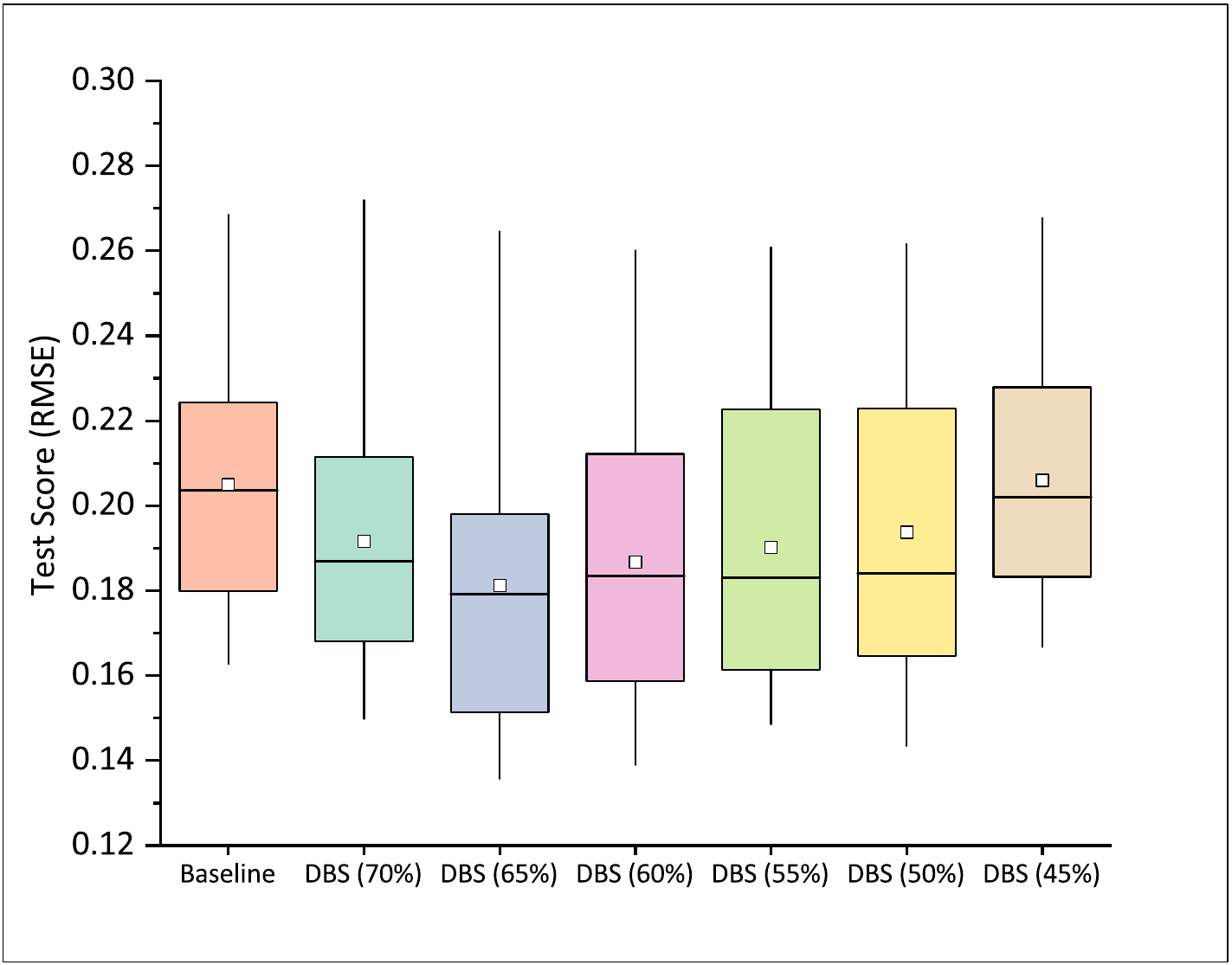}}
    \subfigure[Nguyen-10\label{fig:Test_Nguyen10}]{\includegraphics[width=0.32\textwidth]{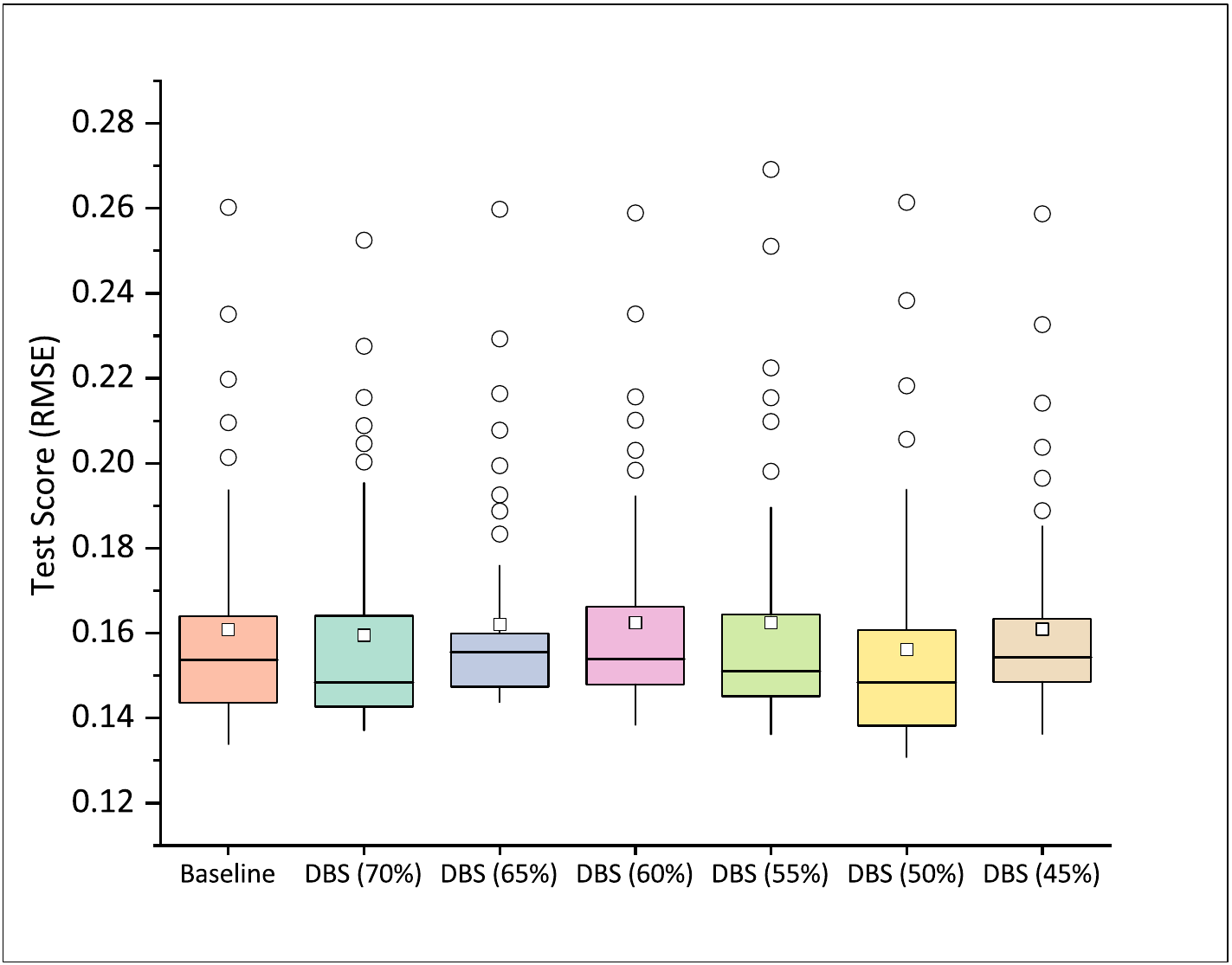}}\\
     \subfigure[Keijzer-5\label{fig:Test_Keijzer5}]{\includegraphics[width=0.32\textwidth]{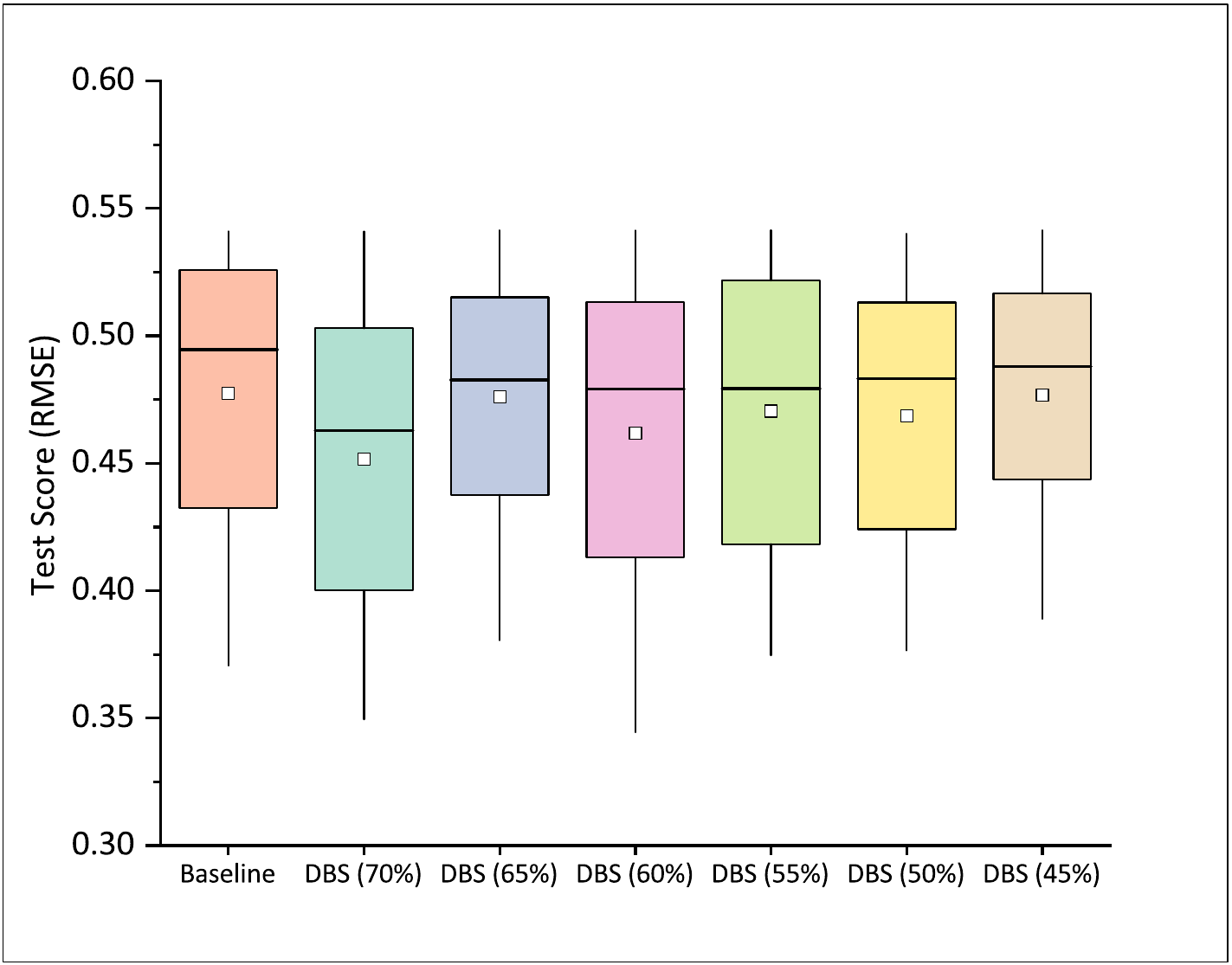}}
    \subfigure[Vladislavleva-5\label{fig:Test_Vlad5}]{\includegraphics[width=0.32\textwidth]{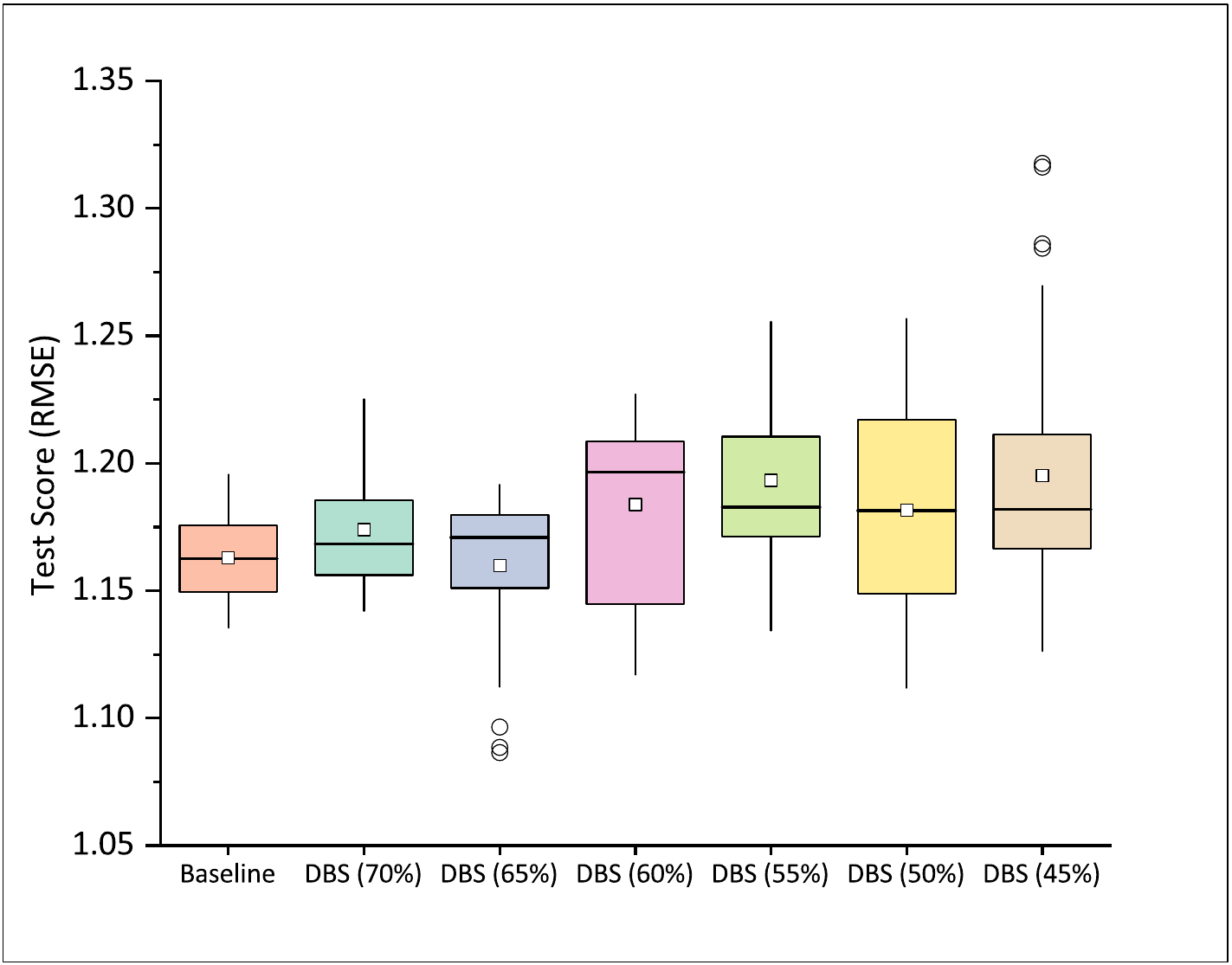}}
    \subfigure[Korns-11\label{fig:Test_korns11}]{\includegraphics[width=0.32\textwidth]{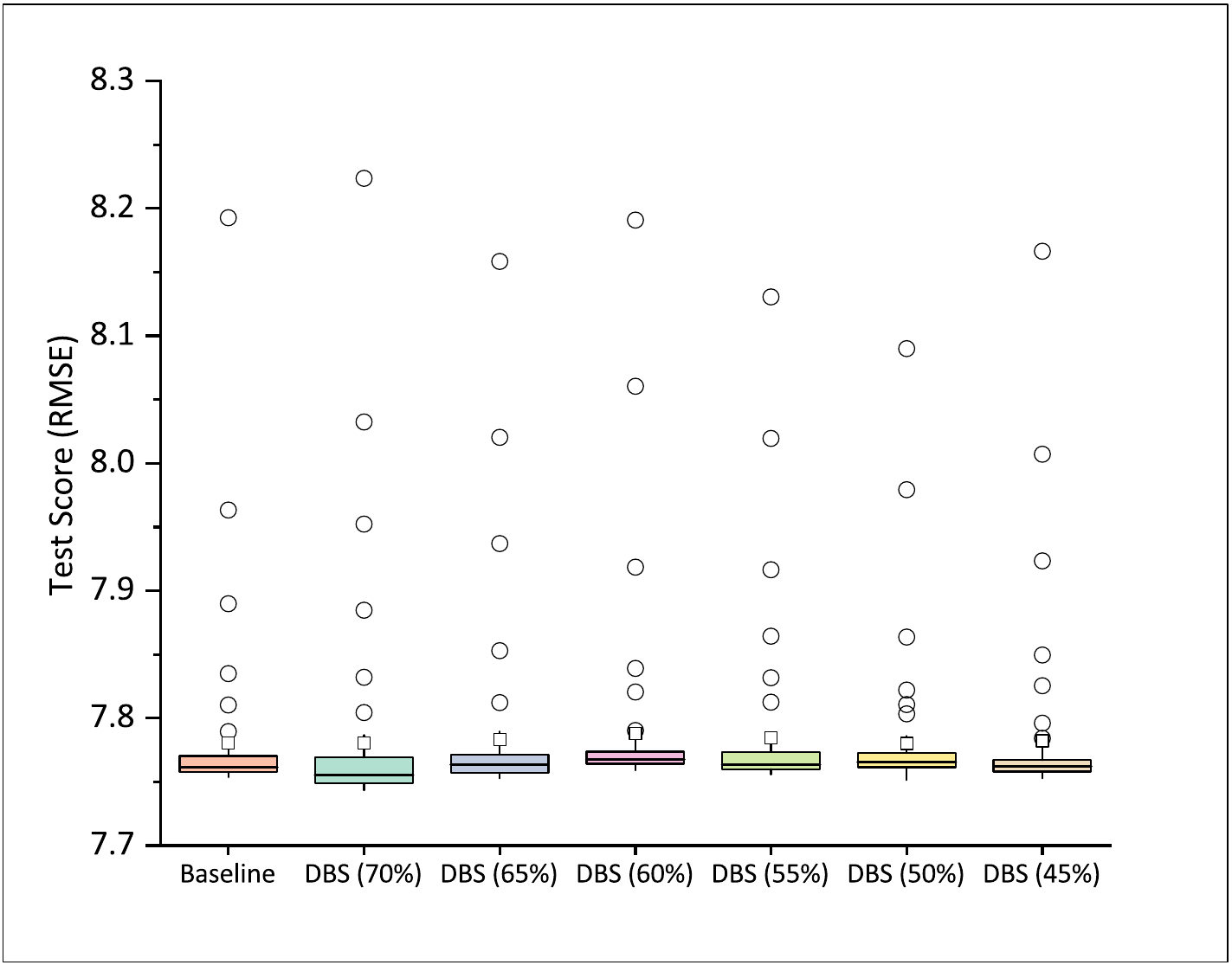}}\\
    \subfigure[Korns-12\label{fig:Test_korns12}]{\includegraphics[width=0.32\textwidth]{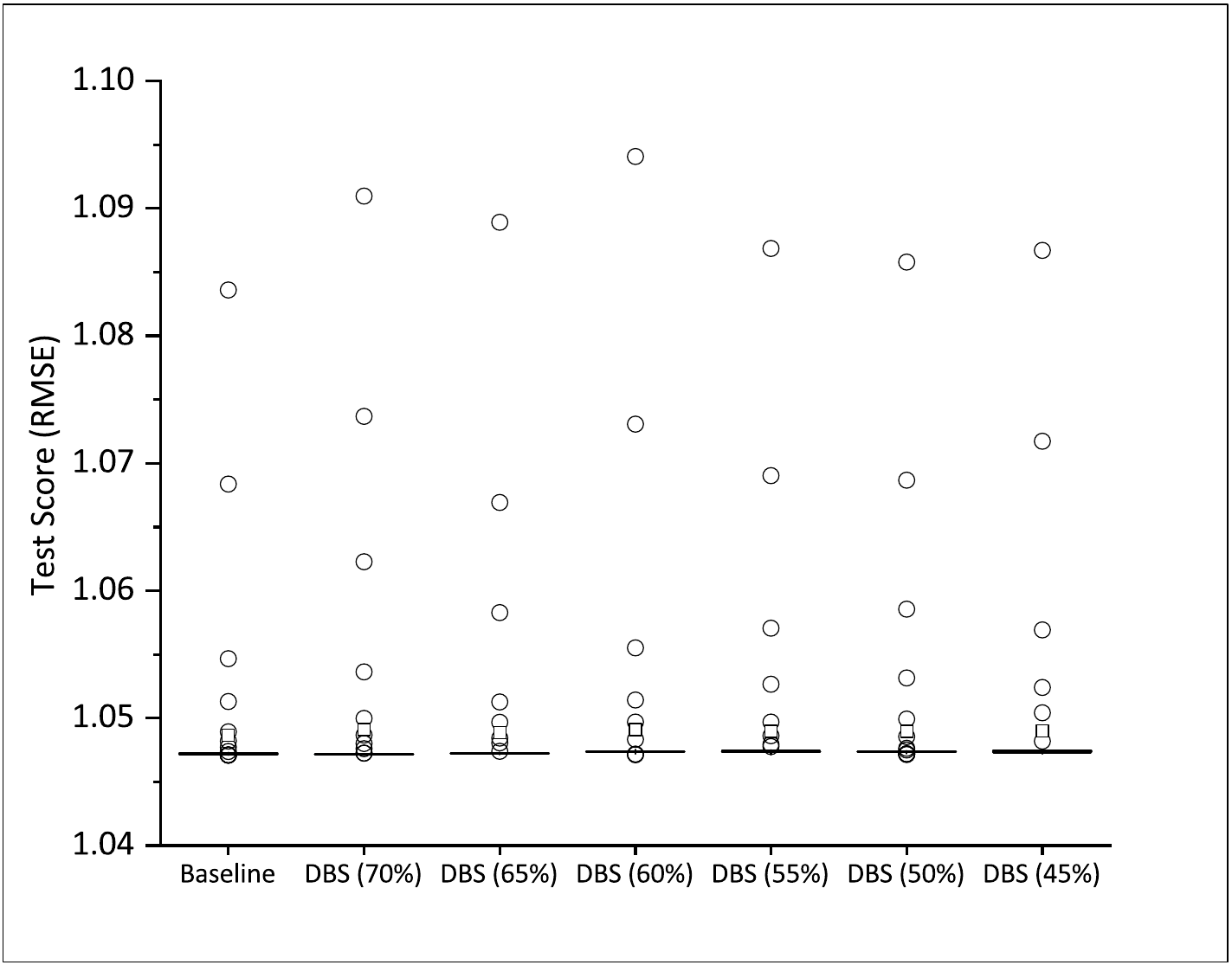}}
    \caption{Mean test score of best solutions across 30 independent runs obtained on synthetic \gls{sr} problems.}
    \label{fig:SynTest}
\end{figure}

Figure \ref{fig:SynTest} shows the testing score for the synthetic SR benchmarks. The test score of the best individuals trained using different subsets of DBS training data against the baseline is shown. In each of the \texttt{Keijzer-9}, \texttt{Keijzer-10}, \texttt{Keijzer-14}, and \texttt{Nguyen-9} problems, most DBS-trained solutions perform better than the baseline. In contrast, in \texttt{Nguyen-10}, \texttt{Keijzer-5}, \texttt{Korns-11}, and \texttt{Korns-12} problems, the results are competitive, and there is no significant difference in performance between any of the DBS-trained solutions and the baseline. In only one case, Figure \ref{fig:Test_Keijzer4} (\texttt{Keijzer-4}), did the baseline generally outperform the DBS. In the \texttt{Vladislavleva-5} problem, shown in Figure \ref{fig:Test_Vlad5}, the baseline outperformed for all sizes except the budget DBS (65\%). In all cases, the DBS-trained sets were significantly faster to train. See Section \ref{sec:TimeAnalysis} for more details.

\begin{figure}[!h]
    \centering
    \subfigure[Airfoil\label{fig:Test_Airfoil}]{\includegraphics[width=0.32\textwidth]{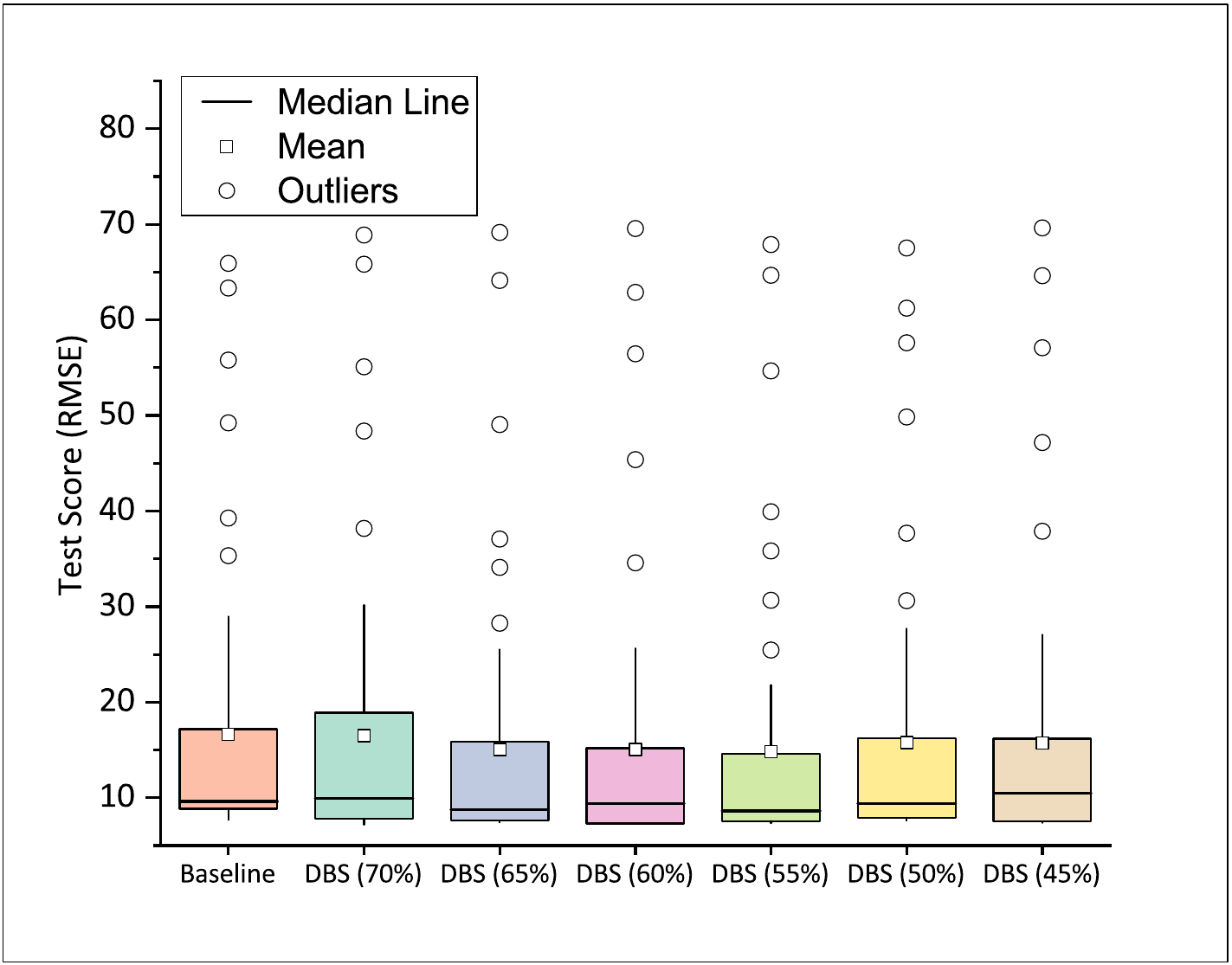}}
    \subfigure[Heating\label{fig:Test_Heating}]{\includegraphics[width=0.32\textwidth]{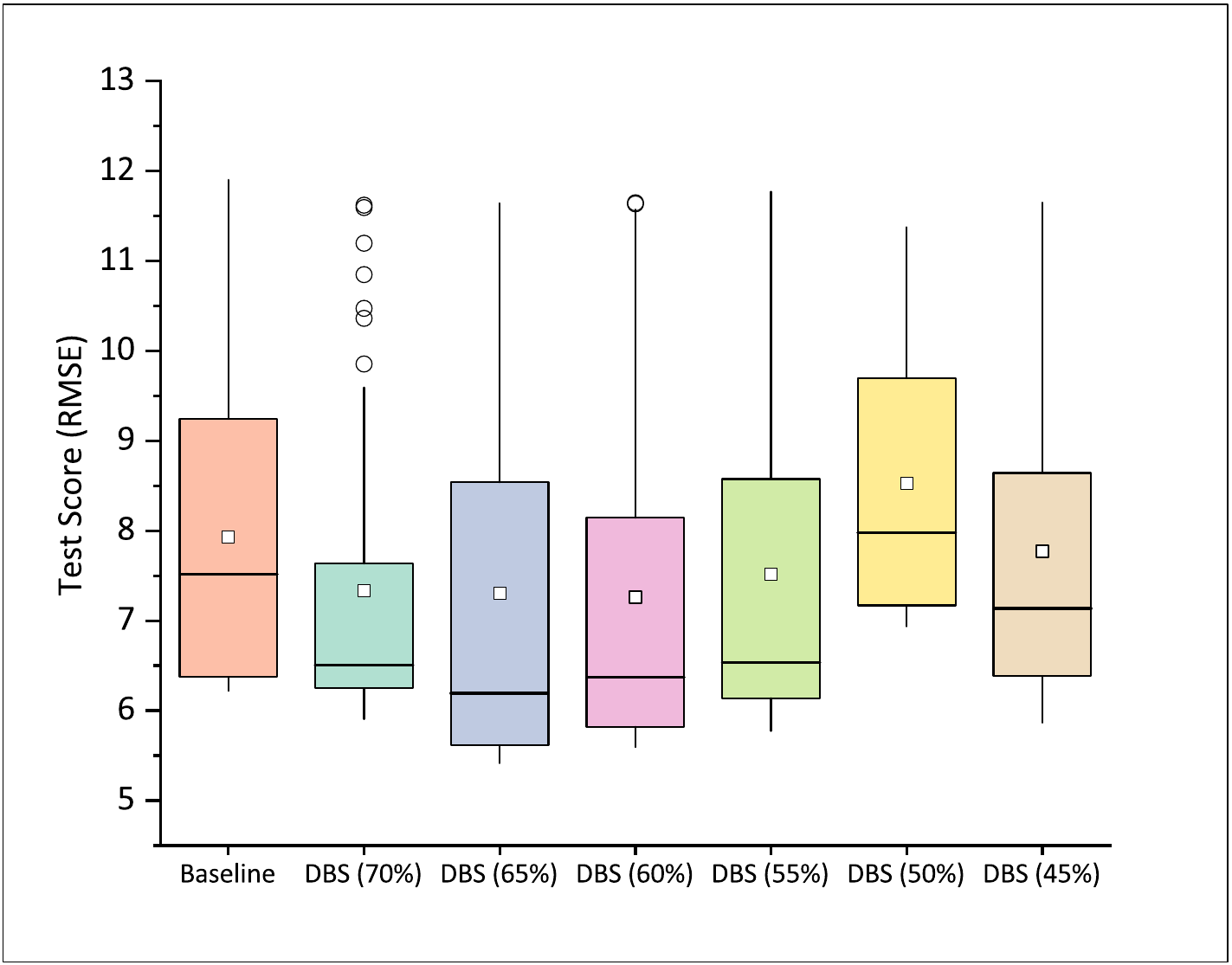}}
   \subfigure[Cooling\label{fig:Test_Cooling}]{\includegraphics[width=0.32\textwidth]{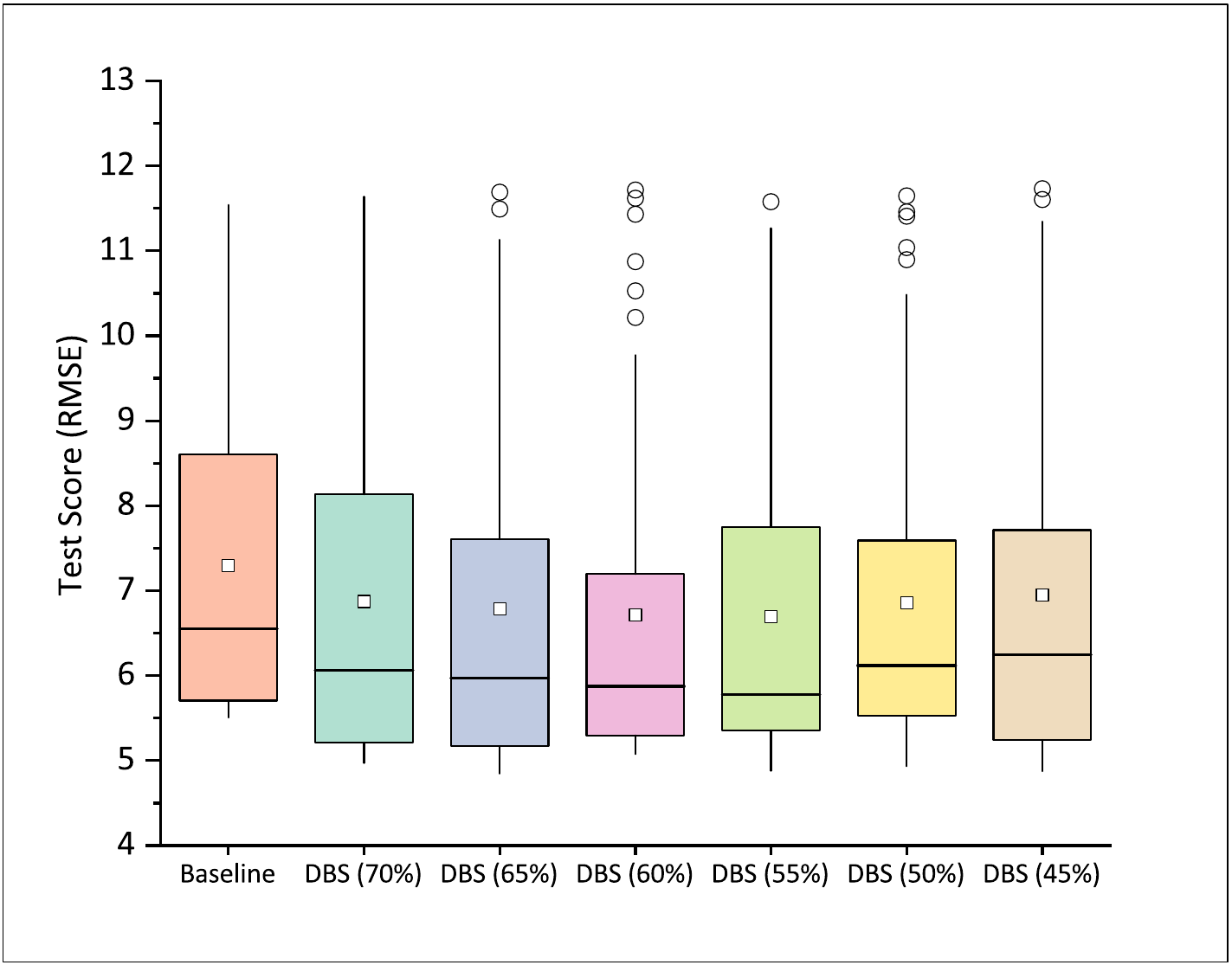}}\\
    \subfigure[Concrete\label{fig:Test_Concrete}]{\includegraphics[width=0.32\textwidth]{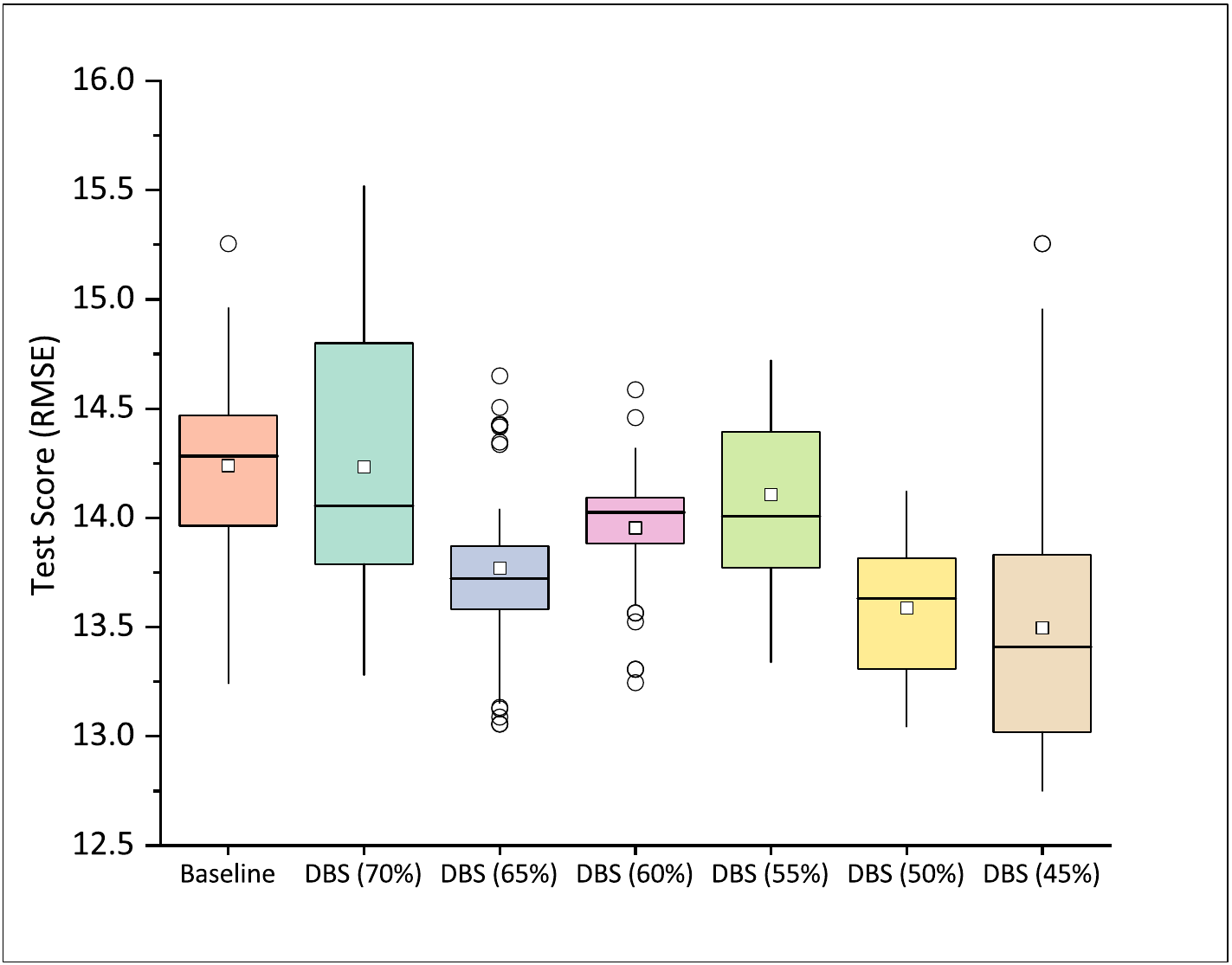}}
    \subfigure[Redwine\label{fig:Test_Redwine}]{\includegraphics[width=0.32\textwidth]{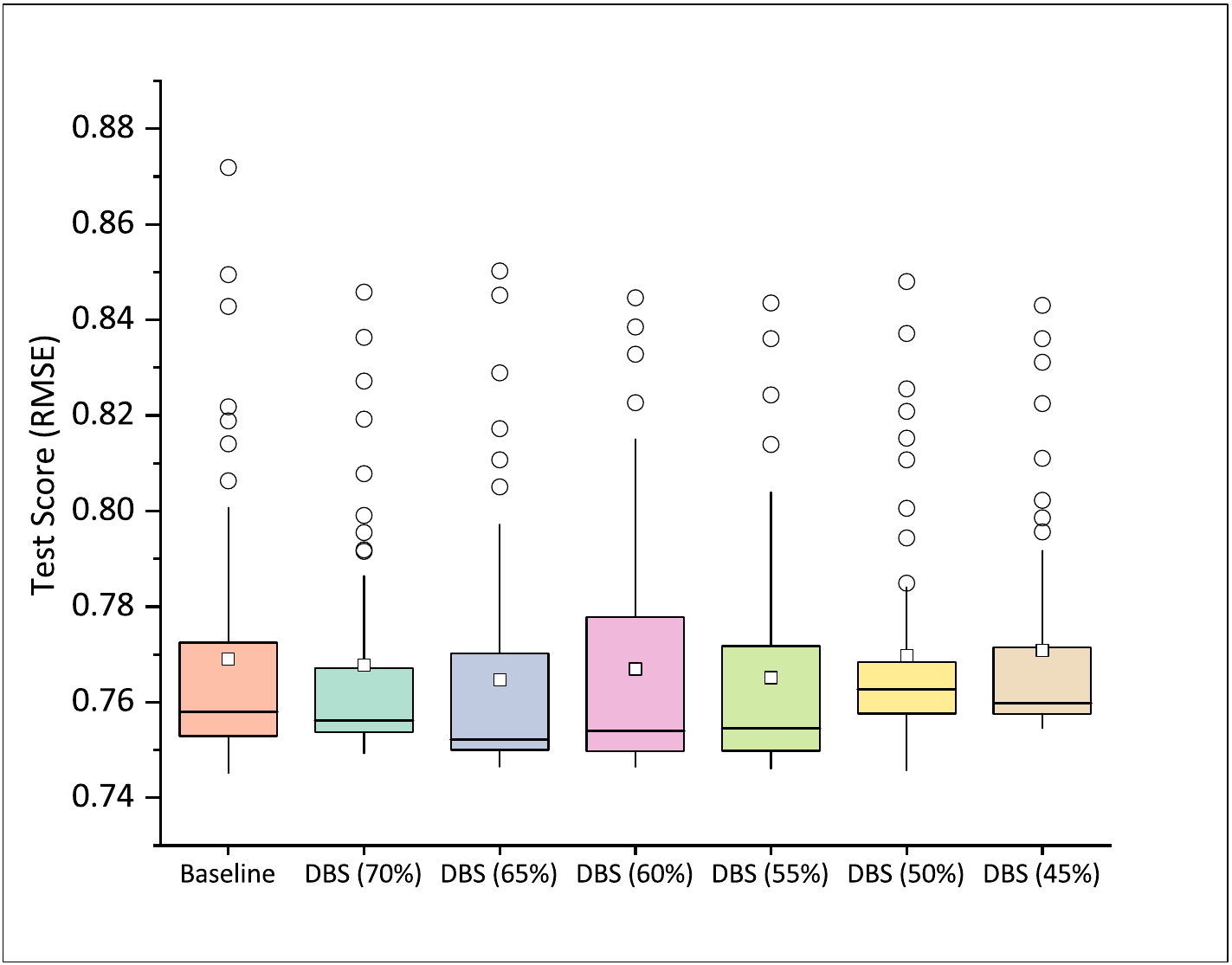}}
    \subfigure[Whitewine\label{fig:Test_Whitewine}]{\includegraphics[width=0.32\textwidth]{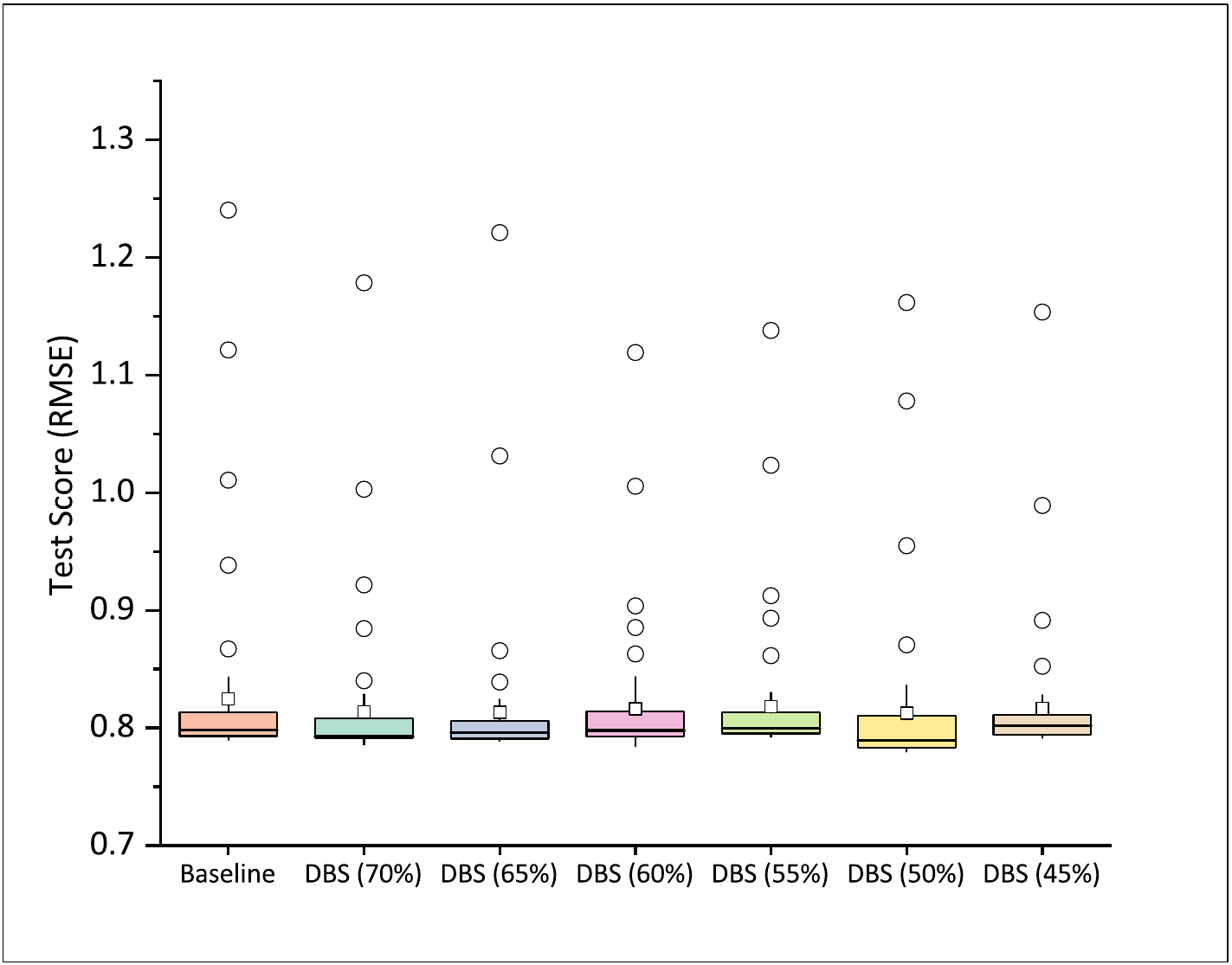}}\\
     \subfigure[Housing\label{fig:Test_Housing}]{\includegraphics[width=0.32\textwidth]{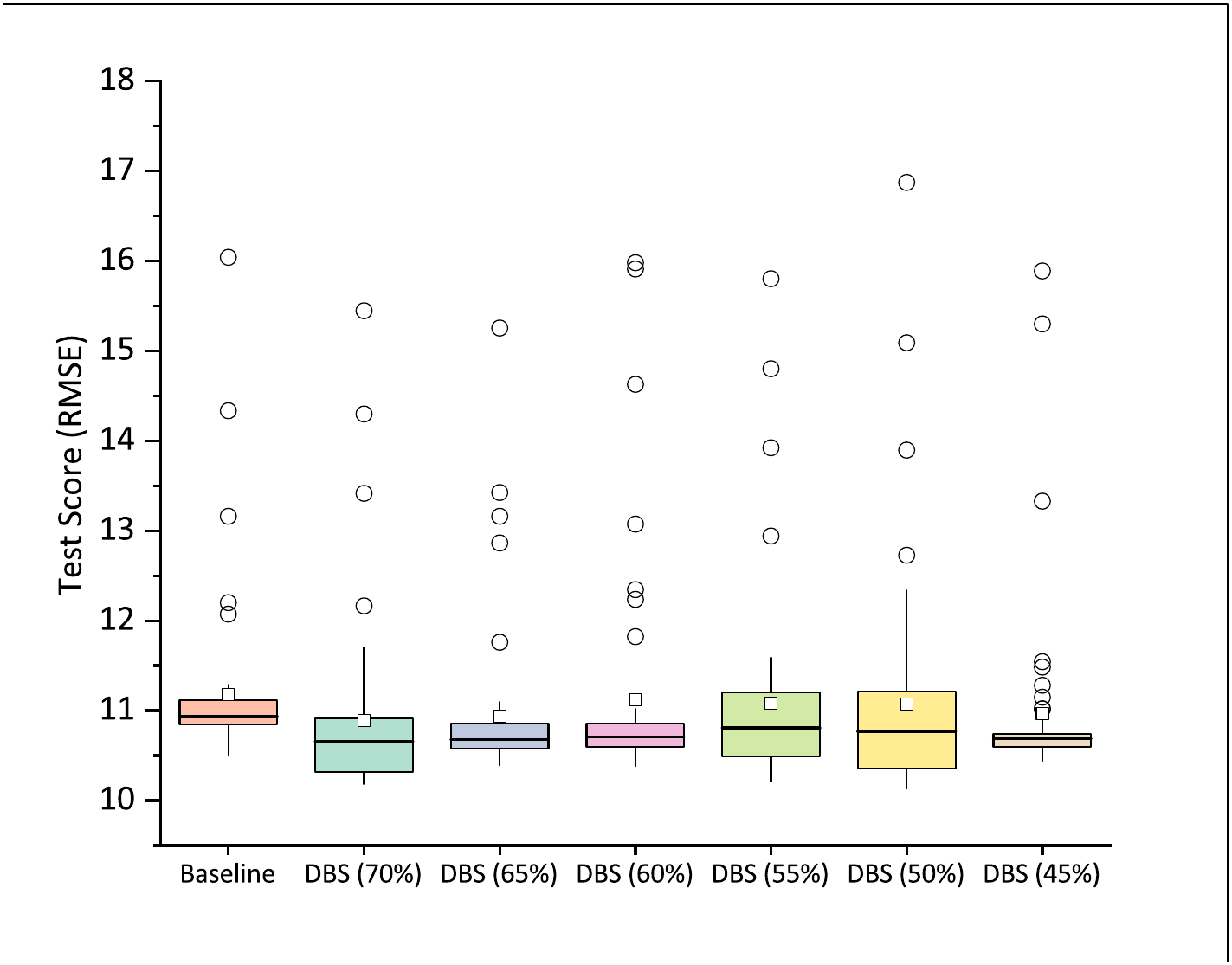}}
    \subfigure[Pollution\label{fig:Test_Pollution}]{\includegraphics[width=0.32\textwidth]{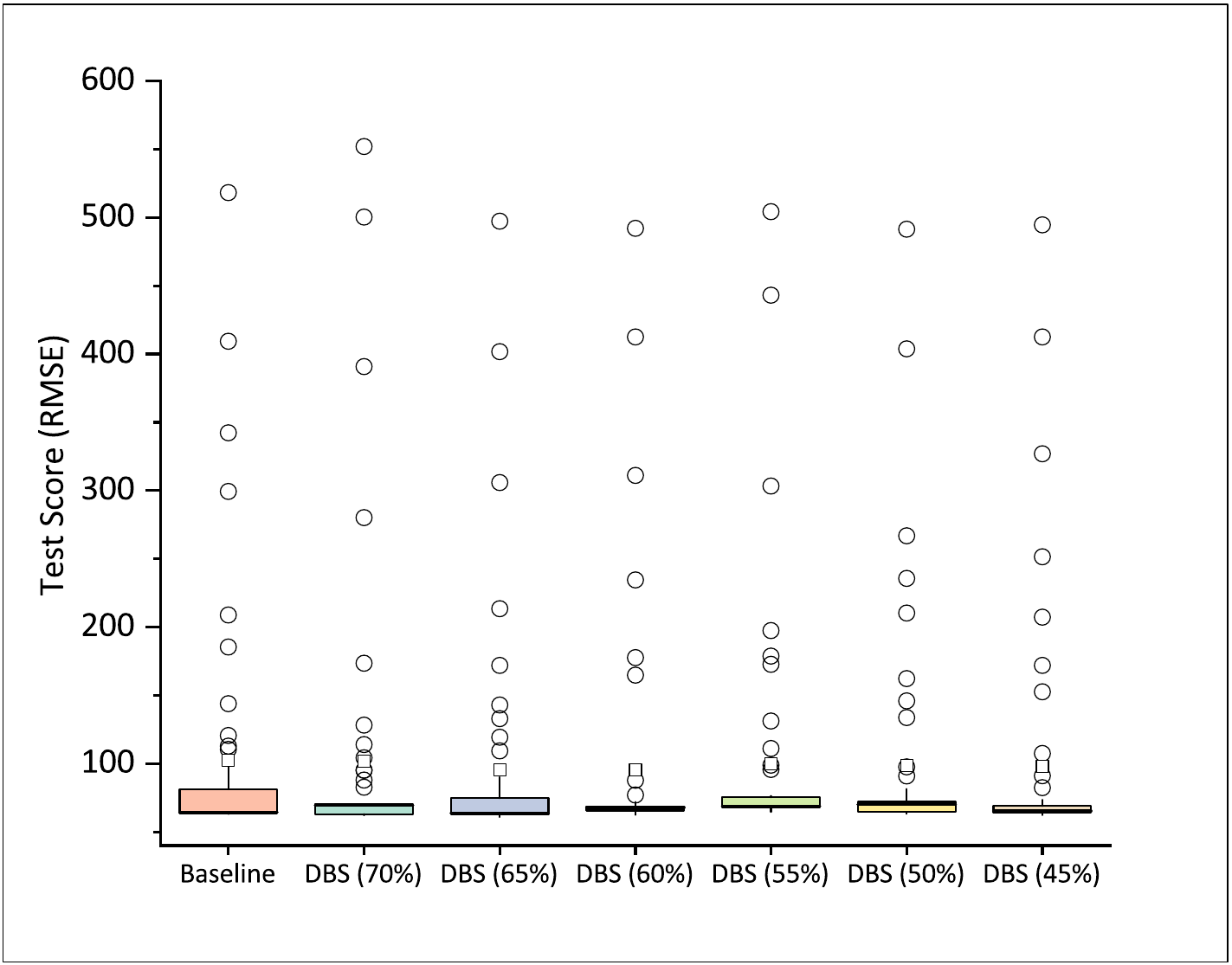}}
    \subfigure[Dowchem\label{fig:Test_Dowchem}]{\includegraphics[width=0.32\textwidth]{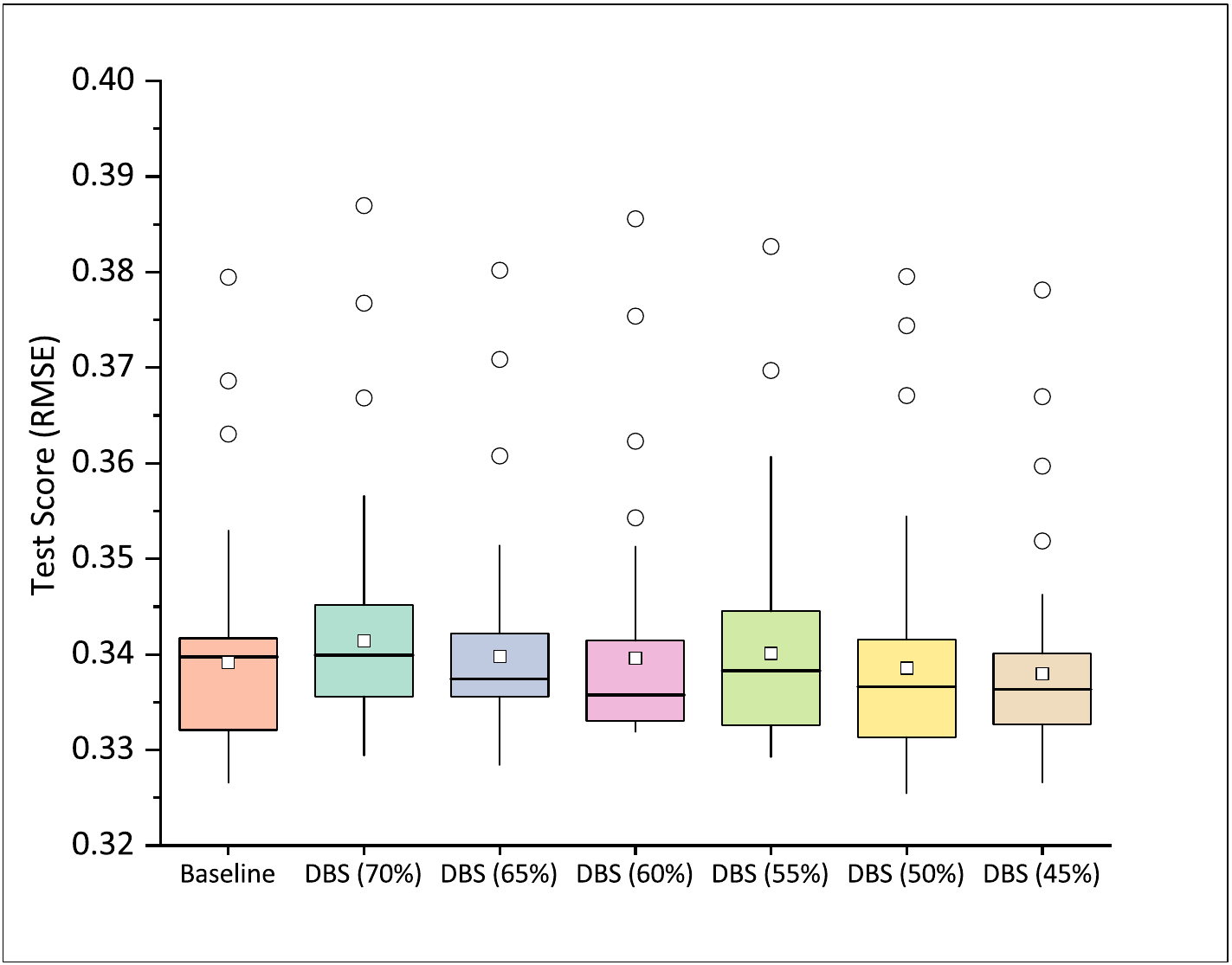}}\\
    \subfigure[Crime\label{fig:Test_Crime}]{\includegraphics[width=0.32\textwidth]{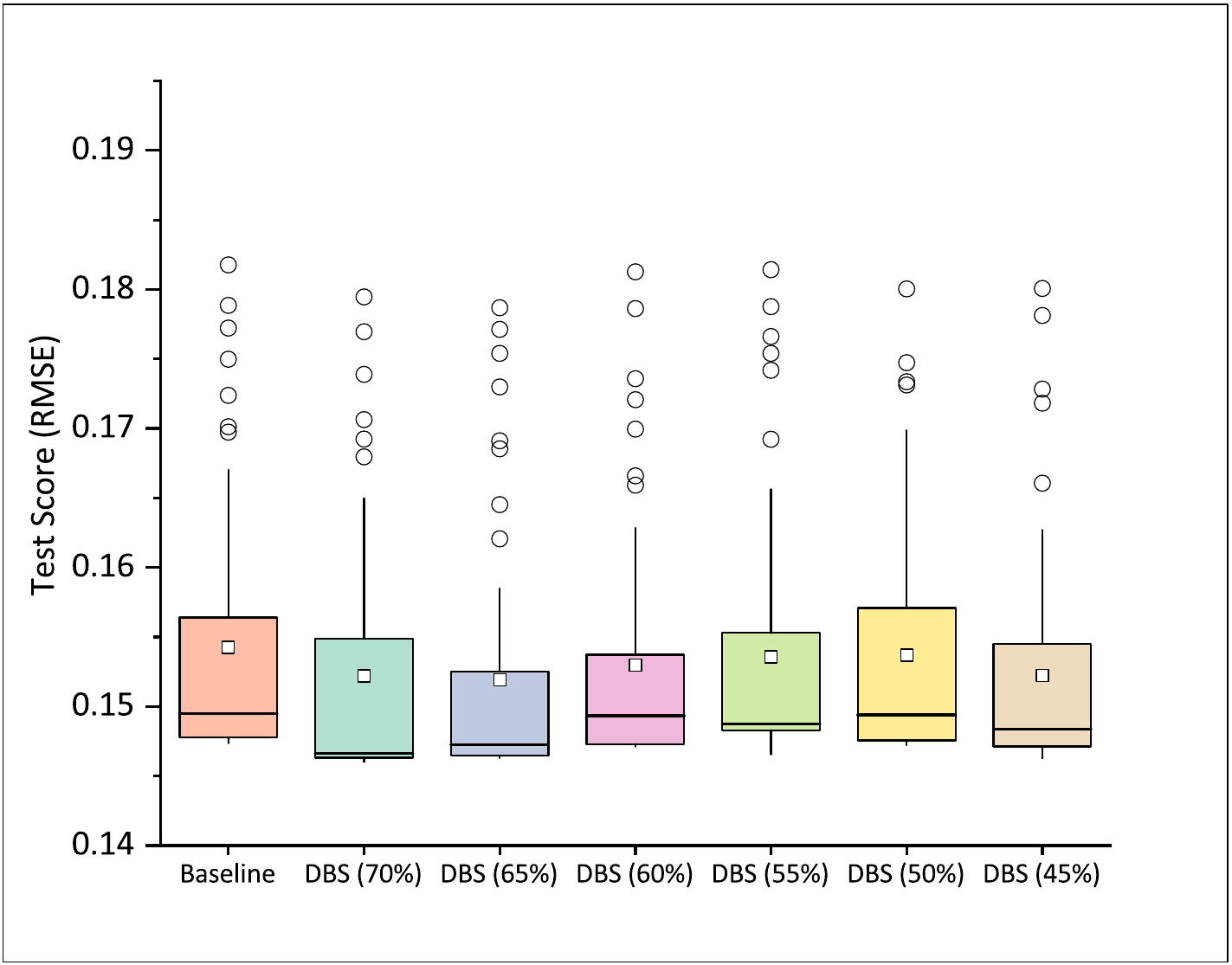}}
    \caption{Mean test score of best solutions across 30 independent runs obtained on real-world \gls{sr} problems.}
    \label{fig:RealTest}
\end{figure}

The test scores of the real-world \gls{sr} benchmarks are shown in Figure \ref{fig:RealTest}. We employed the same strategy used in the previous experiment. The testing scores of the best-trained individuals are reported as the baseline. The training data used in the baseline is used for subset selection using the DBS algorithm. We performed test case selection using the same range of $B$. The test scores of the best individuals trained using training data obtained by the DBS algorithm were tested, and the results were compared. The box plots in Figure \ref{fig:RealTest} show the min, max, mean, and median values of the obtained scores.

In each of the \texttt{airfoil}, \texttt{concrete}, \texttt{whitewine}, \texttt{housing}, and \texttt{pollution} problems, the test score of the best individuals trained using DBS appear to be statistically similar or better than baseline.

In the \texttt{cooling} and \texttt{crime} problems, the DBS-trained solutions outperformed the baseline. In the case of \texttt{heating}, the DBS-trained solutions generally outperform the baseline for all sizes except at DBS (50\%). The solutions trained on DBS data demonstrate superior performance over the baseline for \texttt{redwine}, except when $B$ was set at 50\% and 45\%.

For \texttt{dowchem} problem, the solutions trained on DBS data exhibit higher test scores than the baseline, except when $B$ was set at 70\%.

\subsection{Test Scores for Circuit Design}

\begin{figure}[!h]
    \centering
    \subfigure[Comparator\label{fig:Test_Comp}]{\includegraphics[width=0.4\textwidth]{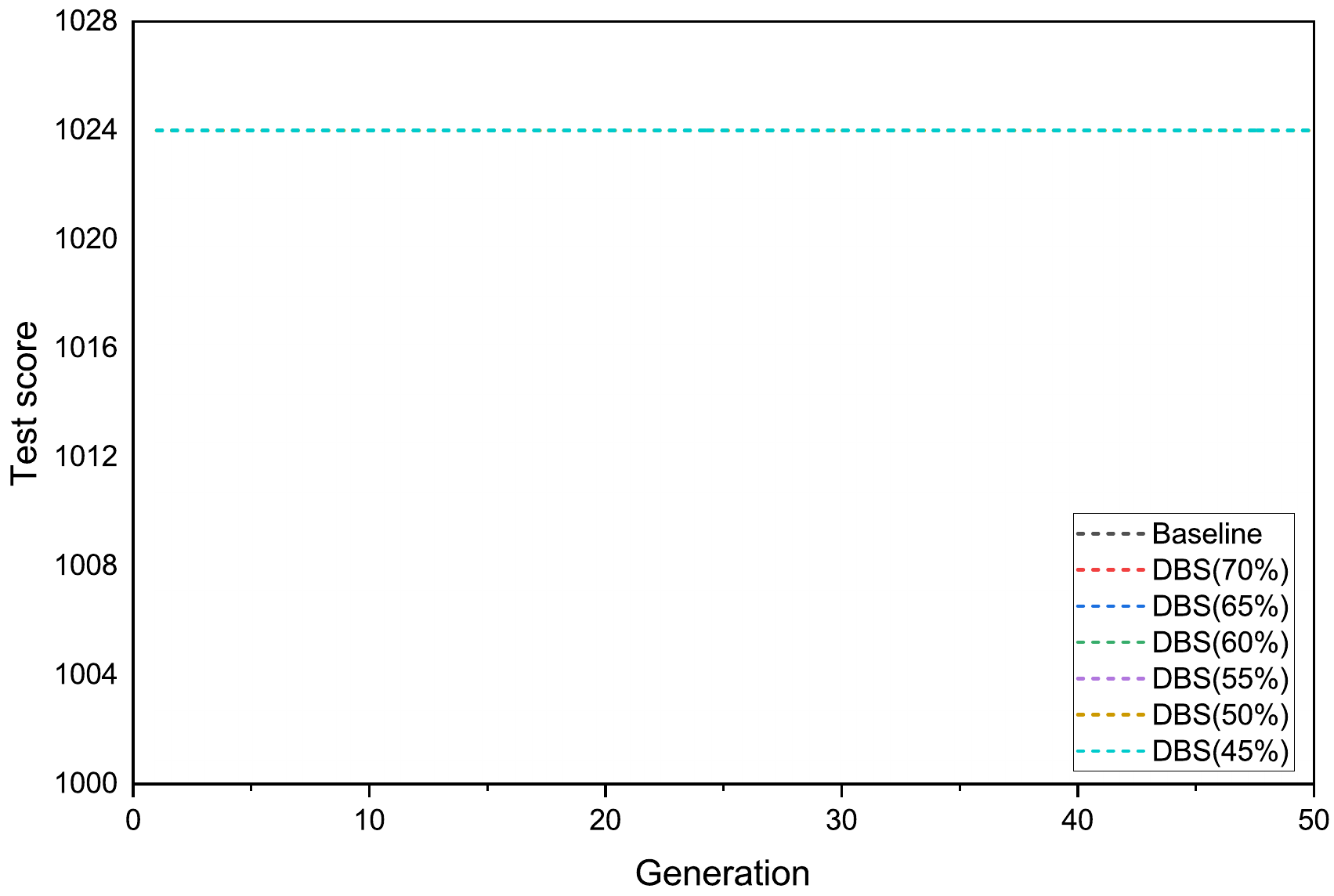}}
    \subfigure[Parity\label{fig:Test_5Parity}]{\includegraphics[width=0.4\textwidth]{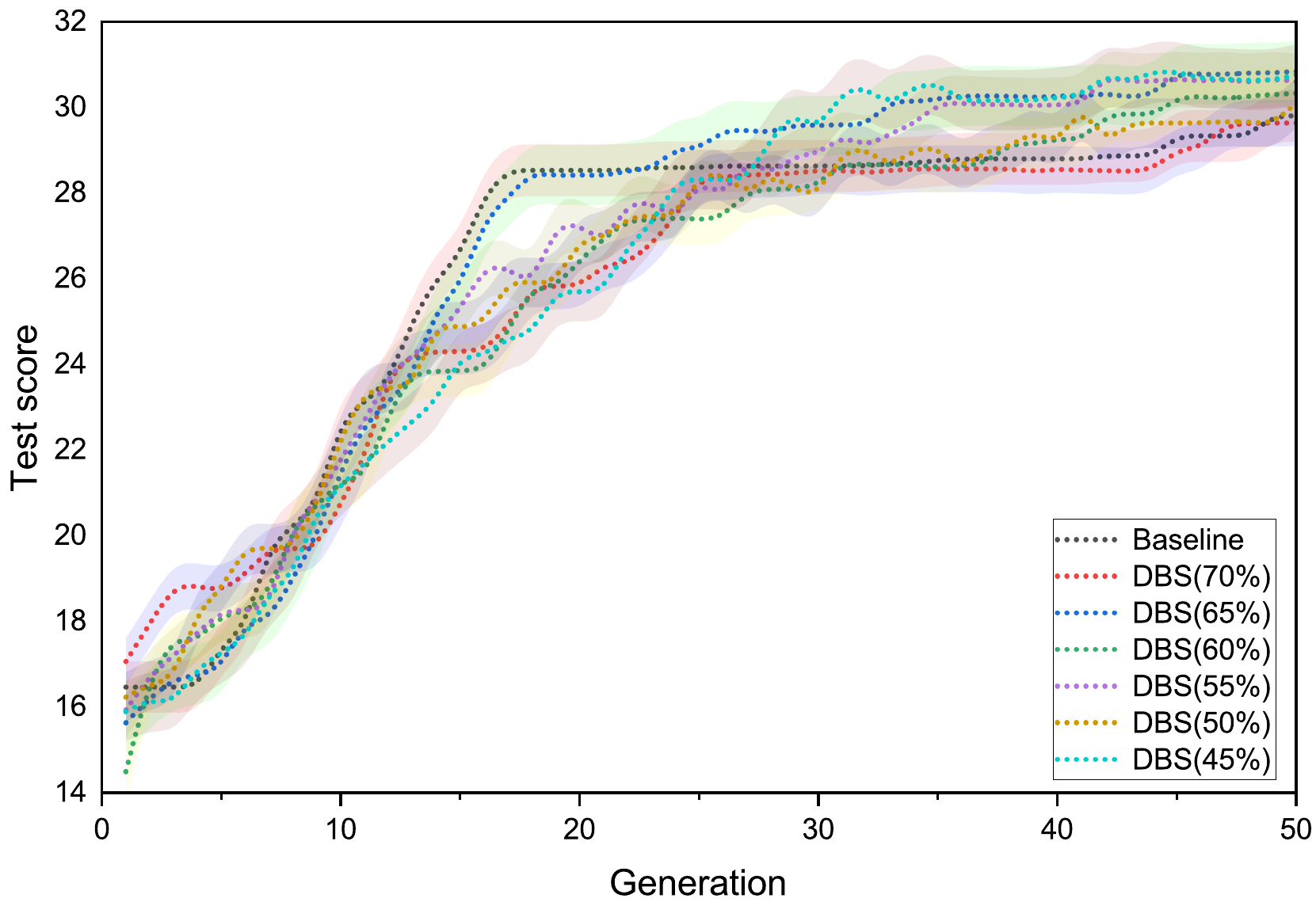}}\\
    \subfigure[Multiplexer\label{fig:Test_Mux}]{\includegraphics[width=0.4\textwidth]{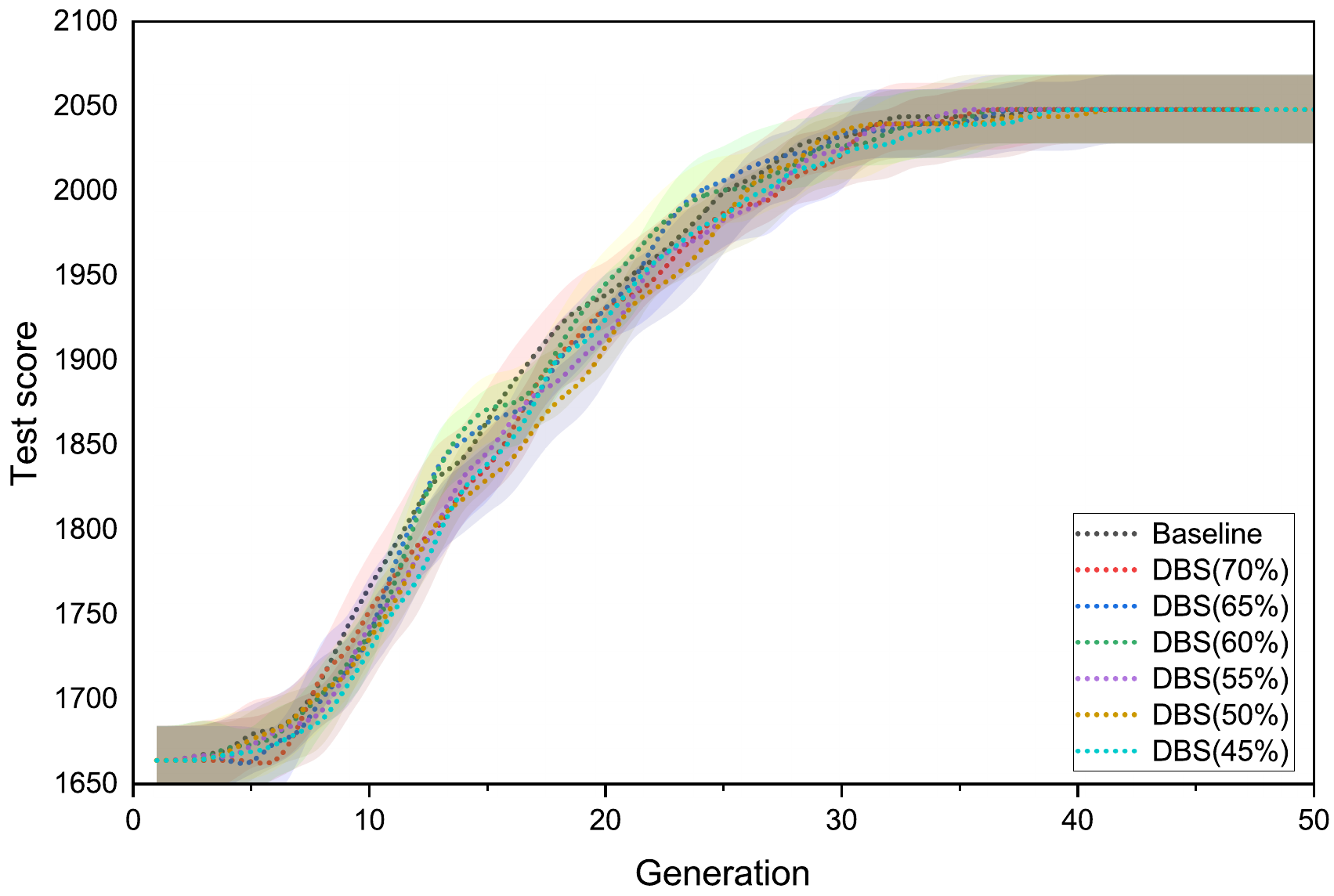}}
    \subfigure[ALU\label{fig:Test_ALU}]{\includegraphics[width=0.4\textwidth]{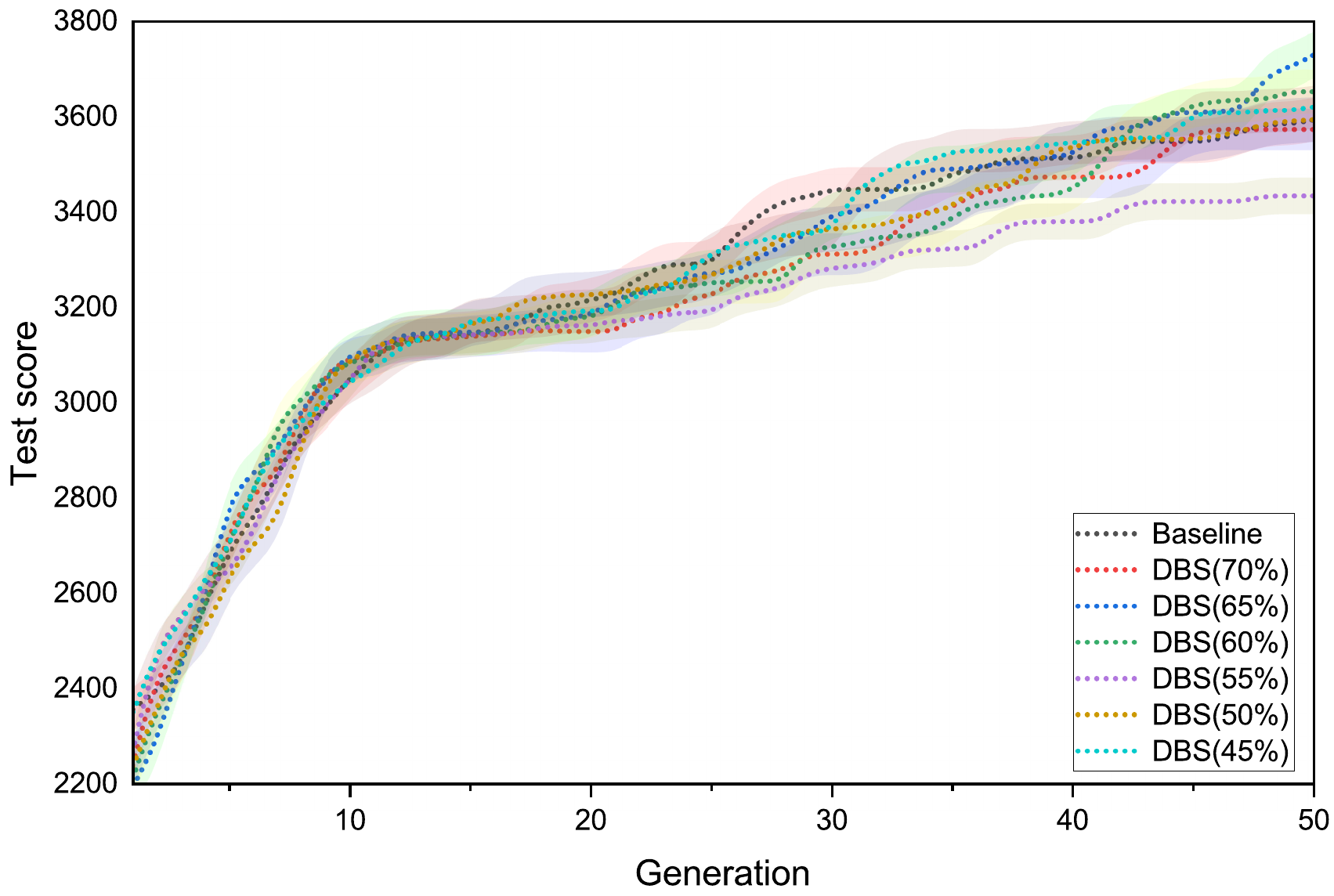}}
    \caption{Mean test score (with std error) of best solutions vs. generation obtained from 30 independent runs on digital circuit problems. Overlapping lines in (a) signify that all training subsets exhibited similar performance.}
    \label{fig:CircuitTest}
\end{figure}

Figure \ref{fig:CircuitTest} displays the outcome for digital circuit benchmarks. It should be noted that, unlike \gls{sr}, where the \gls{rmse} score is calculated, the test score is determined by the number of instances passed out of the total available in the testing data. As we compare the test scores, we plot a generation versus test score graph of the best individuals obtained. This would be helpful in identifying which set of training data could guide a better search. However, we do present the mean of best test score, as we did in the \gls{sr} experiment during the statistical test.

Figure \ref{fig:Test_Comp} represents the test score of the best solutions obtained across 50 generations for the \texttt{comparator}. The mean best test scores of the best individuals across 30 independent runs were observed to be the same. The problem seems trivial, and all experiments on the \texttt{comparator} performed similarly well. 

The test scores of the solutions obtained using the \texttt{parity} dataset are presented in Figure \ref{fig:Test_5Parity}. The best individuals trained using DBS (70\%) could pass as much of the testing data as were passed by the best solution obtained on baseline training data. In the remaining cases, the DBS-trained data produced solutions with higher test coverage than those obtained in the baseline experiment. 

In the case of the \texttt{multiplexer}, the best solutions obtained in the 50th generation were able to maintain a similar test score, as shown in Figure \ref{fig:Test_Mux}. However, close observation of Figure \ref{fig:Test_Mux} indicates that experiments using DBS training data (in some cases) could achieve better test scores in earlier generations than the baseline. 

Figure \ref{fig:Test_ALU} shows the mean best test score of solutions obtained on \texttt{\gls{alu}} dataset. The experiments using DBS training data in most of the cases, except DBS (70\%) and DBS (55\%), were able to get solutions that could achieve better test scores than the solutions obtained in the baseline.

The test results for 24 benchmarks are shown in Figure \ref{fig:SynTest}-\ref{fig:CircuitTest}. We examined the results regarding the best-trained individuals' test scores and showed how they compared to the outcomes of their respective baseline experiments. The test scores presented using box plots and line graphs are good for insight into the performance of the different sets of training data used and are reliable to a certain extent. However, we need some strong mathematical evidence before coming to a conclusion. For this, we perform statistical tests to be able to fairly compare the different methods and validate the results.

We assessed the normality of the data using the Shapiro-Wilk Test at a significance level of $\alpha$=0.05. The results indicate that the data samples do not meet the assumptions necessary for parametric tests, as they do not exhibit a normal distribution. In such cases, non-parametric tests are more appropriate. Non-parametric tests make fewer assumptions about the data distribution and are, therefore, more robust to violations of normality assumptions. 

As the results obtained from distinct training datasets are mutually independent, the Wilcoxon rank-sum\footnote{The Wilcoxon rank-sum test is a non-parametric test that does not rely on the assumption of normality. Shapiro-Wilk Test was performed to check if a more appropriate test could be applied.} (a non-parametric test) is thus employed to determine whether there is any statistically significant difference between the test scores of solutions obtained. We use a significance level of 5\% (i.e.,  $5.0\times 10^{-2}$) for the two hypotheses. 

To test for a statistical difference between the two samples, we compare the \gls{dbs} approaches with the baseline using the null hypothesis (H0) that `there is no difference between the test scores of the baseline and \gls{dbs} approaches.' If the statistical analysis does not reveal a significant difference, we label the outcome as equal or similar, denoted by the symbol ``='' to indicate no substantial disparity between the performance of the baseline and \gls{dbs} approaches. Otherwise, we perform another test, and the following hypotheses are tested:

\begin{itemize}
    \item[-] Null Hypothesis (H0): The models trained using the baseline training data have a better test score than the models trained using the \gls{dbs}-selected training data at a budget $B$;
    \item[-] Alternative Hypothesis (H1): The test score of models trained using \gls{dbs} training data is better than the test score of models trained using baseline training data.
\end{itemize}

After the statistical analysis, the results are marked with a ``+'' symbol to indicate significantly better performance or a ``-'' symbol to indicate significantly worse performance. This hypothesis testing framework allows us to determine whether the \gls{dbs}-selected training data leads to improved model performance compared to the baseline training data under the specified budget constraint.

Table \ref{tab:SynWilcoxonTest} shows the mean test average scores of the best individuals obtained across 30 independent runs. Note that the mean test scores are rounded to 4 decimal points. This can make, sometimes, the test scores at two or more DBS budgets appear to be the same, but in reality, it is extremely uncommon to obtain the same test score after carrying out a large number of runs, in our case, 30 runs. All DBS test results were statistically tested against the corresponding baseline. The p-values are calculated, and accordingly, the output is reported in Table \ref{tab:SynWilcoxonTest}. 
\begin{table}[!h]
    \centering
    \caption{Results of the Wilcoxon test for SR \textbf{synthetic benchmarks} considering a significance level of $\alpha = 0.05$. Values shown are the mean best test scores across 30 independent runs. The symbols +, =, - indicate whether the corresponding results for DBS are significantly better, not significantly different, or worse than baseline, respectively. The last row summarizes this information.}
    \label{tab:SynWilcoxonTest}
    \resizebox{\columnwidth}{!}{
    \begin{tabular}{l c c c c c c r}
    
    \cmidrule(lr){1-8}
{Benchmarks} & {Baseline} & {DBS (70\%)} & {DBS (65\%)} & {DBS (60\%)} & {DBS (55\%)} & {DBS (50\%)} & {DBS (45\%)}\\
    \cmidrule(lr){1-8}
Keijzer-4  & {0.2088} & 0.217- & 0.2208- & 0.2323- & 0.2217- & 0.2234- & 0.2135=\\
Keijzer-9  & {0.7724} & {0.7462}+ & {0.6893}+ & {0.7278}+ & {0.6955}+ & {0.7059}+ & {0.7238}+ \\
Keijzer-10 & {0.317}  & {0.2788}+ & {0.2628}+ & {0.2301}+ & {0.225}+  & {0.1651}+      & {0.133}+\\
Keijzer-14 & {0.7064} & {0.5533}+ & {0.5603}+ & {0.552}+  & {0.5742}+ & {0.5308}+ & {0.522}+\\
Nguyen-9   & {0.205}  & {0.1915}+ & {0.1812}+ & {0.1868}+ & {0.1902}+ & {0.1937}+ & 0.206=        \\
Nguyen-10  & {0.1608} & {0.1595}+ & 0.162=        & 0.1625- & 0.1625- & {0.1562}+ & 0.1609=\\
Keijzer-5  & {0.4774} & {0.4515}+ & 0.4761= & {0.4617}+ & {0.4705}+ & {0.4686}+ & 0.4767= \\
Vladislavleva-5 & {1.1629} & 1.1739- & 1.1599= & 1.1837- & 1.1933- & 1.1816- & 1.1952-\\
Korns-11   & {7.7804} & {7.7806}+ & 7.783- & 7.7877- & 7.7846- & 7.7802- & 7.7821= \\
Korns-12   & {1.0486} & 1.0491= & 1.0488- & 1.049- & 1.049- & 1.049- & 1.049- \\
\cmidrule(lr){1-8}
Total +/=/-   & -- & 7/1/2 & 4/3/3 & 5/0/5 & 5/0/5  & 6/0/4 & 3/5/2 \\
\cmidrule(lr){1-8}
\end{tabular}}
\end{table}

Table \ref{tab:statTest_realWorld} summarizes the outcomes of the statistical analysis conducted on real-world \gls{sr} benchmarks. Instead of solely relying on the box plots, this strengthens the results and aids in deciding whether there is sufficient evidence to ``reject'' a hypothesis or to ``support'' the proposed approach.

\begin{table}[!h]
    \centering
    \caption{Results of the Wilcoxon test for SR \textbf{real-world benchmarks} considering a significance level of $\alpha = 0.05$. Values shown are the mean best test scores across 30 independent runs. The symbols +, =, - indicate whether the corresponding results for DBS are significantly better, not significantly different, or worse than the baseline. The last row summarizes this information.}
    \label{tab:statTest_realWorld}
    \resizebox{\columnwidth}{!}{
    \begin{tabular}{l c c c c c c r}
    
    \cmidrule(lr){1-8}
{Benchmarks} & {Baseline} & {DBS (70\%)} & {DBS (65\%)} & {DBS (60\%)} & {DBS (55\%)} & {DBS (50\%)} & {DBS (45\%)}\\
\cmidrule(lr){1-8}
airfoil & 16.6198 & 16.5061= & {15.0541}+ & {15.051}+ & {14.8581}+ & {15.7739}+ & {15.7069}+ \\
heating & 7.929 & {7.3309}+ & {7.303}+ & {7.2609}+ & {7.5126}+ & 8.5253- & {7.7708}+ \\
cooling & 7.2973 & {6.8732}+ & {6.787}+ & {6.7121}+ & {6.6915}+ & {6.8563}+ & {6.9468}+ \\
concrete & 14.239 &	14.233= & {13.769}+ & {13.954}+ & {14.107}+ & {13.586}+ & {13.497}+\\
redwine & 0.769 & 0.7677= & {0.7647}+ & {0.7669}+ & {0.7651}+ & 0.7698- & 0.7708-\\
whitlewine & 0.8246 & {0.8137}+ & {0.8133}+ & {0.8162}+ & 0.8181= & {0.8126}+ & 0.8164=\\ 
housing & 11.1763 & {10.8883}+ & {10.9344}+ & {11.1181}+ & 11.0832= & 11.0703= & {10.9704}+\\
pollution & 102.291	& 102.0317=	& {95.3476}+ & 95.2581= & 99.7087=	& 98.7799=	& 98.1966= \\
dowchem & 0.3392 & 0.3414- & 0.3398= & 0.3396= & 0.3401- & {0.3386}+ & {0.338}+ \\
crime & 0.1543 & {0.1522}+ & {0.1519}+ & {0.153}+ & {0.1536}+ & {0.1537}+ & {0.1522}+\\
\cmidrule(lr){1-8}
Total +/=/-   & -- & 5/4/1 & 9/1/0 & 8/2/0 & 6/3/1  & 6/2/2 & 7/2/1 \\
\cmidrule(lr){1-8}
\end{tabular}}
\end{table}

We discover that the DBS algorithm performs better in contrast to how well it performed with synthetic SR problems because real-world SR problems present a higher number of features and level of complexity than synthetic SR benchmarks. While examining the findings for each benchmark, DBS surpassed the baseline at one or more test case selection budgets while achieving competitive performance at other budgets.

Table \ref{tab:CircuitWilcoxonTest} displays the findings of the Wilcoxon rank-sum test conducted on the test score for digital circuits on different DBS training budgets against the baseline. We followed the same strategy for the statistical test as used in the case of \gls{sr}. It is evident from Table \ref{tab:SynWilcoxonTest}-\ref{tab:CircuitWilcoxonTest} that the proposed DBS algorithm helps select a subset of test cases to train models that perform at least as well as—and sometimes significantly better, in terms of achieving better test score, than the considered baseline.

\begin{table}[!h]
    \centering
    \caption{Results of the Wilcoxon test for \textbf{digital circuit} benchmarks considering a significance level of $\alpha = 0.05$. Values shown are the mean best test scores across 30 independent runs. The symbols +, =, - indicate whether the corresponding results for DBS are significantly better, not significantly different, or worse than the baseline. The last row summarizes this information.}
    \label{tab:CircuitWilcoxonTest}
    \resizebox{\columnwidth}{!}{
    \begin{tabular}{l c c c c c c r}
    \cmidrule(lr){1-8}
{Benchmarks} & {Baseline} & {DBS (70\%)} & {DBS (65\%)} & {DBS (60\%)} & {DBS (55\%)} & {DBS (50\%)} & {DBS (45\%)}\\
    \cmidrule(lr){1-8}
Comparator   & 1024     & 1024=     & 1024=     & 1024=     & 1024=     & 1024=     & 1024=\\
Parity & 26.334   & 25.78067- & 26.756+   & 25.816-   & 26.46667= & 26.07267= & 26.18533=\\
MUX    & 1930.923 & 1922.645- & {1927.595}= & {1928.789}= & 1922.048- & 1919.659- & 1919.829-\\
ALU    & 3250.112 & 3212.179- & {3255.58}=  & 3226.407- & 3166.453- & 3229.893- & {3260.591}+\\
\cmidrule(lr){1-8}
Total +/=/-   & -- & 0/1/3 & 1/3/0 & 0/2/2 & 0/2/2 &	0/2/2 &	1/2/1\\
\cmidrule(lr){1-8}
\end{tabular}}
\end{table}

Although $B$ depends on several factors, including the difficulty of the problem, resource constraints, and the type of data being utilized, the results in Table \ref{tab:statTest_realWorld} show that DBS (65\%) and DBS (60\%) consistently outperformed or matched the baseline across all problems. Likewise, in Table \ref{tab:CircuitWilcoxonTest}, DBS (65\%) never underperformed the baseline on any of the benchmarks. This can be a promising starting point when applying for DBS. Nevertheless, users can define the budget based on their requirements and available resources.

\subsection{Time Analysis}
\label{sec:TimeAnalysis}
It is crucial to measure the time efficiency and, hence, the computational cost of the method. To compare the GE run time, we performed a time analysis for all the problems investigated in this paper; where appropriate, we also present the time taken by the DBS algorithm in selecting the subset of test cases.

\begin{figure}[!h]
    \centering
    \subfigure[Keijzer-4\label{fig:Time_Keijzer4}]{\includegraphics[width=0.32\textwidth]{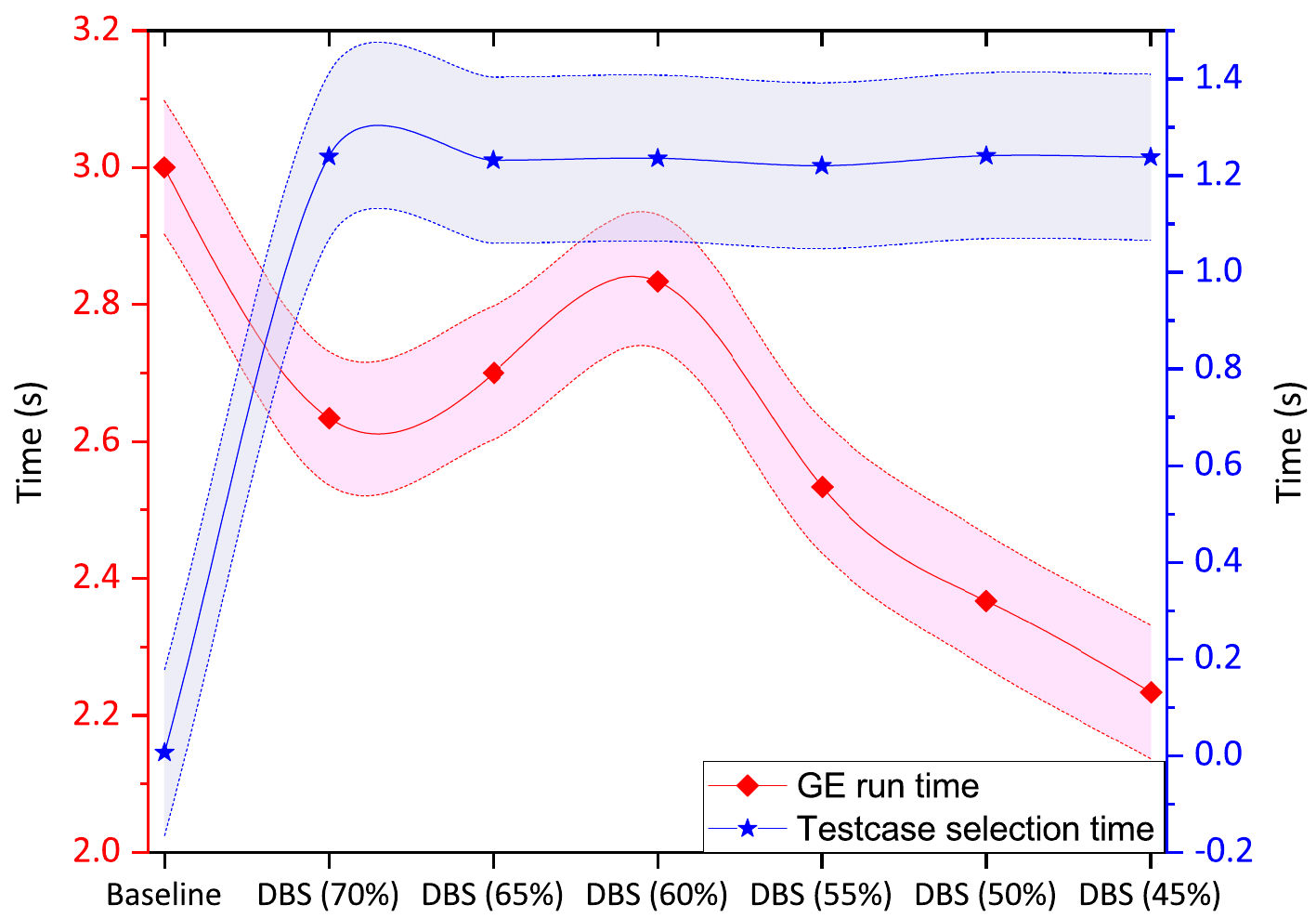}}
    \subfigure[Keijzer-9\label{fig:Time_Keijzer9}]{\includegraphics[width=0.32\textwidth]{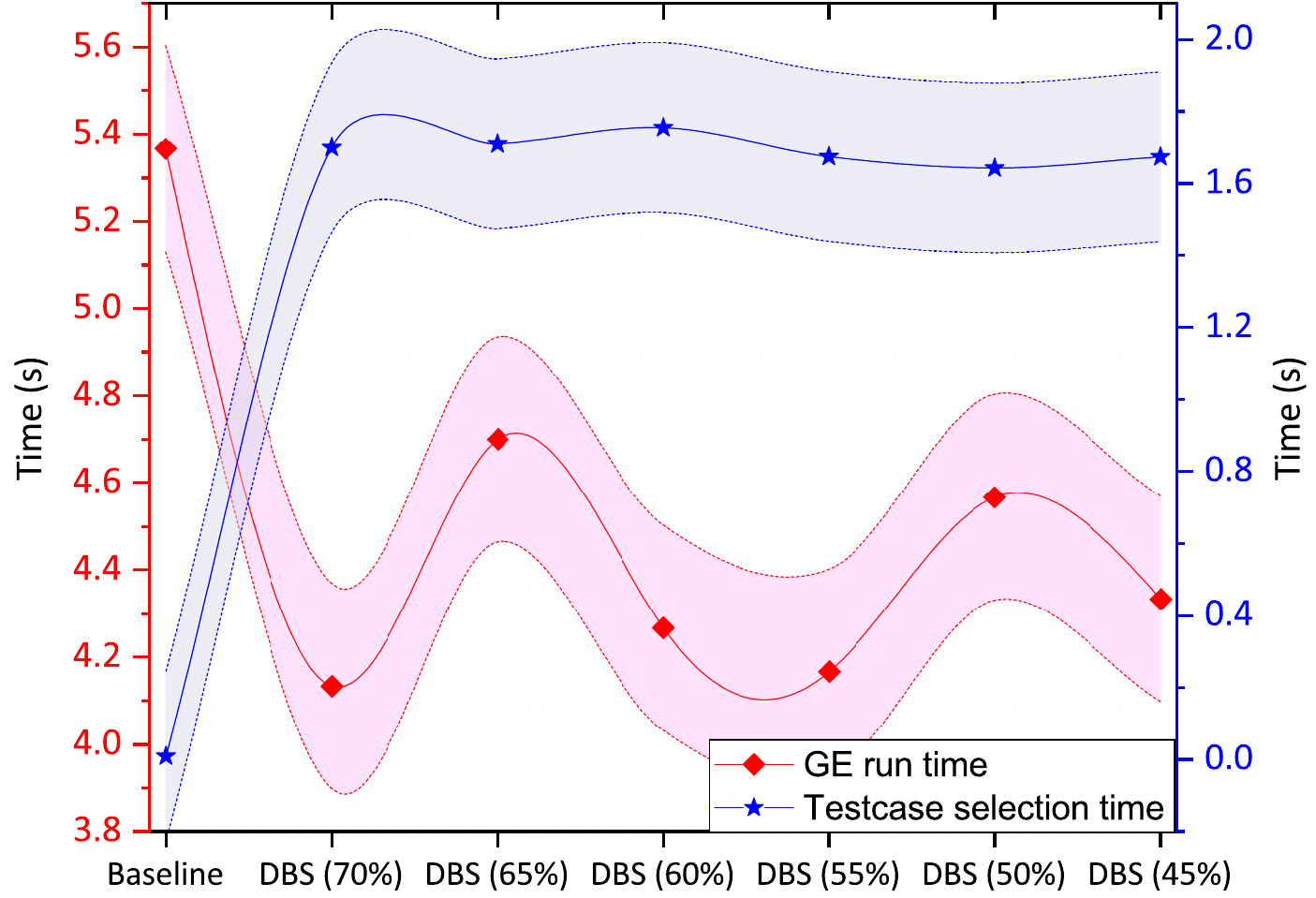}}
   \subfigure[Keijzer-10\label{fig:Time_Keijzer10}]{\includegraphics[width=0.32\textwidth]{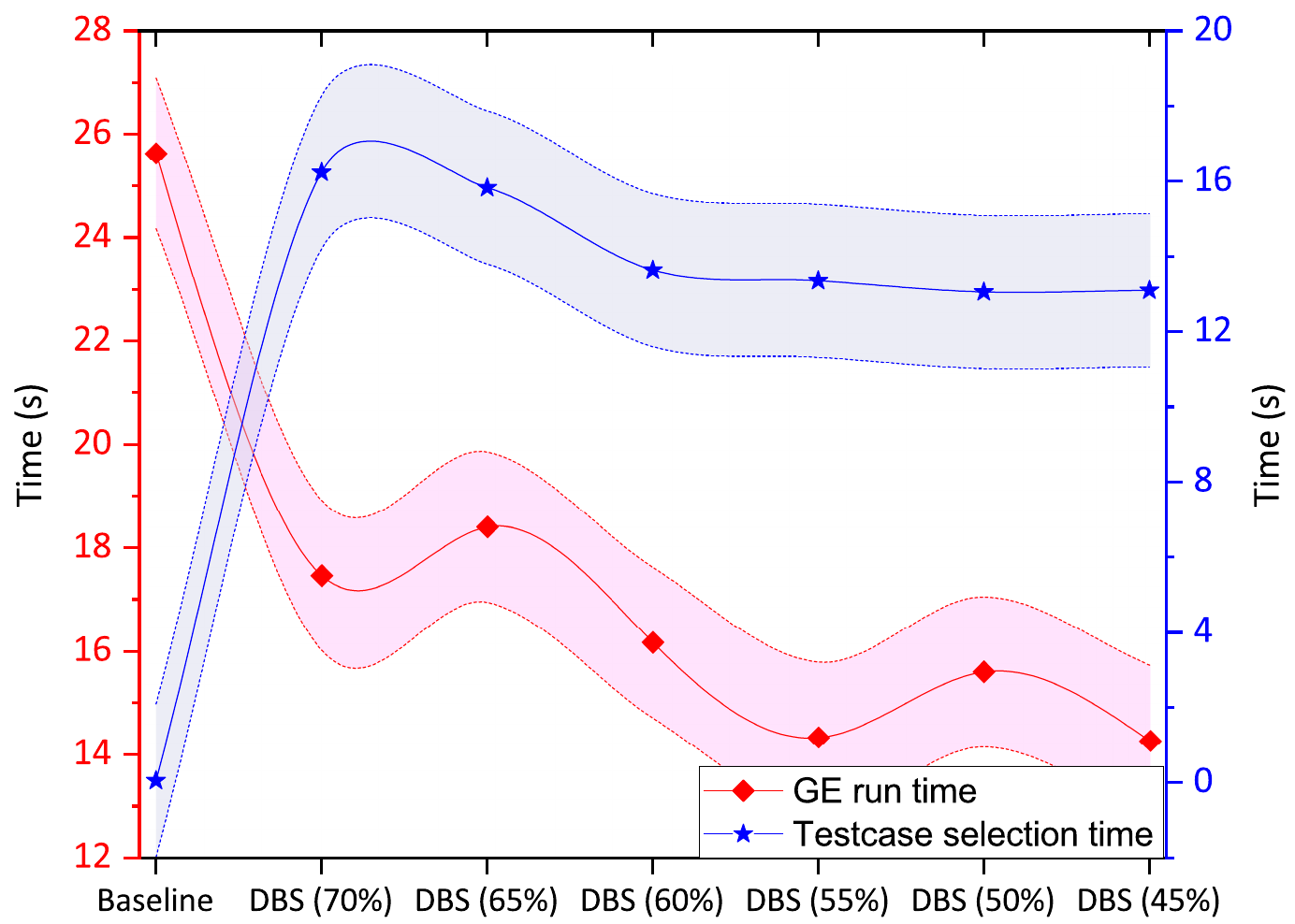}}\\
    \subfigure[Keijzer-14\label{fig:Time_Keijzer14}]{\includegraphics[width=0.32\textwidth]{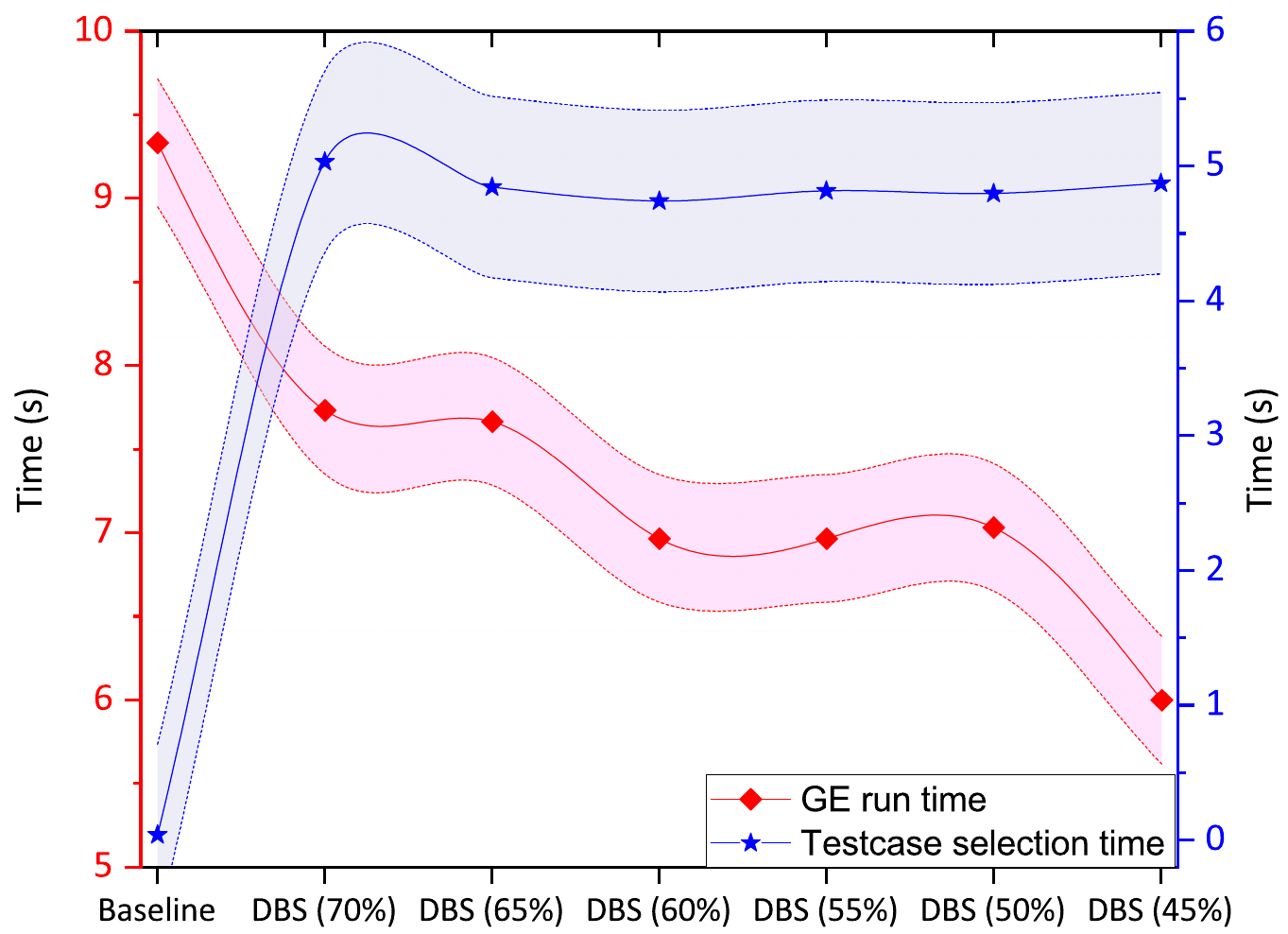}}
    \subfigure[Nguyen-9\label{fig:Time_Nguyen9}]{\includegraphics[width=0.32\textwidth]{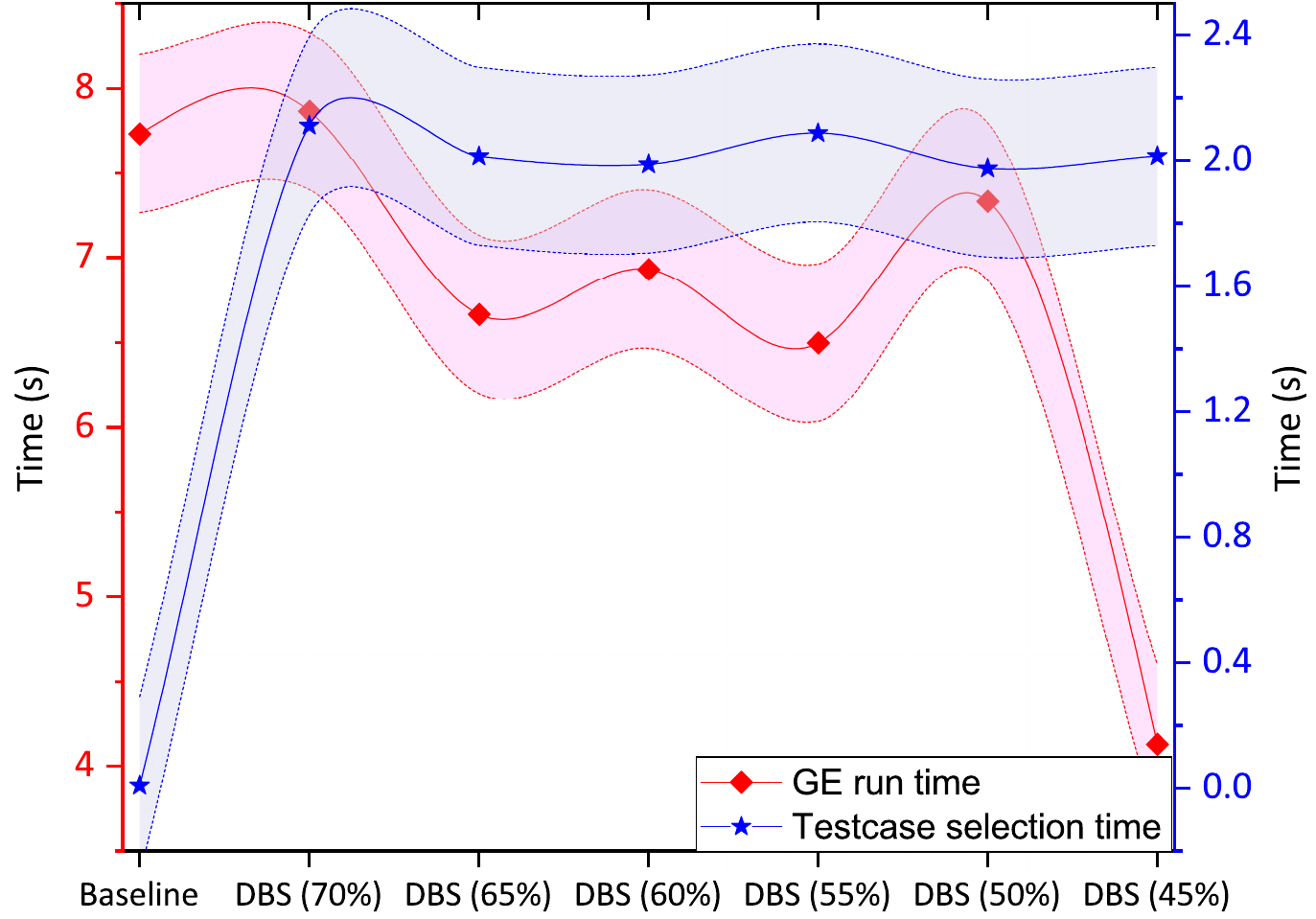}}
    \subfigure[Nguyen-10\label{fig:Time_Nguyen10}]{\includegraphics[width=0.32\textwidth]{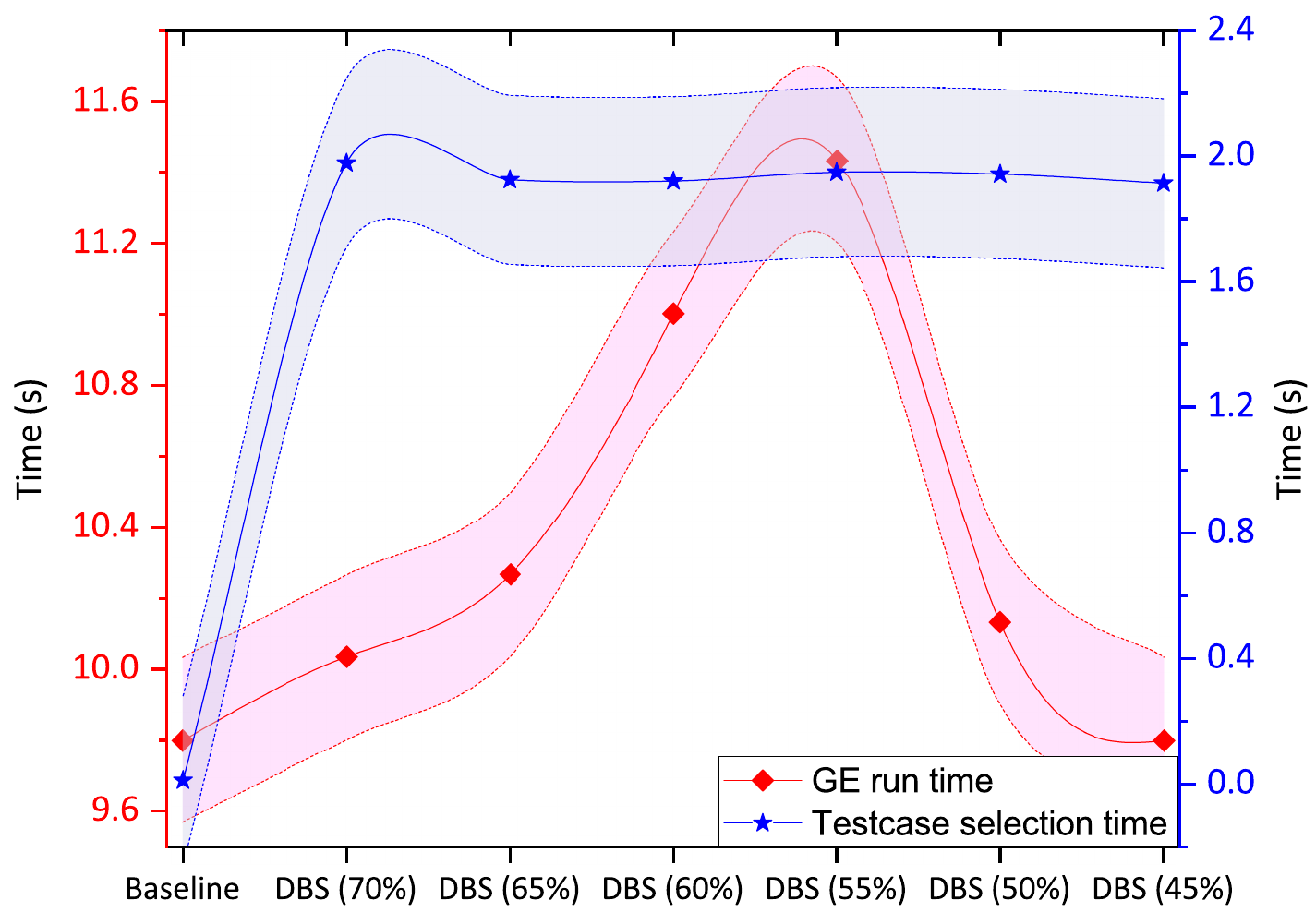}}\\
     \subfigure[Keijzer-5\label{fig:Time_Keijzer5}]{\includegraphics[width=0.32\textwidth]{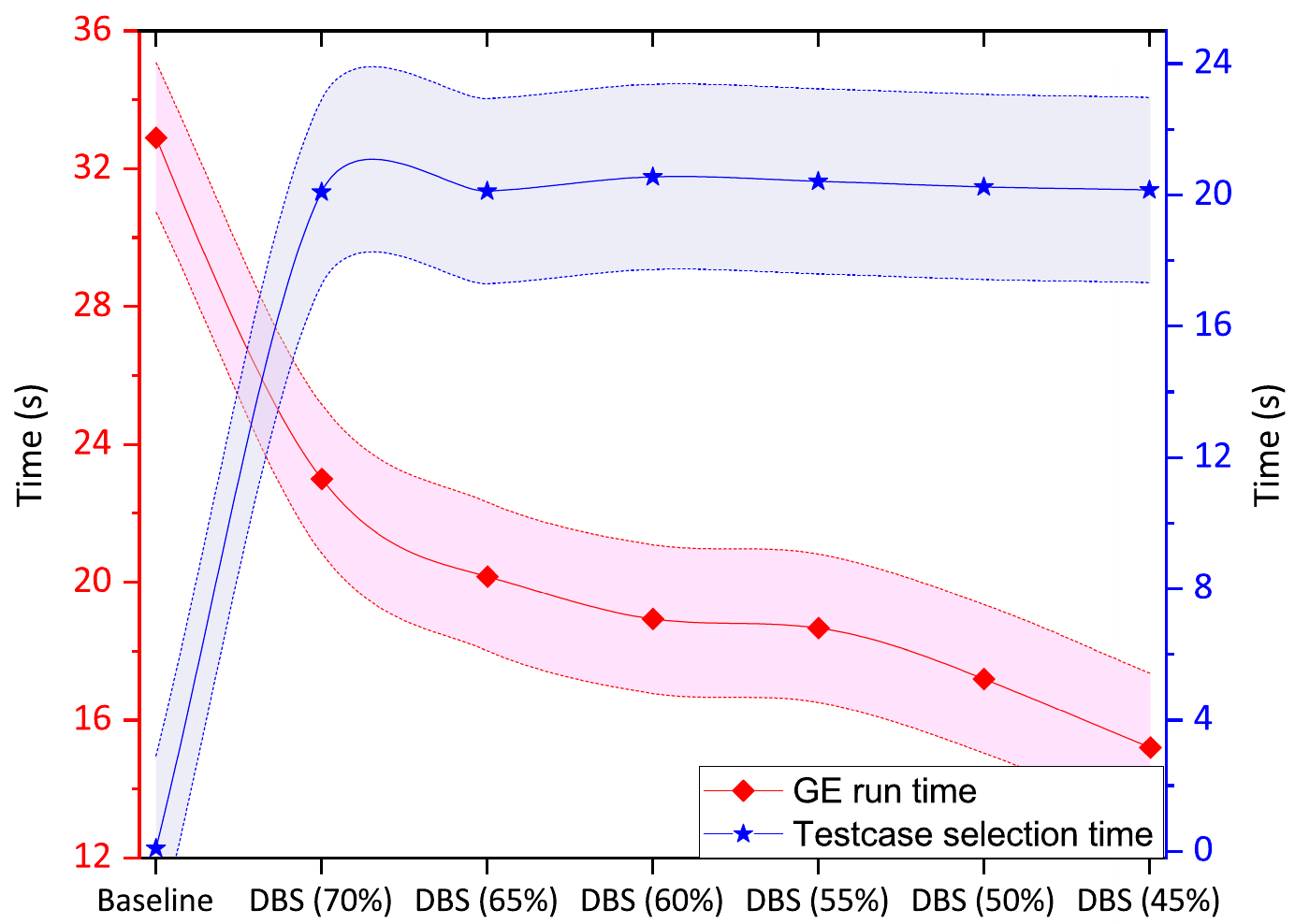}}
    \subfigure[Vladislavleva-5\label{fig:Time_Vlad5}]{\includegraphics[width=0.32\textwidth]{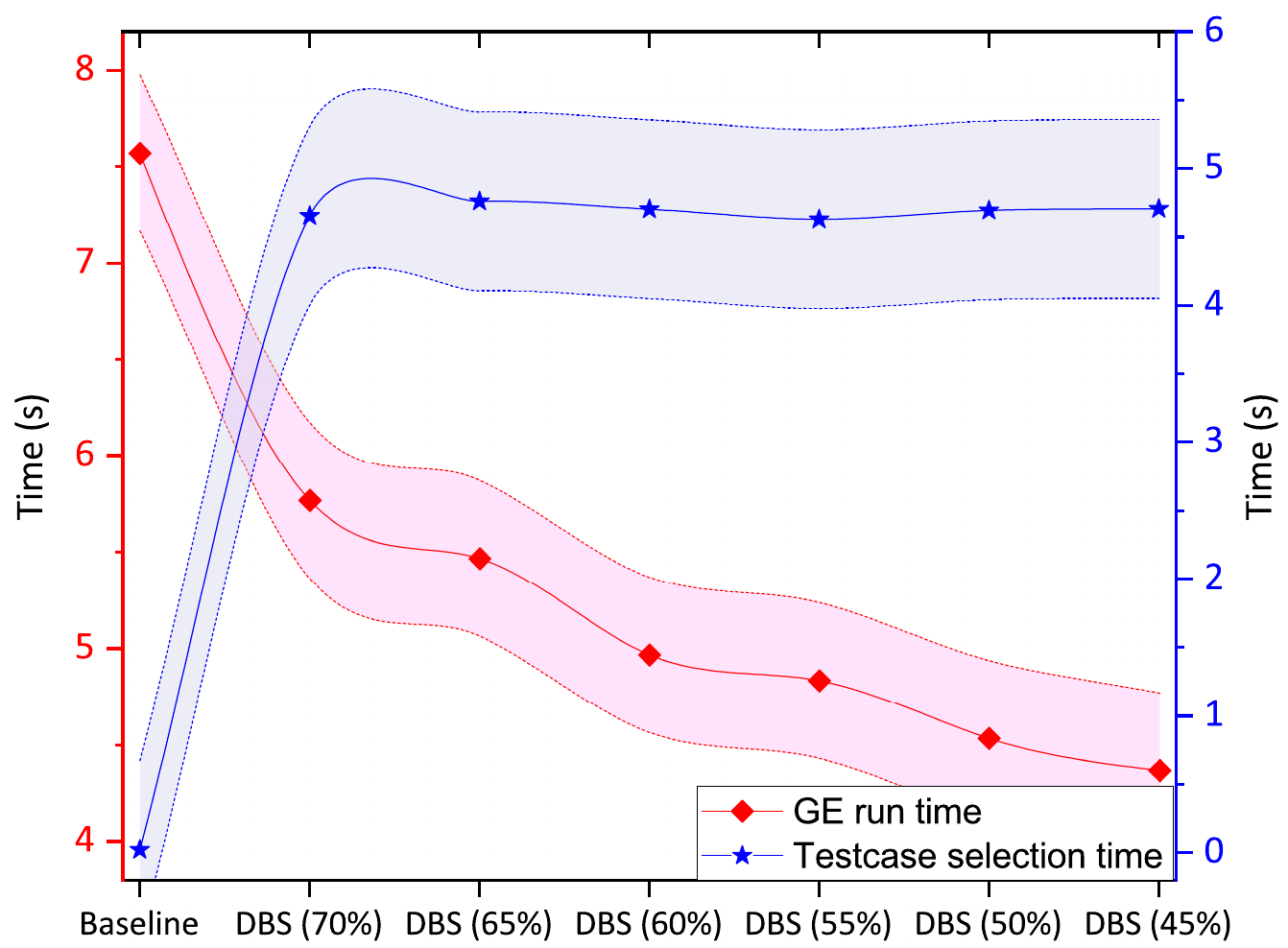}}
    \subfigure[Korns-11\label{fig:Time_korns11}]{\includegraphics[width=0.32\textwidth]{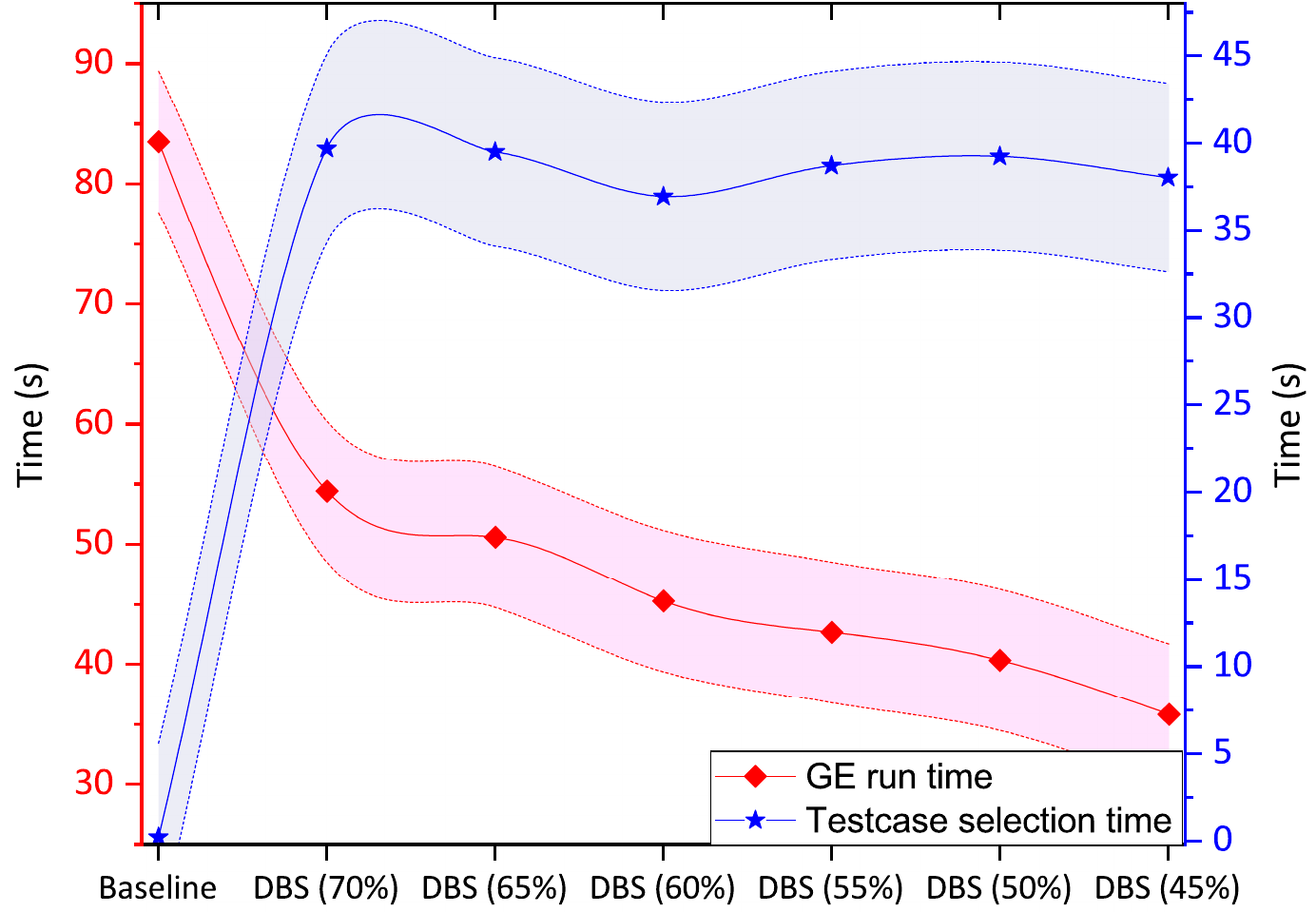}}\\
    \subfigure[Korns-12\label{fig:Time_korns12}]{\includegraphics[width=0.32\textwidth]{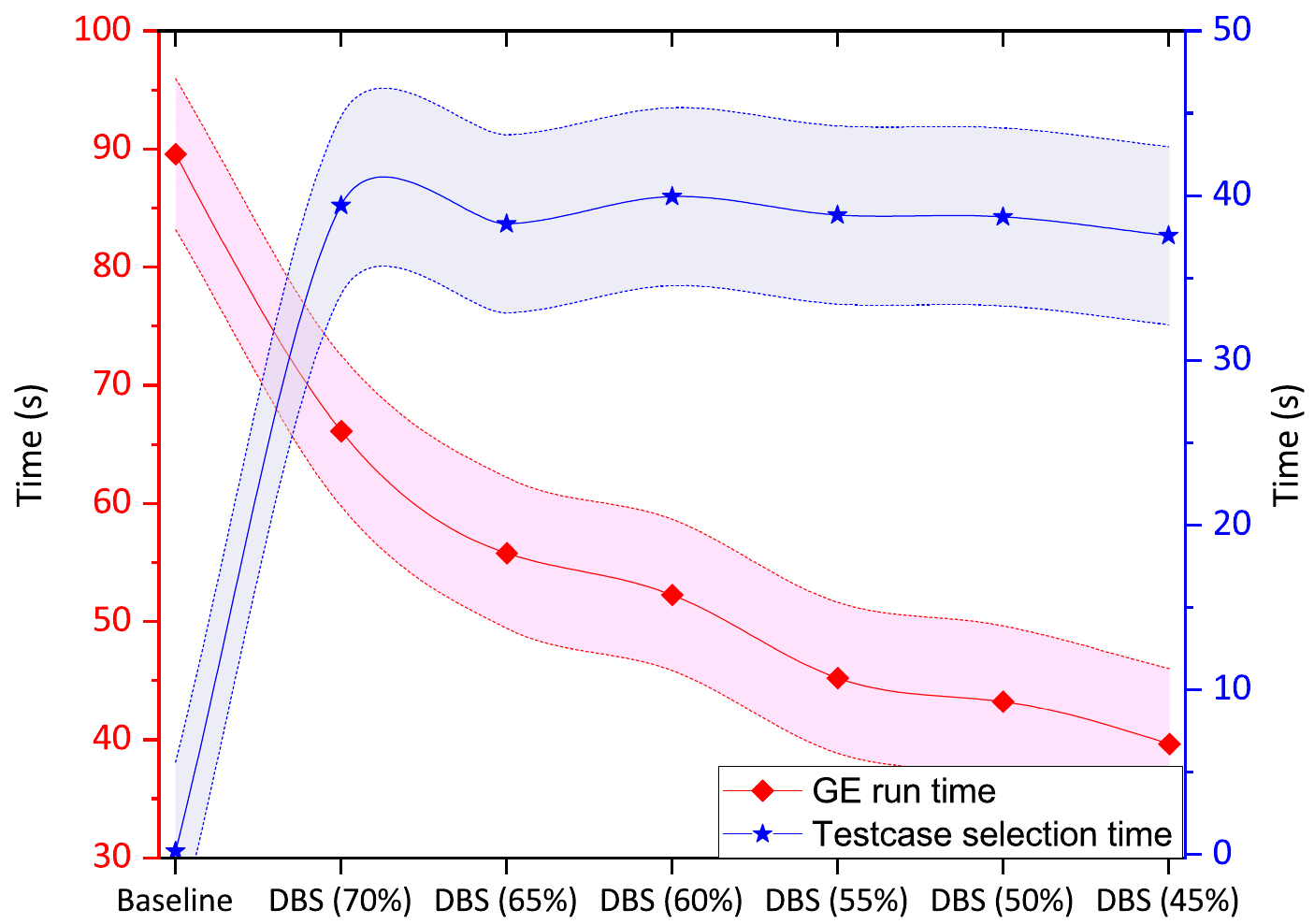}}
    \caption{Total time taken in test case selection and average time taken per GE run along with std error for synthetic \gls{sr} benchmarks.}
    \label{fig:SynTime}
\end{figure}

The graphs in Figure \ref{fig:SynTime} give the results of the time analysis for synthetic SR benchmarks. In all  10 of those benchmarks, the maximum test selection time by DBS is at the budget DBS (70\%) and lower. A decrease in $B$ decreases the test case selection time. Of course, the time taken in preparing baseline data is lower (see the blue line in Figure \ref{fig:SynTime}) as its simple task was to split the dataset into the ratio of 70\%-30\%.

To measure the time cost for each GE run, we calculate the total training time taken across 30 independent runs and take an average of that. A time analysis graph for \texttt{Keijzer-4} is shown in Figure \ref{fig:Time_Keijzer4}. The shaded region in the line graph shows standard error. Using DBS training data, the model training took less time than the baseline experiment did. However, the expected time consumption using DBS (60\%) should have been lower than that at DBS (70\%). If we look closely at the numerical data, the time difference between the two DBS instances is in milliseconds, which is too small to track. The alternate strategy might involve performing tens of thousands of runs to obtain a better value; however, this is not practical owing to time restrictions. The depiction of the \texttt{Keijzer-9} instance in Figure \ref{fig:Time_Keijzer9} illustrates a similar finding to \texttt{Keijzer-4}. The training time for a DBS with a lesser budget is likewise anticipated to be shorter than one with a higher budget. The time difference is too small, and that is because the overall run time is not long enough to distinguish the difference. In the case of \texttt{Keijzer-10}, we see a distinct pattern, and a possible reason is that the benchmark poses a higher number of instances that increase the total run time, as shown in Figure \ref{fig:Time_Keijzer10}. A similar trend is observed for \texttt{Keijzer-14} in Figure \ref{fig:Time_Keijzer14}. In the case of \texttt{Nguyen-9} in Figure \ref{fig:Time_Nguyen9}, the training duration in baseline and DBS (70\%) is comparable; however, in other cases, we see a pattern similar to \texttt{Keijzer-9} (Figure \ref{fig:Time_Keijzer9}). 
One potential explanation is the total number of instances in both problems is $\approx$ 1k. However, in most instances, the DBS training period is shorter than the baseline training time. For \texttt{Nguyen-10}, Figure 6 shows a completely different pattern. Even though DBS (55\%) has fewer instances than the baseline, the training period is longer. In order to determine the potential cause of such instances where smaller training data takes longer to process than bigger training data, we made the decision to examine the effective solution size, which is further covered in Section \ref{subsec:IndSize}. The training times for \texttt{Keijzer-5}, \texttt{Vladislavleva-5}, \texttt{Korns-11}, and \texttt{Korns-12} problems are evident and as expected. This is primarily due to the longer training period and/or increased problem complexity.
Briefly, the baseline run time for all the algorithms is significantly higher, as observed in Figure \ref{fig:SynTime}. For example, in the case of \texttt{Keijzer-10}, the worst-case scenario of DBS (70\%) is 9 seconds lower than the baseline.

\begin{figure}[t!]
    \centering
    \subfigure[Airfoil\label{fig:Time_Airfoil}]{\includegraphics[width=0.32\textwidth]{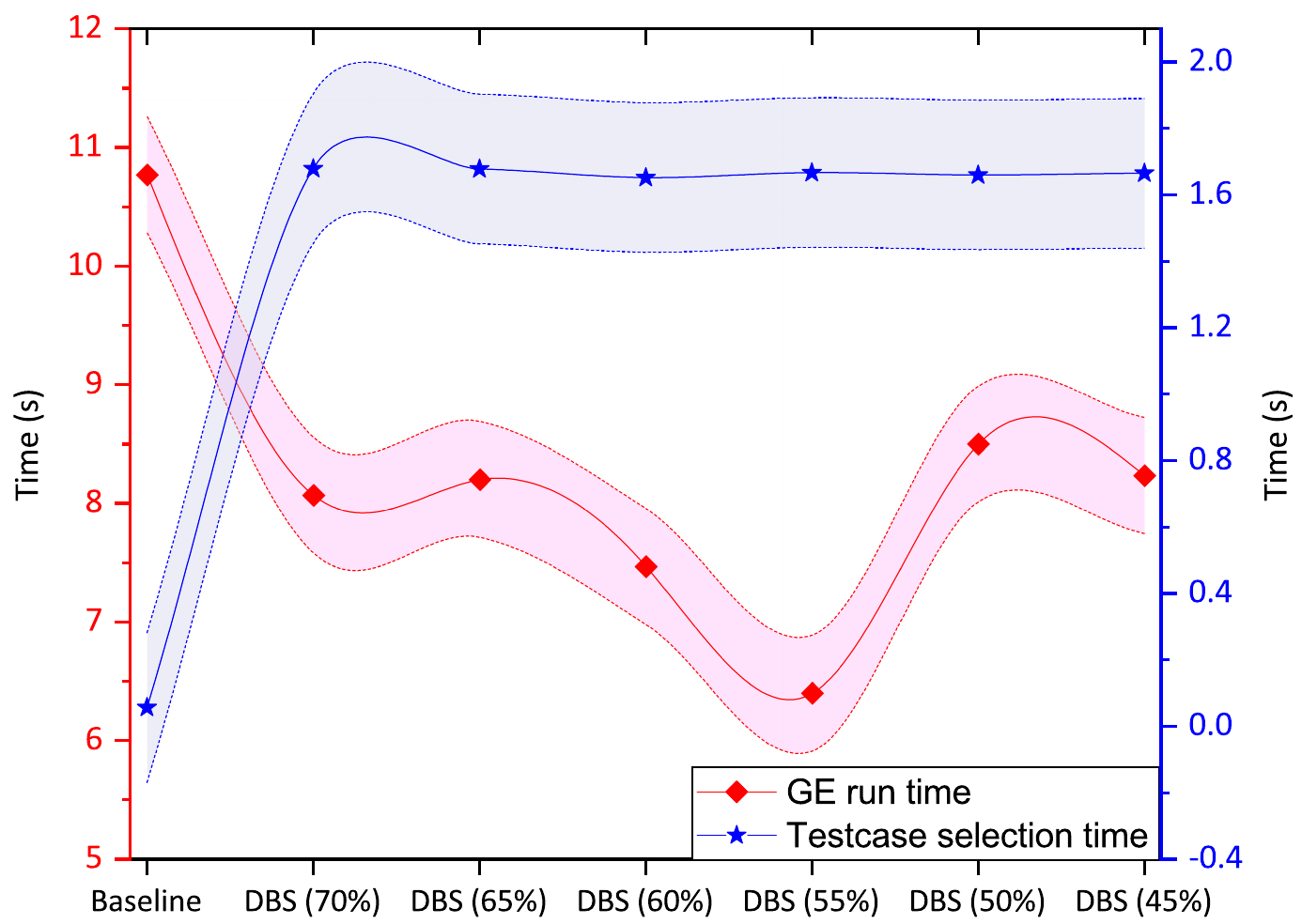}}
    \subfigure[Heating\label{fig:Time_Heating}]{\includegraphics[width=0.32\textwidth]{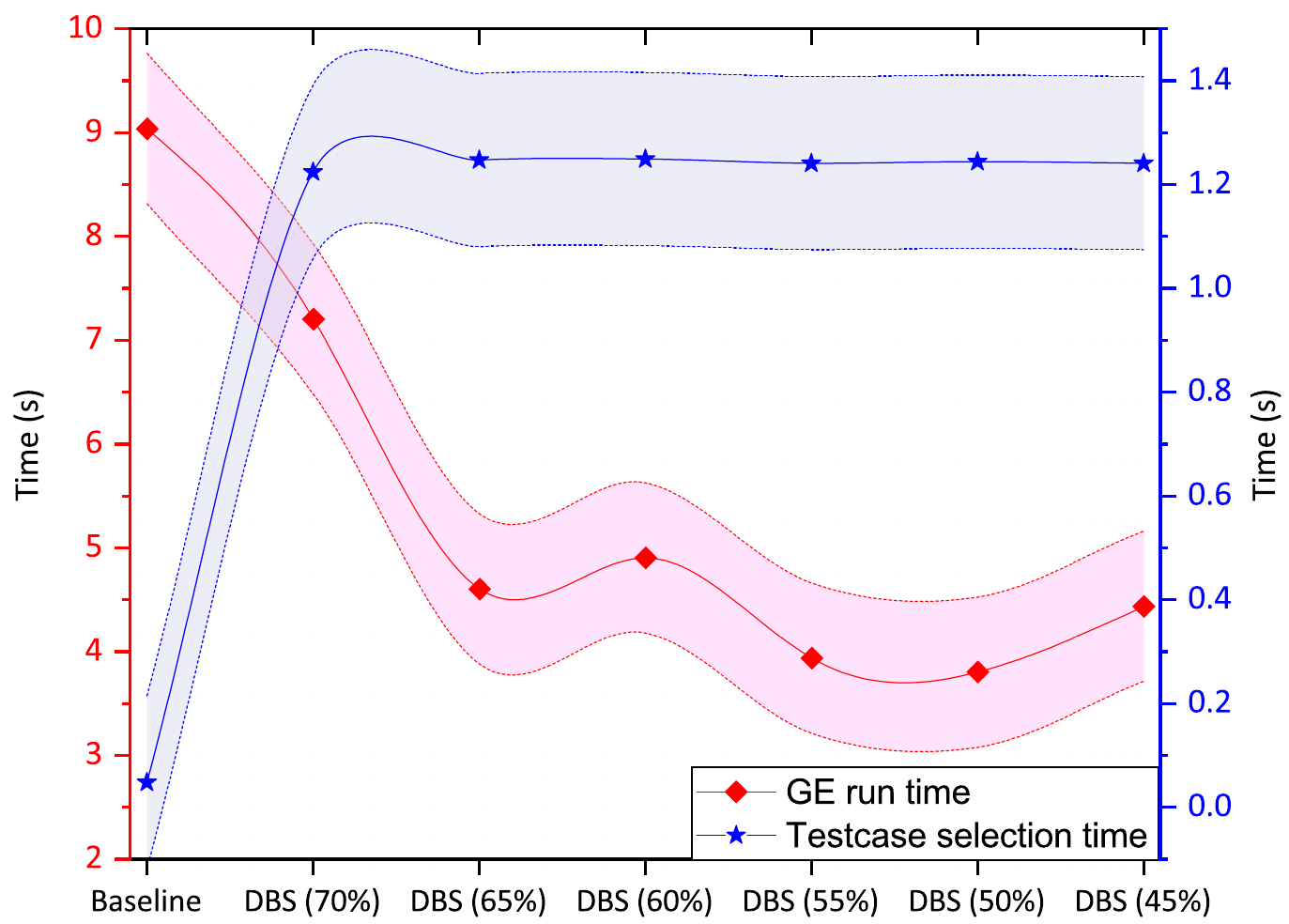}}
   \subfigure[Cooling\label{fig:Time_Cooling}]{\includegraphics[width=0.32\textwidth]{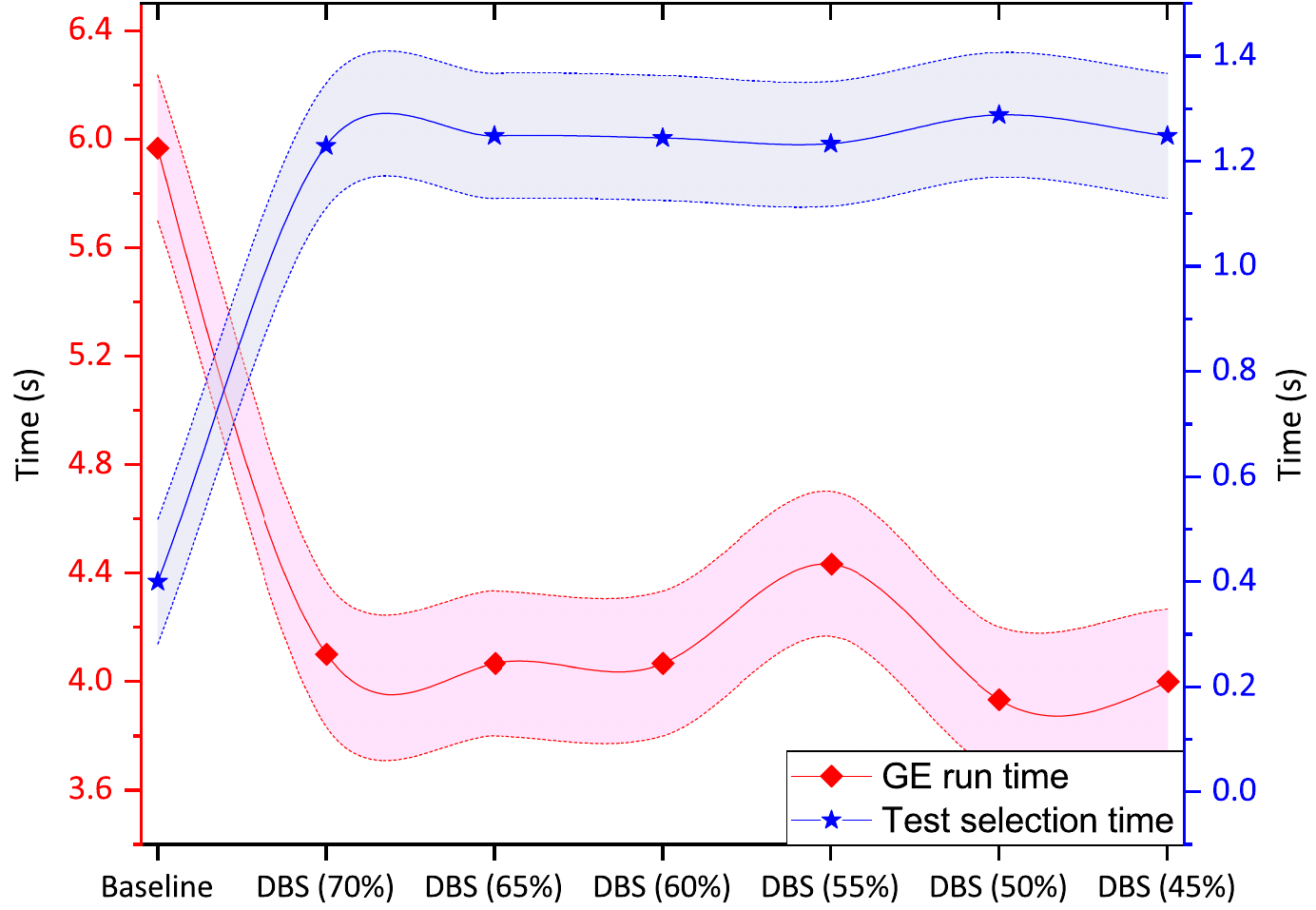}}\\
    \subfigure[Concrete\label{fig:Time_Concrete}]{\includegraphics[width=0.32\textwidth]{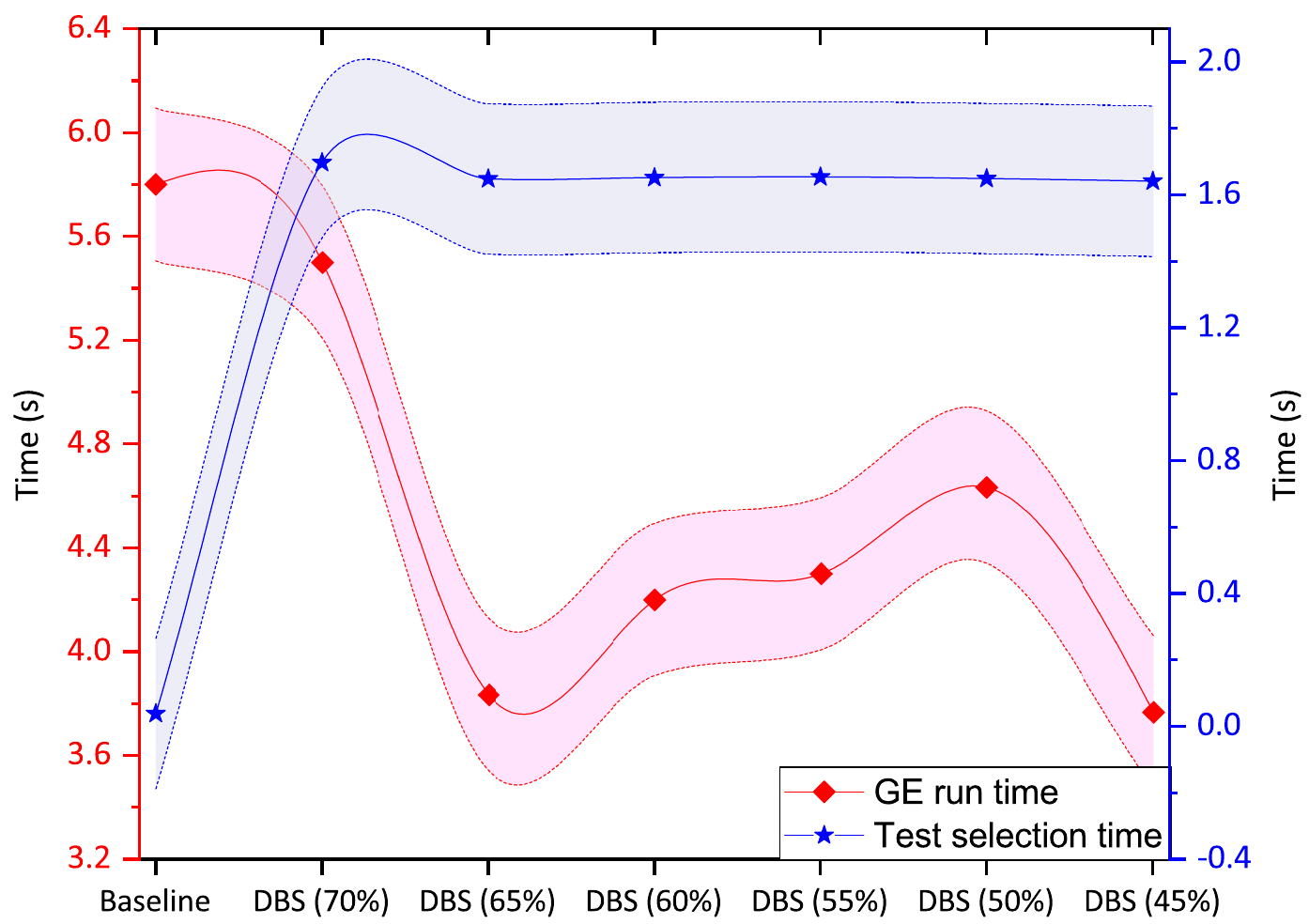}}
    \subfigure[Redwine\label{fig:Time_Redwine}]{\includegraphics[width=0.32\textwidth]{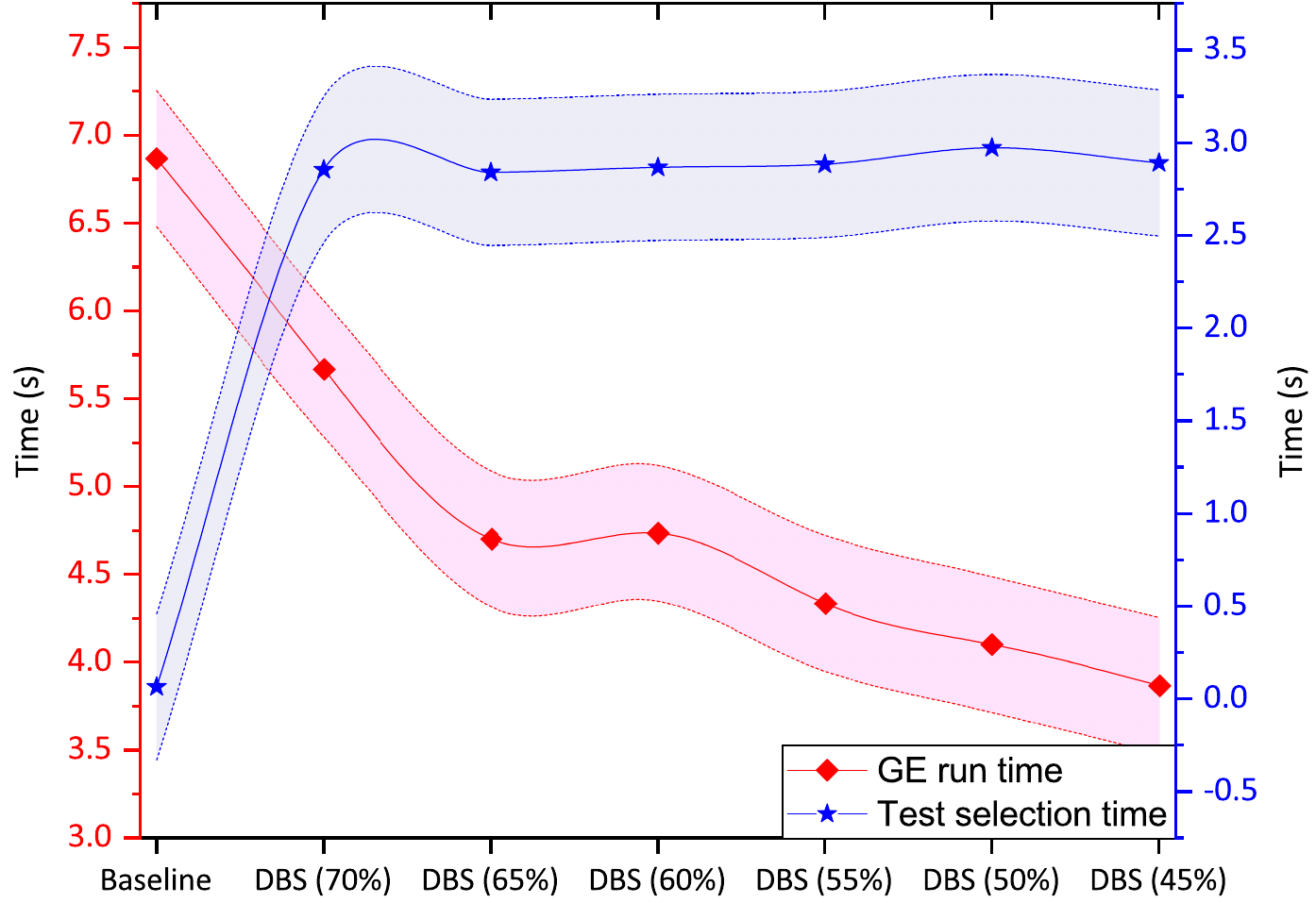}}
    \subfigure[Whitewine\label{fig:Time_Whitewine}]{\includegraphics[width=0.32\textwidth]{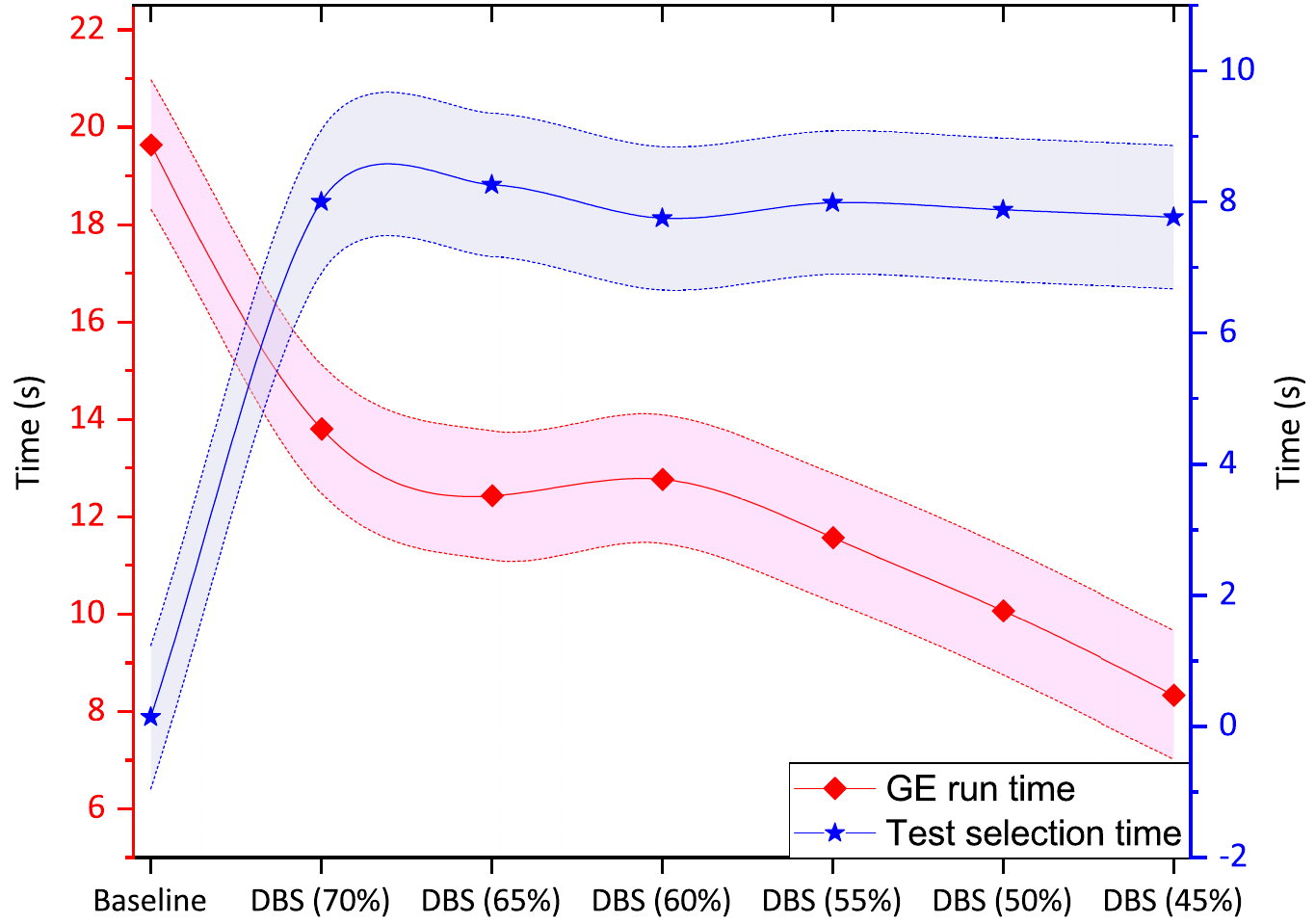}}\\
     \subfigure[Housing\label{fig:Time_Housing}]{\includegraphics[width=0.32\textwidth]{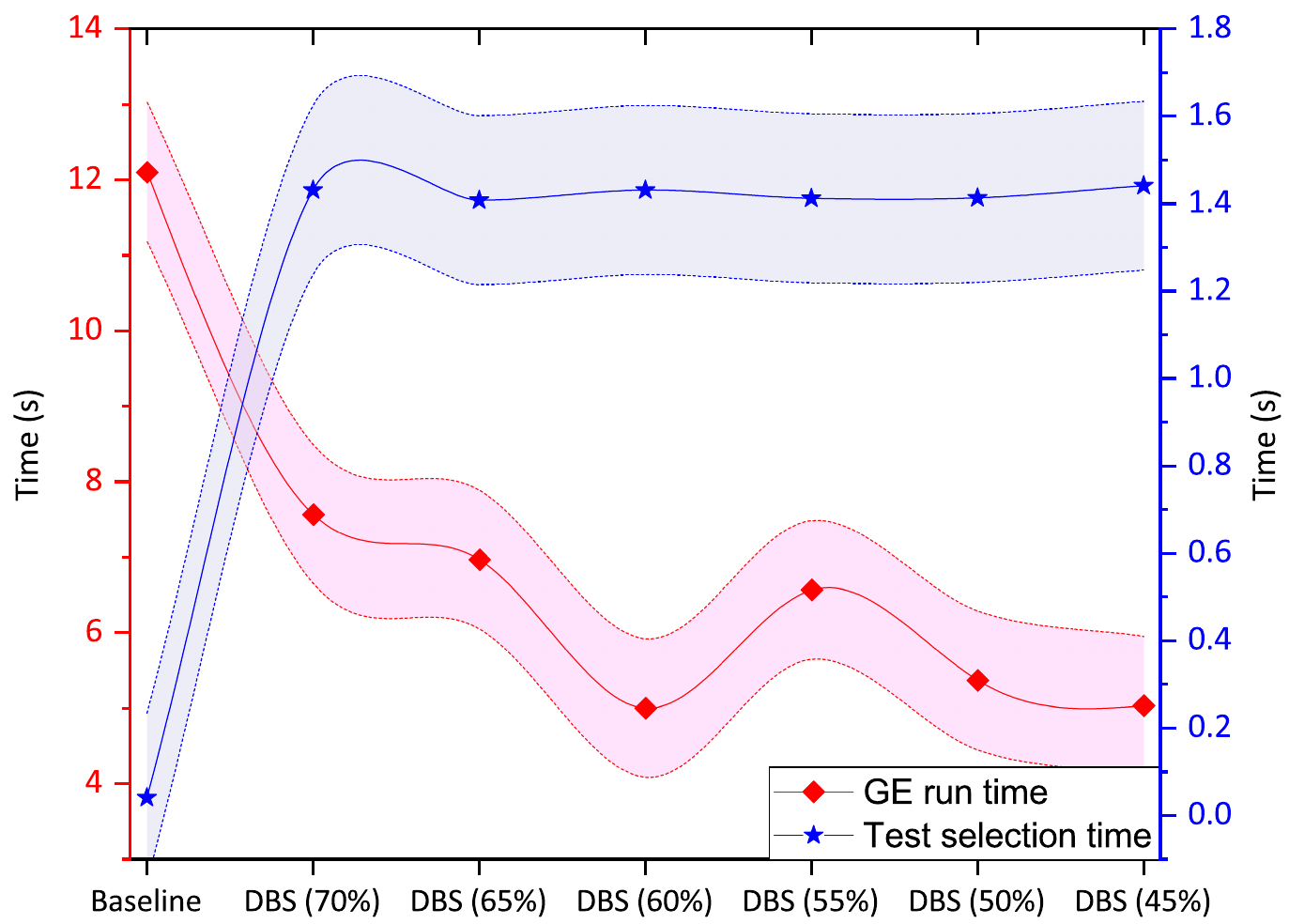}}
    \subfigure[Pollution\label{fig:Time_Pollution}]{\includegraphics[width=0.32\textwidth]{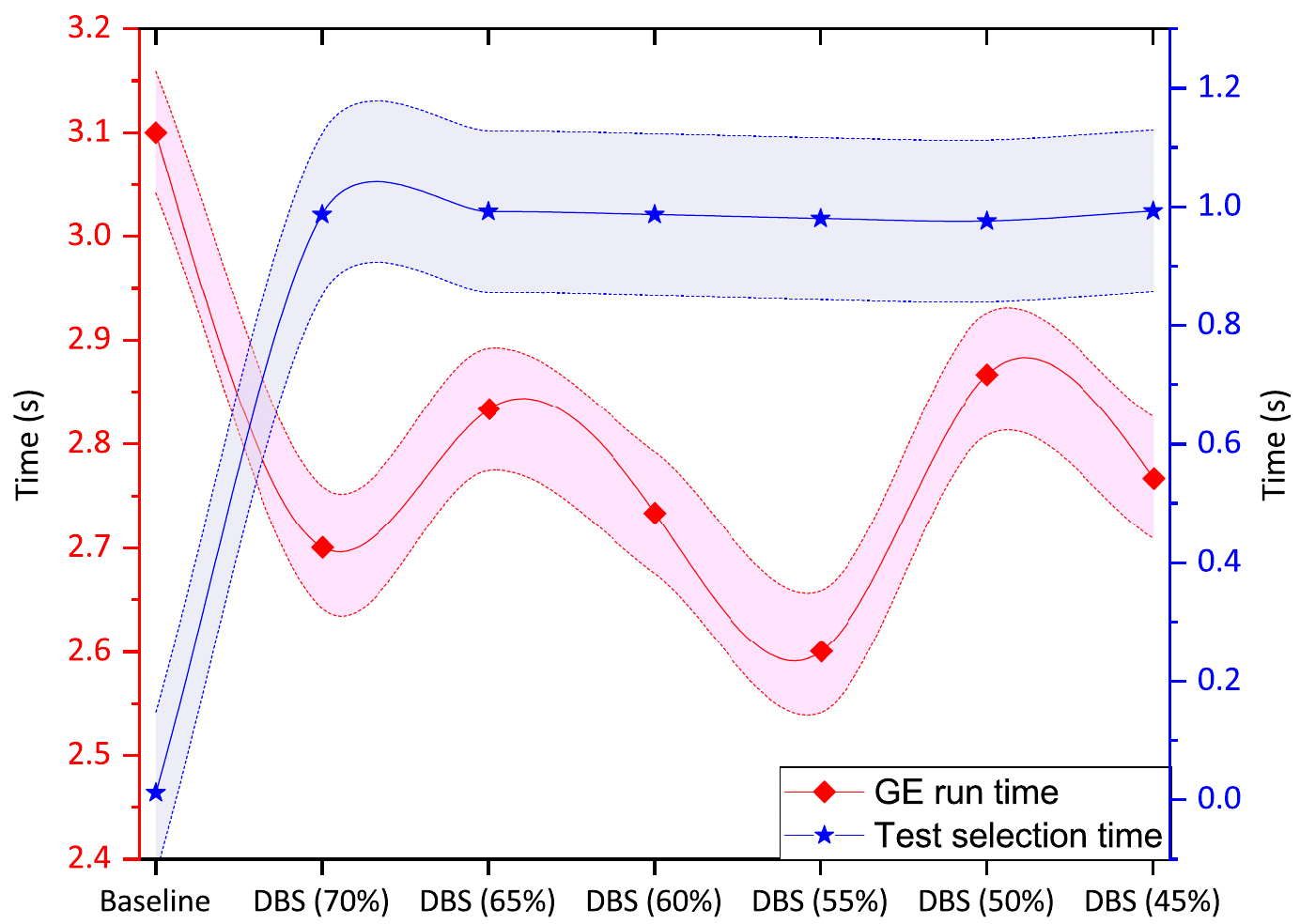}}
    \subfigure[Dowchem\label{fig:Time_Dowchem}]{\includegraphics[width=0.32\textwidth]{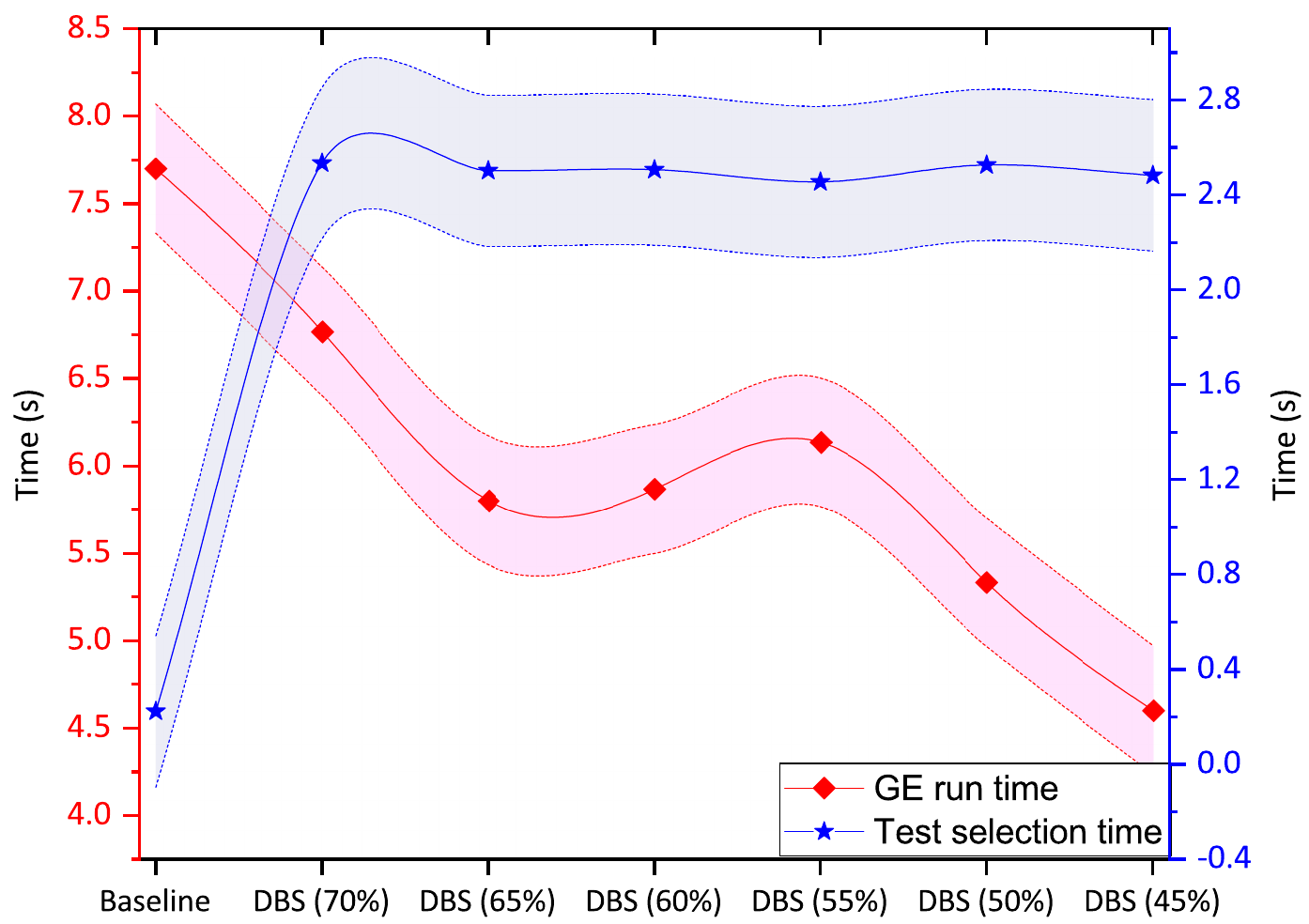}}\\
    \subfigure[Crime\label{fig:Time_Crime}]{\includegraphics[width=0.32\textwidth]{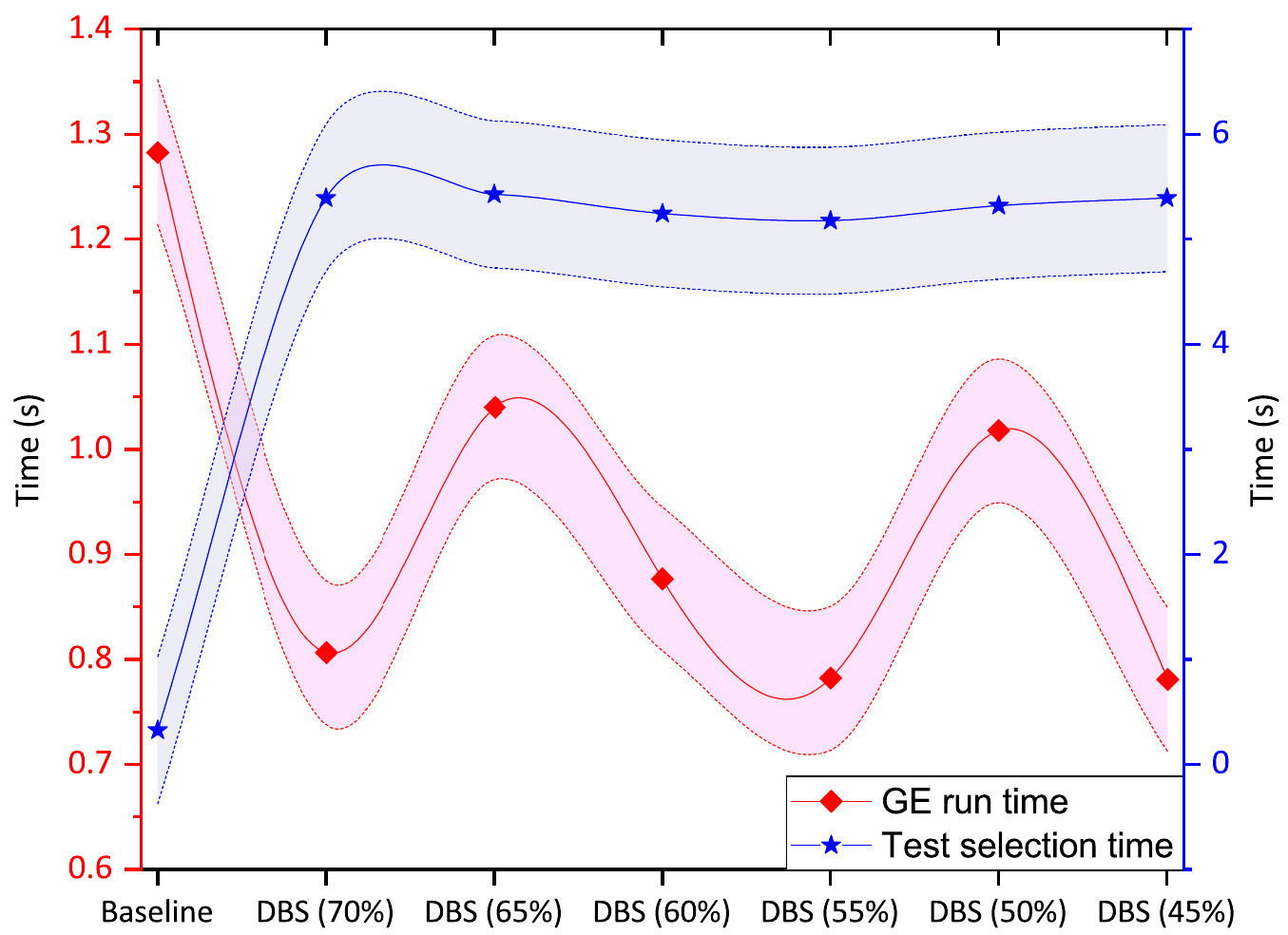}}
    \caption{Total time taken in test case selection and average time taken per GE run along with std error for real-world \gls{sr} benchmarks.}
    \label{fig:RealTime}
\end{figure}

The GE run time or training time analysis of real-world SR experiments is shown in Figure \ref{fig:RealTime}. In comparison to synthetic SR datasets, these benchmarks have more instances. The complexity of the problem has increased as well. The training time in most cases is typically predictable (in terms of graph trend) and, in some ways, evident. The blue line displays the test case selection time for the baseline and DBS approaches. The baseline strategy typically takes less time to choose test cases than the DBS approach as it does not involve any complex steps as in DBS. The DBS took nearly the same or less time on most test case selection budgets.

Figure \ref{fig:Time_Airfoil} depicts the amount of training time required for the \texttt{airfoil} experiment. The DBS training set significantly reduces the GE run time compared to the baseline training period. However, the time at DBS (50\%) is longer than at DBS (55\%). We hypothesize that this may be because the experiment employing DBS (50\%) produced greater individual (solution) sizes. We do want to conduct an analysis of solution size to further study it, even if our main objective is to compare the time efficiency of DBS with the baseline, which is solved in the majority of cases. The time analysis in Figure \ref{fig:Time_Heating} for the \texttt{heating} dataset is as expected. Although using DBS training data at some budgets appears to take a bit longer than their neighboring higher budget, the time difference is relatively small, which is another instance where having thousands of runs is necessary to identify a clear trend. Figures \ref{fig:Time_Cooling} and \ref{fig:Time_Concrete} show a pattern that is similar. Given that these problems have an equal amount of features and instances, it is conceivable that this is why identical patterns in \texttt{heating}, \texttt{cooling}, and \texttt{concrete} run-time are observed. As can be seen in Figures \ref{fig:Time_Redwine}-\ref{fig:Time_Whitewine}, and \ref{fig:Time_Housing}, the training times for the \texttt{redwine}, \texttt{whitewine}, and \texttt{housing} experiments are as anticipated. A shorter run time in these experiments is accomplished since the fitness evaluation is lower with smaller training data using DBS. The strong trend is evident, and a large number of instances and features in the dataset may be a contributing factor. Figure \ref{fig:Time_Pollution} of the \texttt{pollution} experiment shows a pattern for training time that is comparable to that shown in Figure \ref{fig:Time_Concrete}. This may be related to solution size or the fact that the number of instances is low. Figure \ref{fig:Time_Dowchem} displays the \gls{ge} run time for \texttt{dowchem} experiment. In the case of \texttt{crime} experiments, as in Figure \ref{fig:Time_Crime}, the GE run time of experiments using different DBS budgets varies arbitrarily by small fractions. This results from the run duration being so short that managing such minute variations is challenging. 

The DBS training data, in fact, required less time to train the model in most instances, as can be seen by looking at most of the graphs presented above. This study indicates that training these models on training data selected using DBS is quicker and yields results comparable to or better than those produced in the baseline experiments.

\begin{figure}[!h]
    \centering
    \subfigure[Comparator\label{fig:Time_Comp}]{\includegraphics[width=0.4\textwidth]{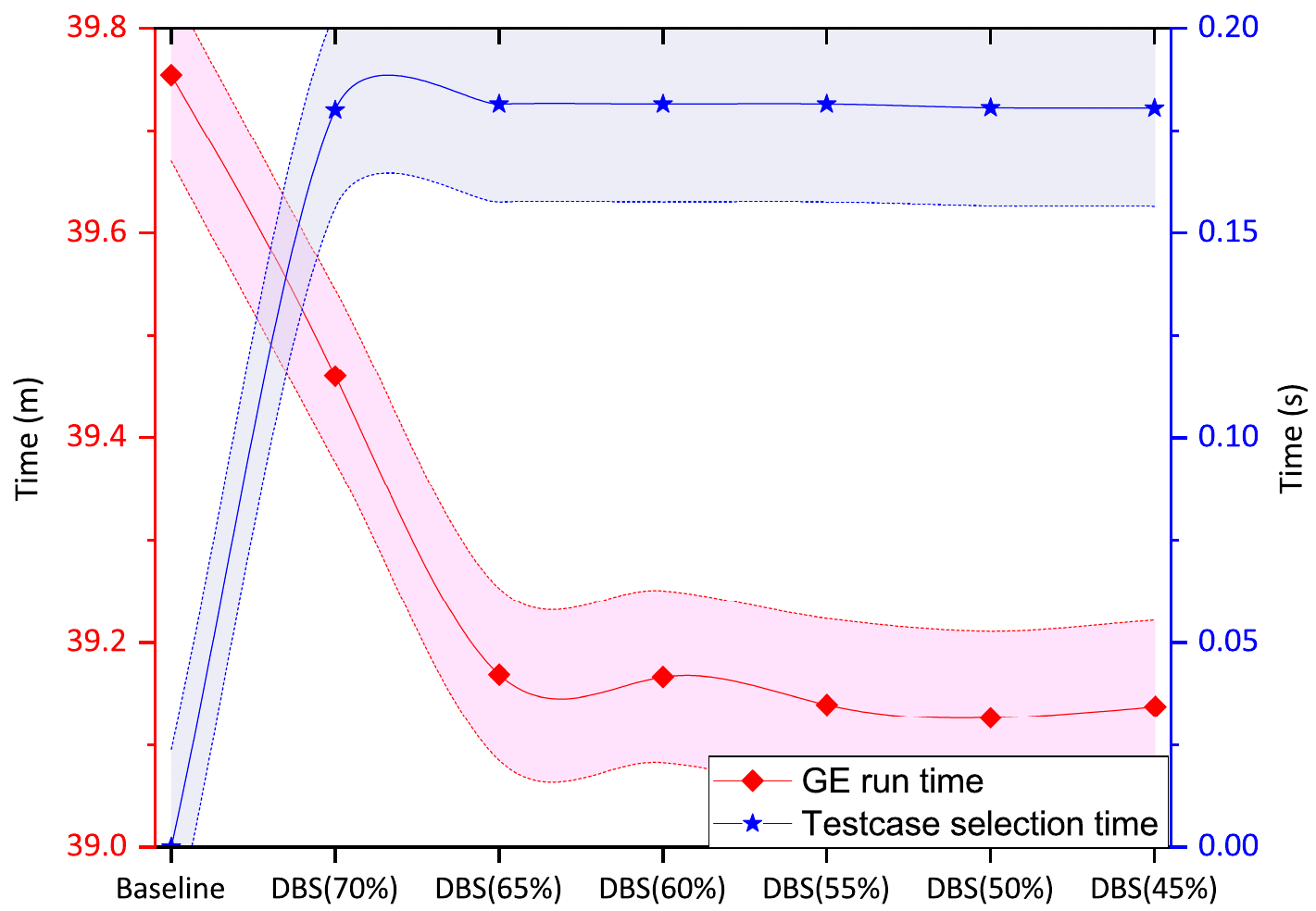}}
    \subfigure[Parity\label{fig:Time_Parity}]{\includegraphics[width=0.4\textwidth]{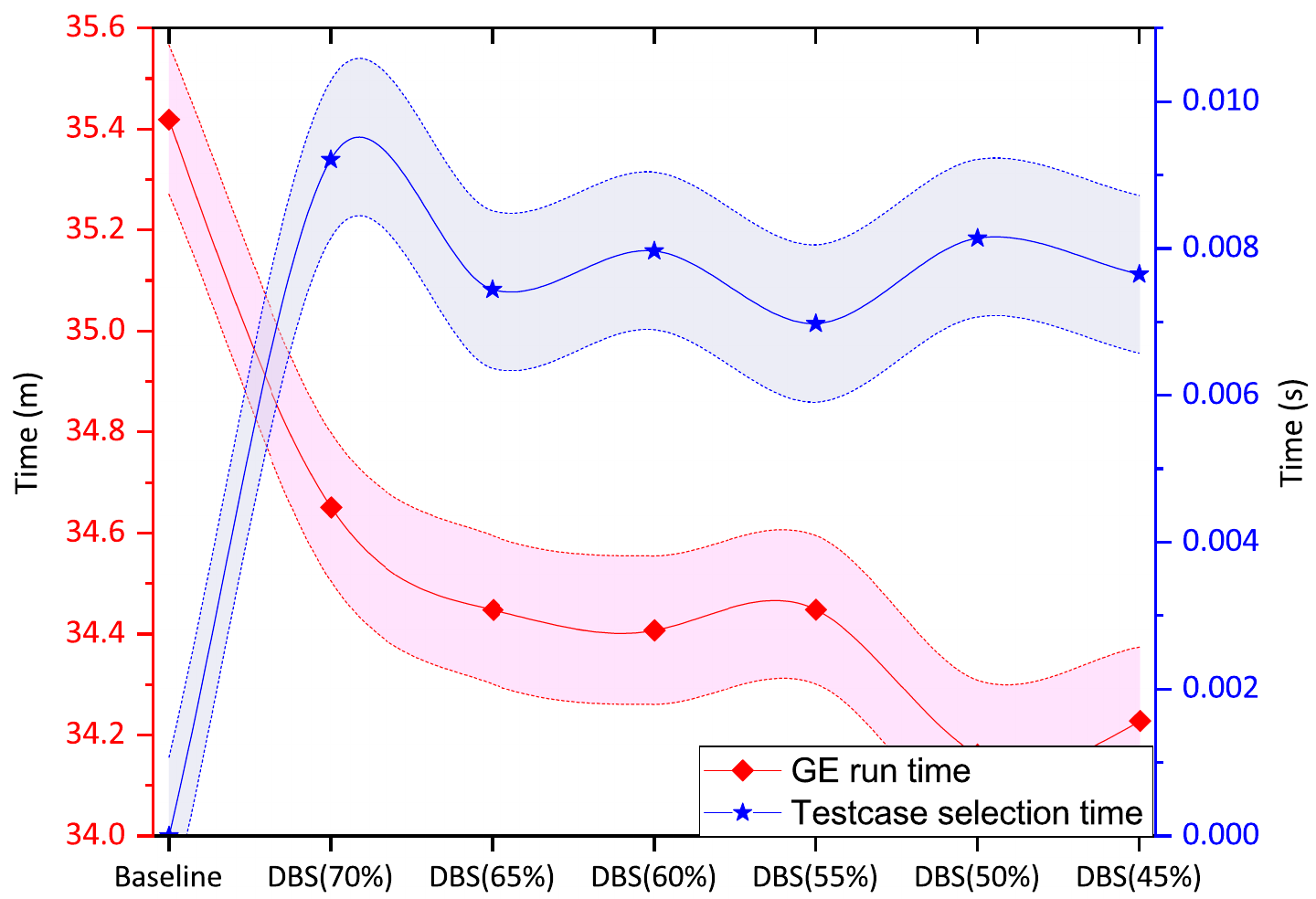}}\\
   \subfigure[Multiplexer\label{fig:Time_MUX}]{\includegraphics[width=0.4\textwidth]{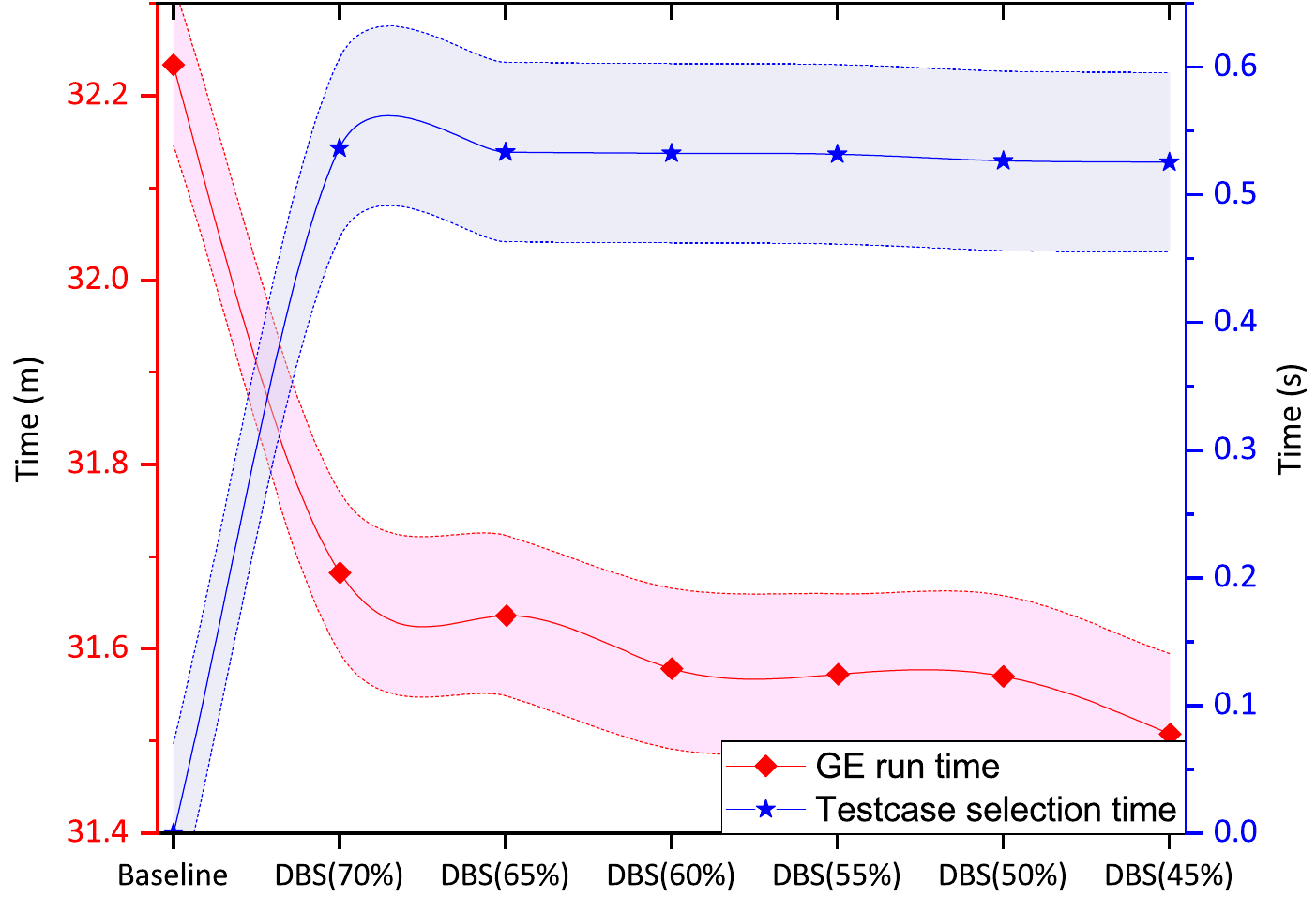}}
    \subfigure[ALU\label{fig:Time_ALU}]{\includegraphics[width=0.4\textwidth]{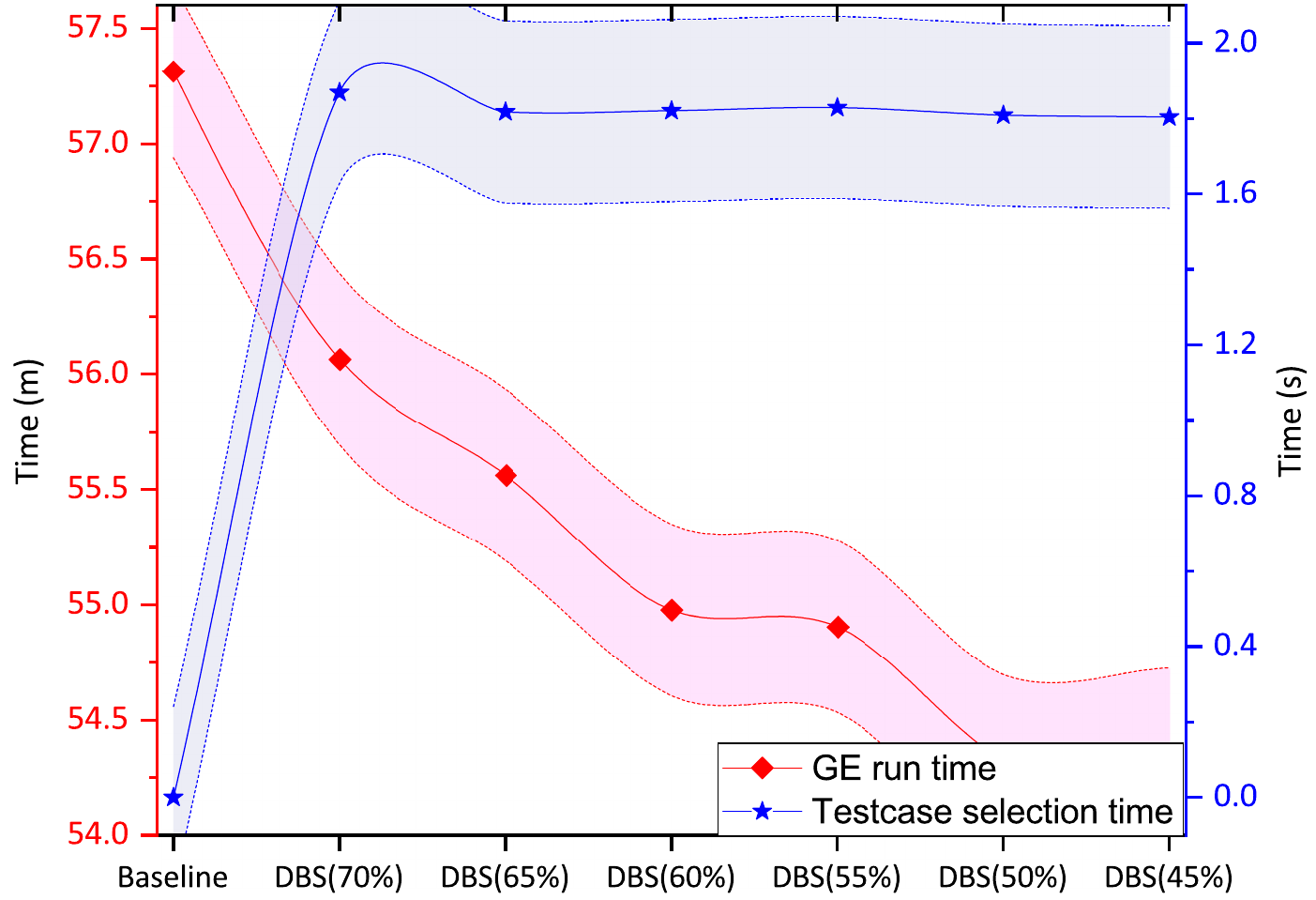}}
    \caption{Total time taken in test case selection (in seconds) and average time taken per GE run (in minutes) along with std error for digital circuit benchmarks.}
    \label{fig:CircuitTime}
\end{figure}

The training time or GE run time analysis for digital circuit benchmarks is presented in Figure \ref{fig:CircuitTime}. The DBS algorithm has a significant contribution in speeding up the evolutionary process and hence has a smaller run time than the baseline.

It is evident from the time analysis in Figures \ref{fig:SynTime}-\ref{fig:CircuitTime} that test case selection using DBS can greatly reduce the fitness evaluation time, and hence the total \gls{ge} run time.

\subsection{Solution Size}
\label{subsec:IndSize}
To determine if the training data selected using DBS impacts the size of solutions achieved, we analyze the effective individual size of the best solutions over 30 independent runs carried out on each instance of training data used on 24 benchmarks. This has several advantages, as it can give insight into their impact on GE run time and whether or not bloat occurs. We use the same approach for all of the benchmarks used.

\begin{figure}[!h]
    \centering
    \subfigure[Keijzer-4\label{fig:Eff_Keijzer4}]{\includegraphics[width=0.32\textwidth]{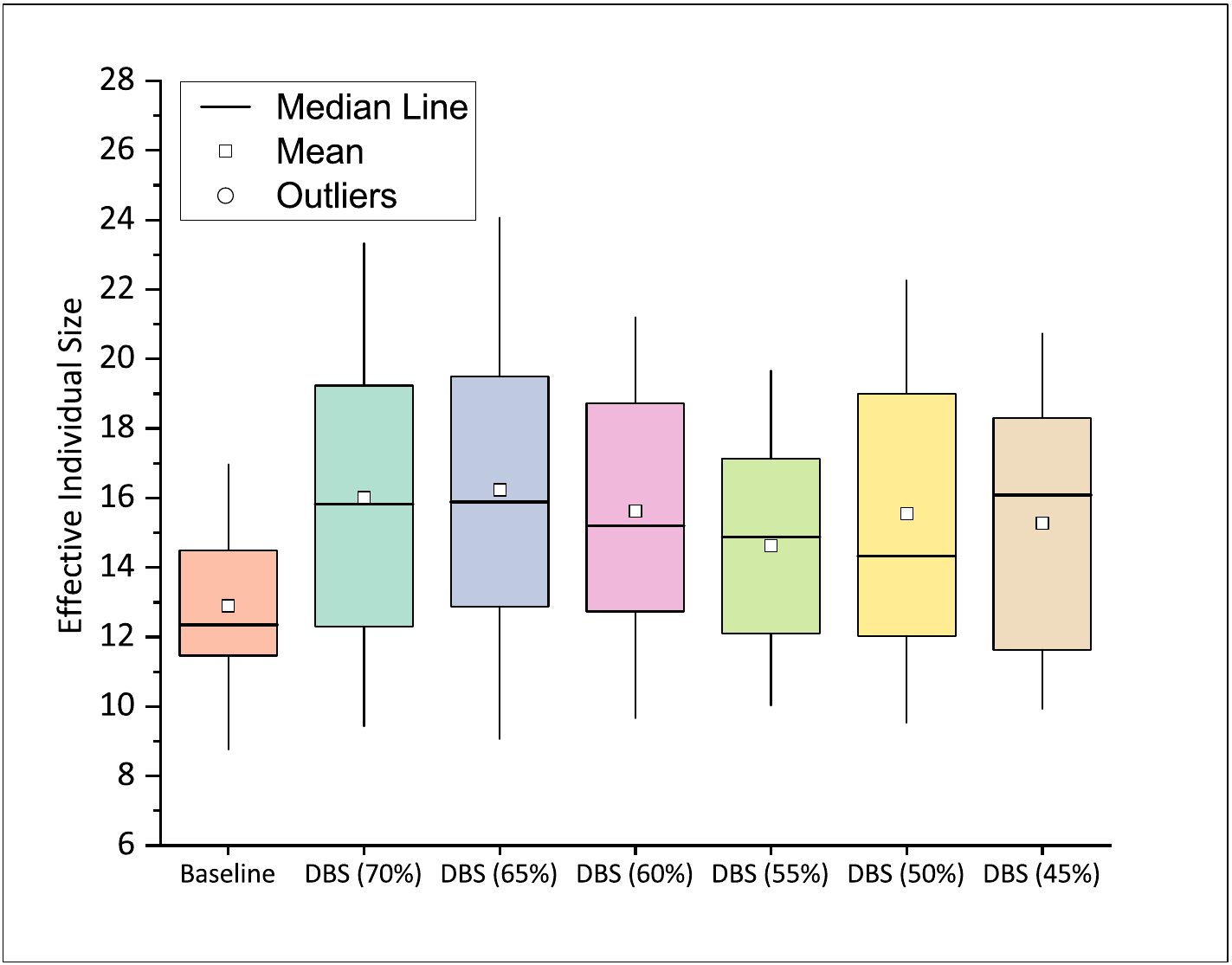}}
    \subfigure[Keijzer-9\label{fig:Eff_Keijzer9}]{\includegraphics[width=0.32\textwidth]{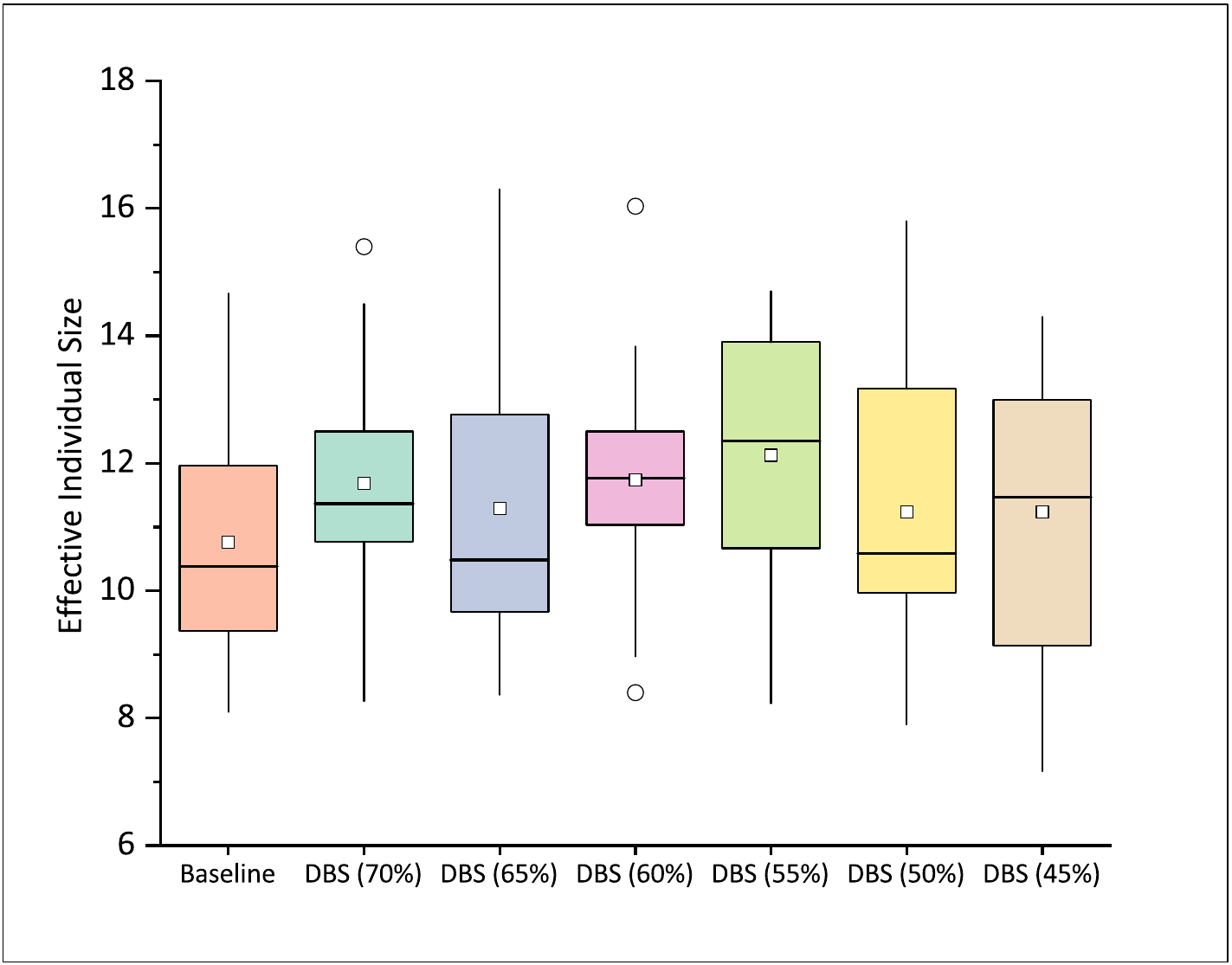}}
   \subfigure[Keijzer-10\label{fig:Eff_Keijzer10}]{\includegraphics[width=0.32\textwidth]{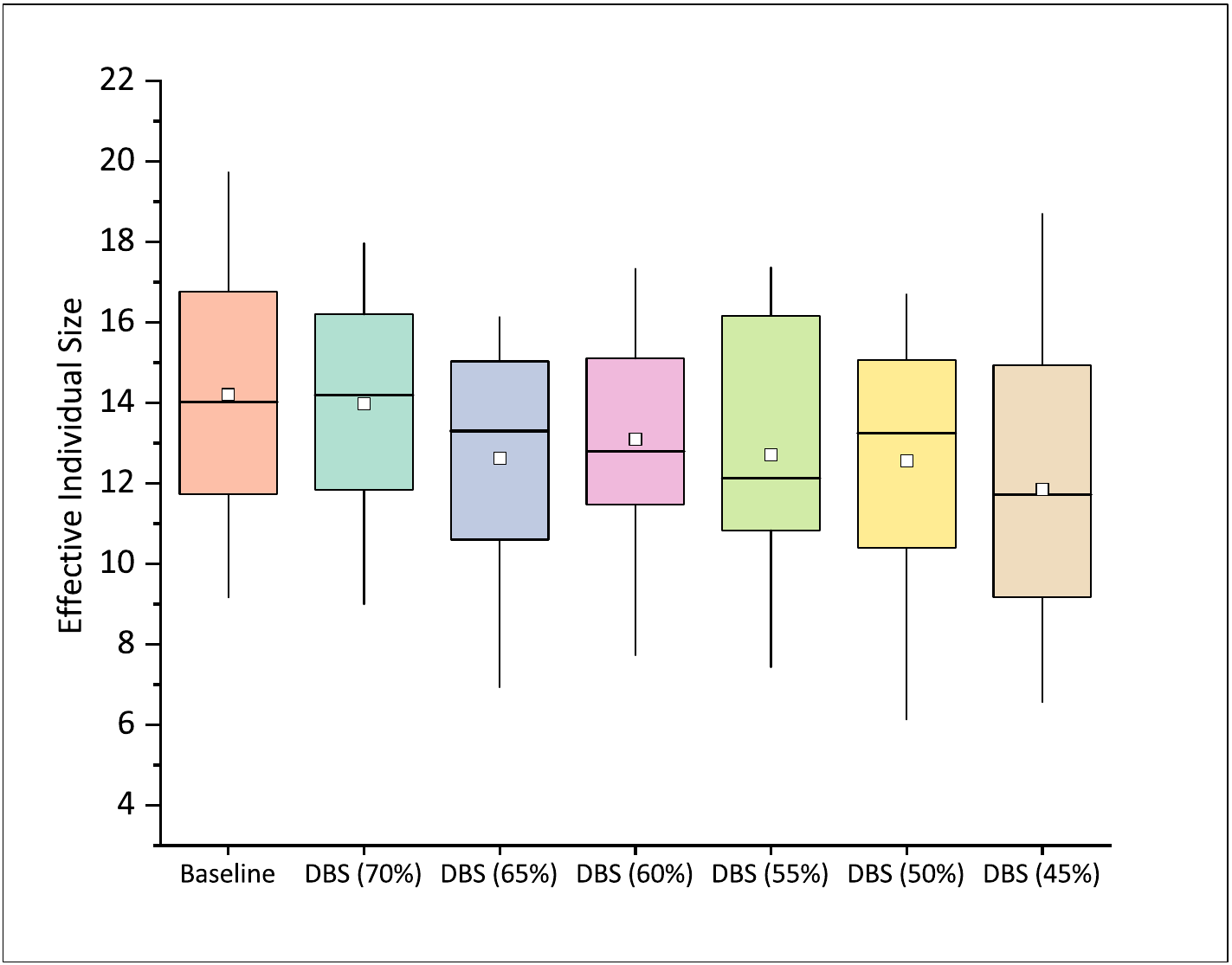}}\\
    \subfigure[Keijzer-14\label{fig:Eff_Keijzer14}]{\includegraphics[width=0.32\textwidth]{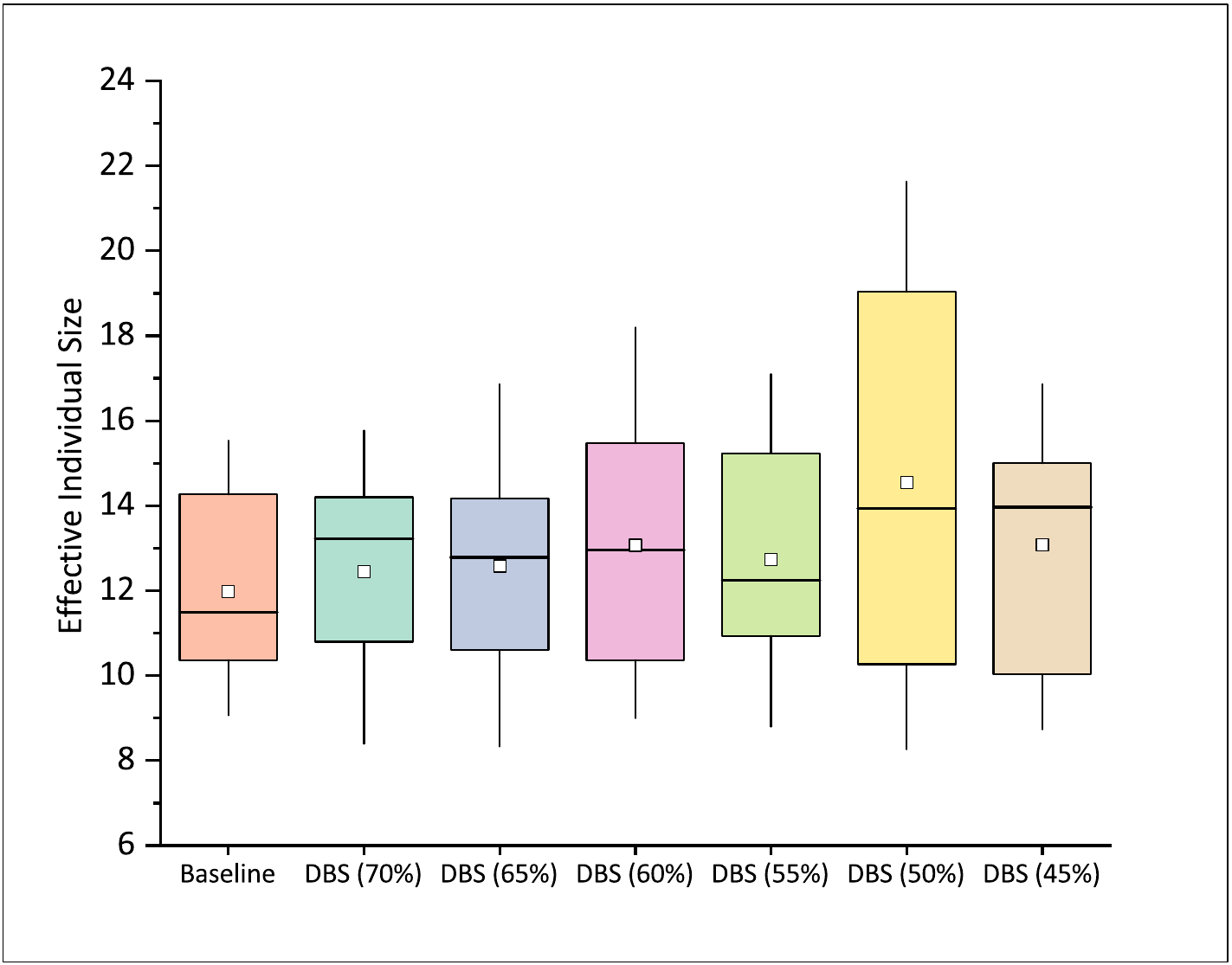}}
    \subfigure[Nguyen-9\label{fig:Eff_Nguyen9}]{\includegraphics[width=0.32\textwidth]{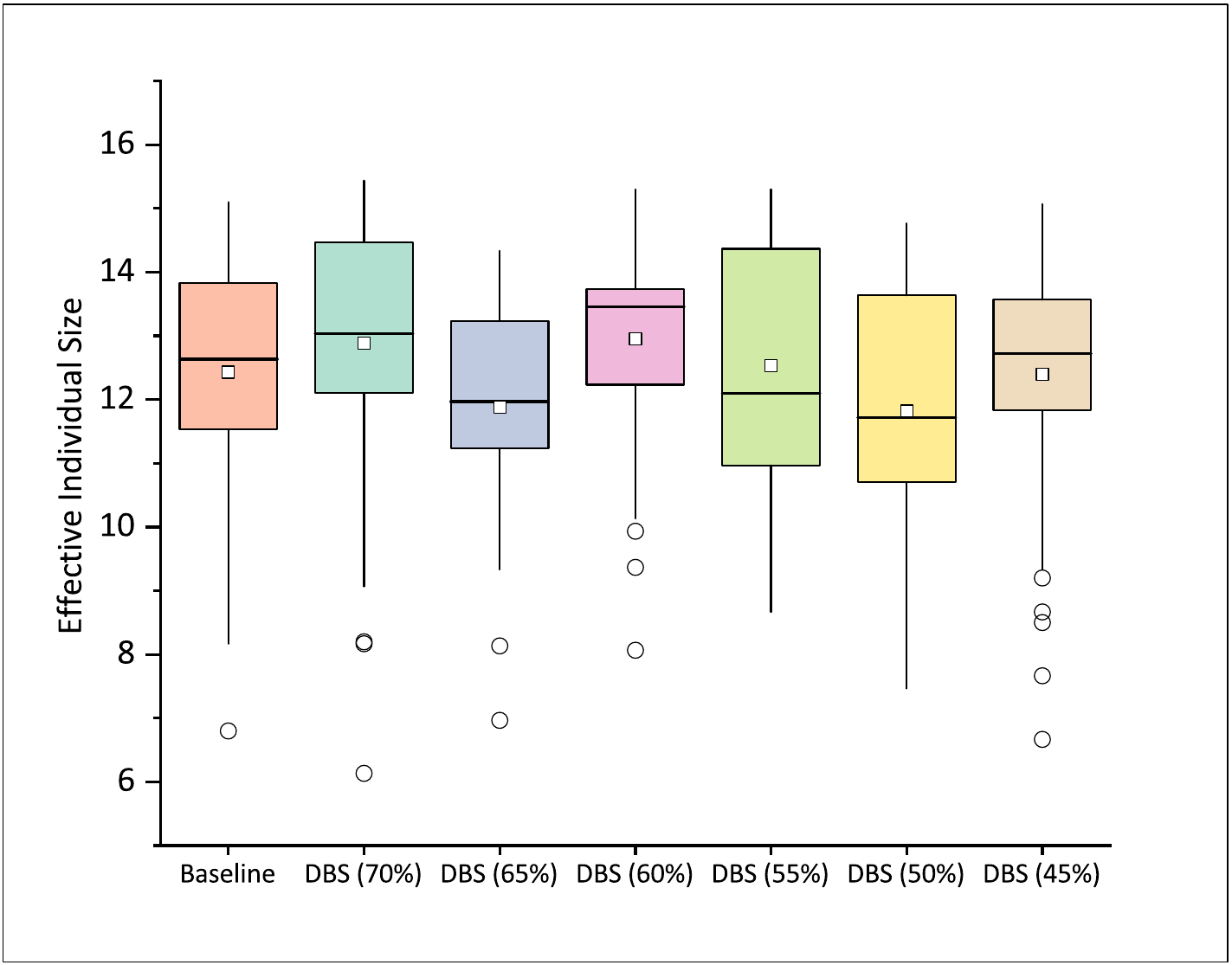}}
    \subfigure[Nguyen-10\label{fig:Eff_Nguyen10}]{\includegraphics[width=0.32\textwidth]{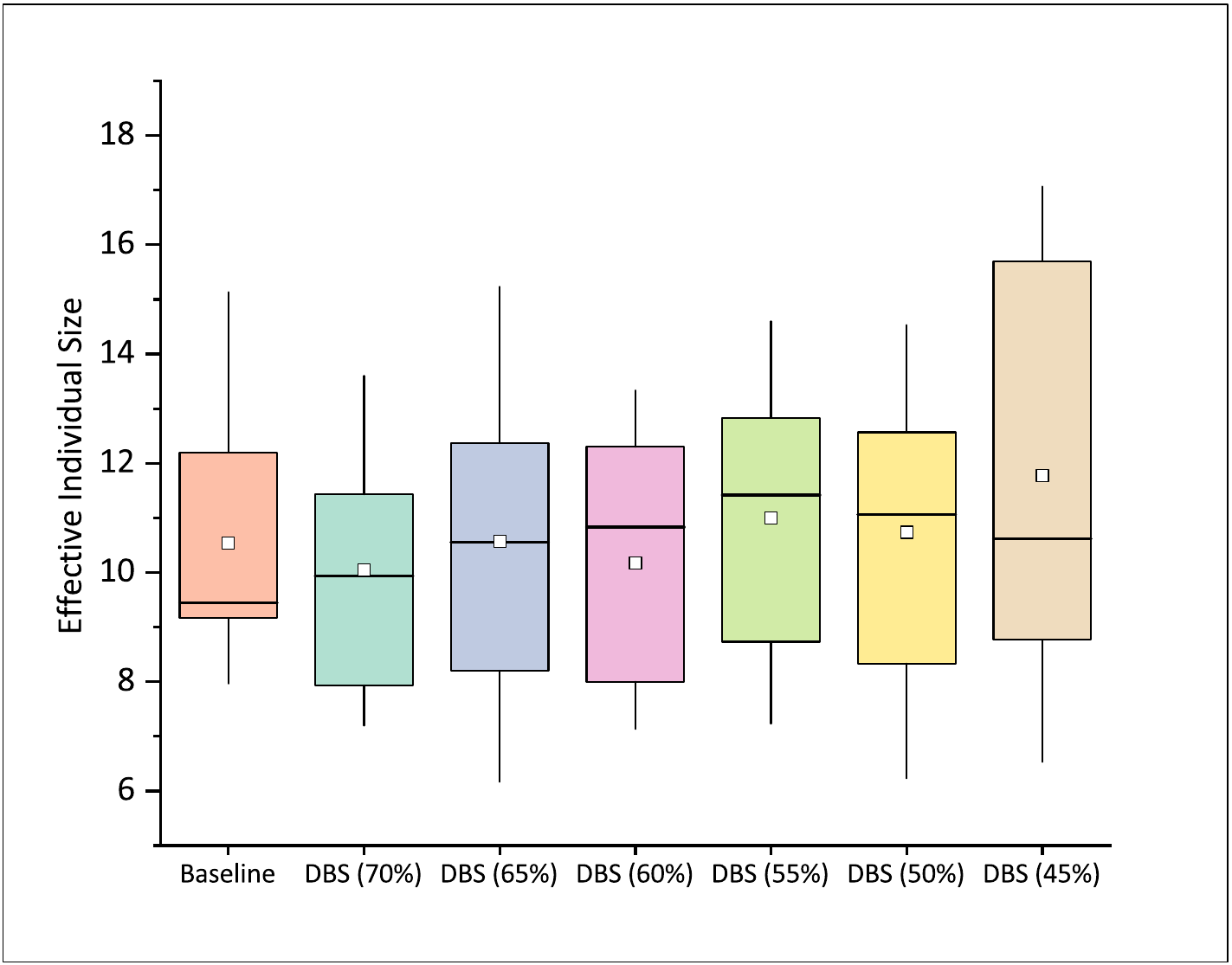}}\\
     \subfigure[Keijzer-5\label{fig:Eff_Keijzer5}]{\includegraphics[width=0.32\textwidth]{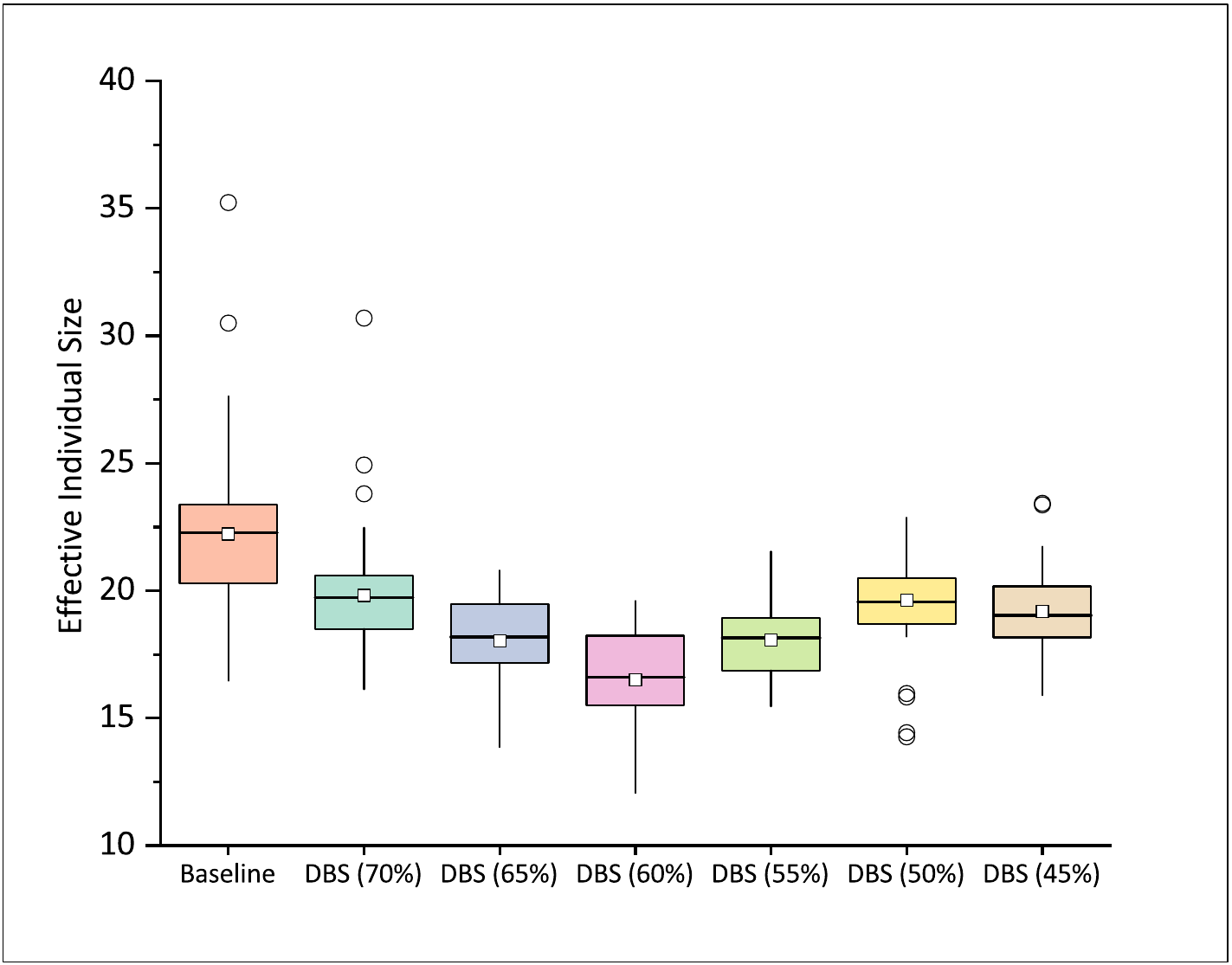}}
    \subfigure[Vladislavleva-5\label{fig:Eff_Vlad5}]{\includegraphics[width=0.32\textwidth]{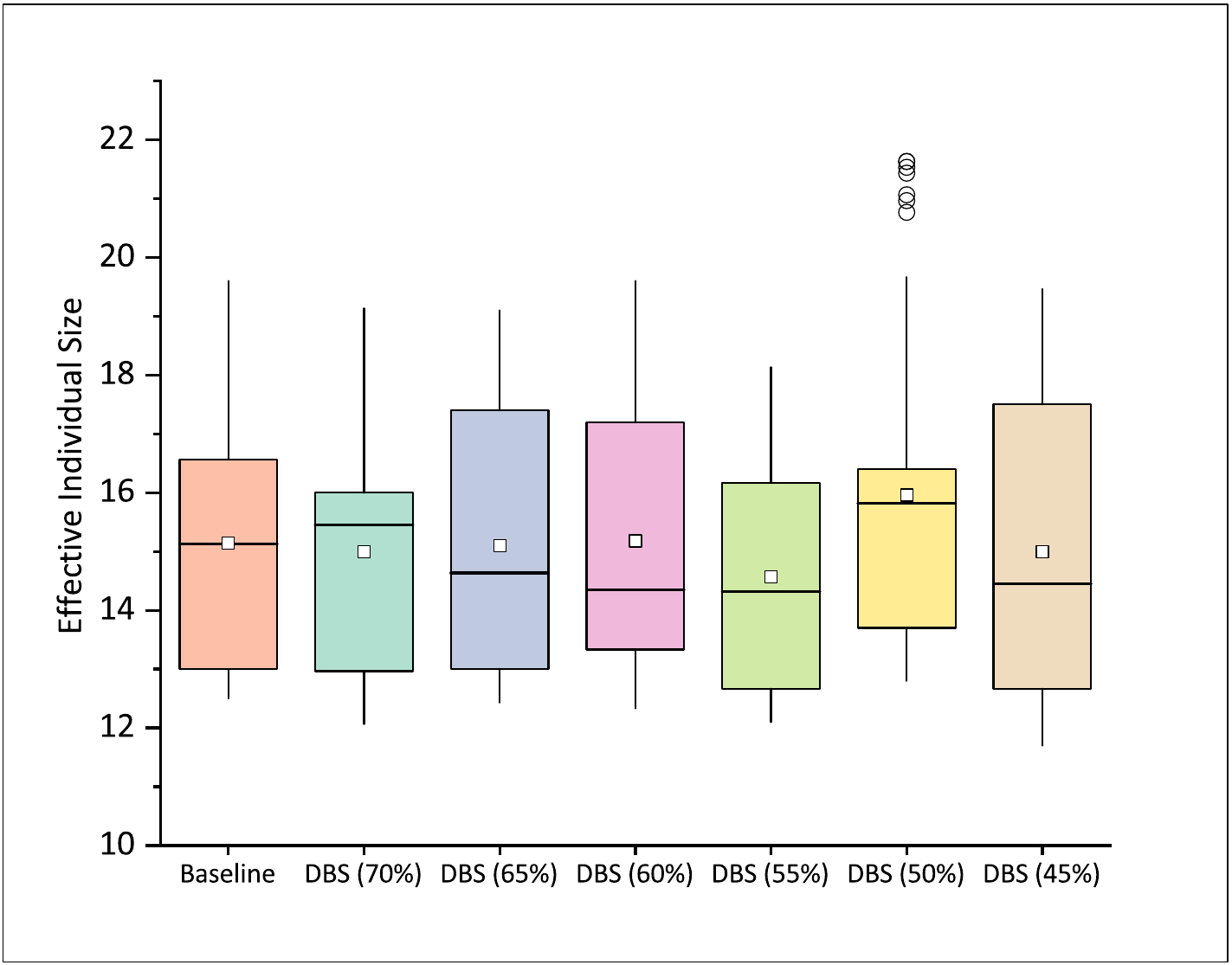}}
    \subfigure[Korns-11\label{fig:Eff_korns11}]{\includegraphics[width=0.32\textwidth]{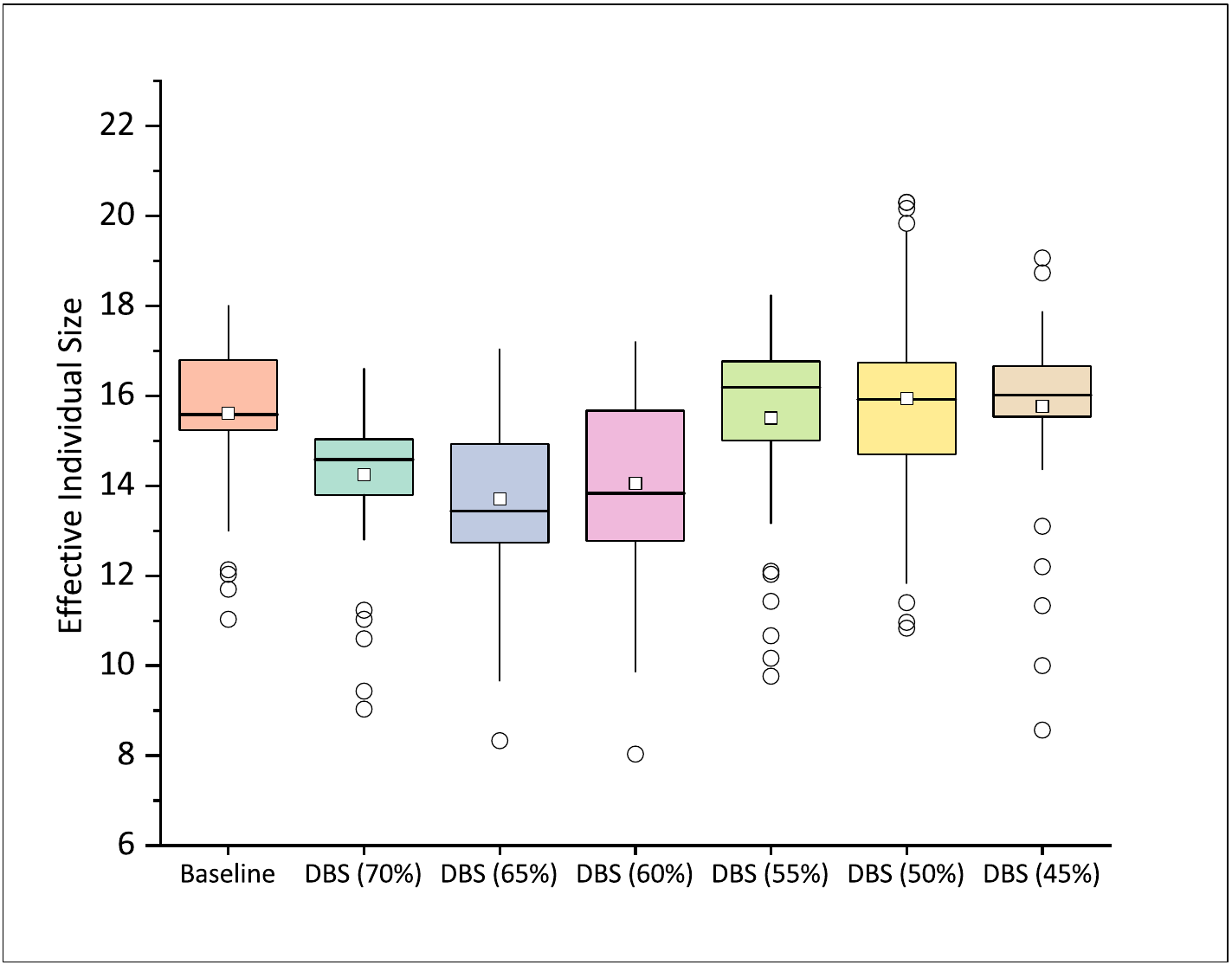}}\\
    \subfigure[Korns-12\label{fig:Eff_korns12}]{\includegraphics[width=0.32\textwidth]{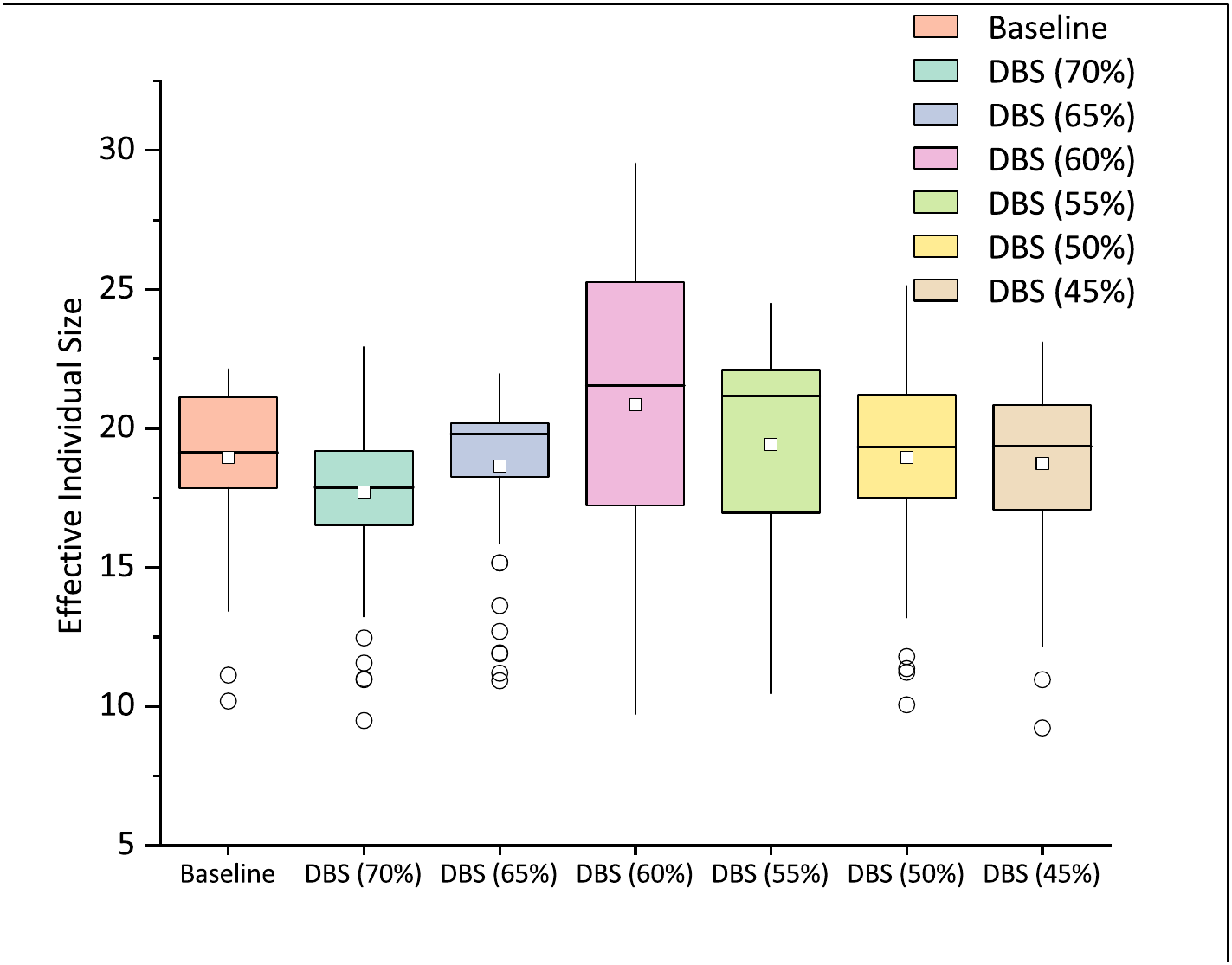}}
    \caption{Mean effective individual size of best solutions obtained on synthetic SR benchmarks.}
    \label{fig:SynEff}
\end{figure}

Figure \ref{fig:SynEff} presents the effective individual size of 10 synthetic SR benchmarks. With the exception of \texttt{Keijzer-4} and \texttt{Keijzer-14}, the solutions obtained using the DBS training data have effective individual sizes that are either similar to or smaller than the baseline. 

Recalling Figure \ref{fig:Time_Keijzer4} through \ref{fig:Time_Nguyen9}, we can observe that the time analysis either shows the expected pattern of lowering run time while decreasing the training data or if it doesn't, the time difference is too small. In the experiment with \texttt{Nguyen-10} in Figure \ref{fig:Eff_Nguyen10}, DBS (55\%) takes noticeably longer to train than the experiments with bigger training data. The fact that DBS's mean effective individual size at DBS (55\%) is larger than that of individuals trained on a higher training budget is one possible explanation. The results for the following experiments are presented in Figure \ref{fig:Eff_Keijzer5} through Figure \ref{fig:Eff_korns12}. The GE run time graphs for the same in Figure \ref{fig:Time_Keijzer5}-\ref{fig:Time_korns12} show the outcome as anticipated. Usually, a modest variation in effective individual size has less of an influence if the training size, number of features, and GE run-time are large.

Furthermore, we concur that it is challenging to relate how individual size affects the overall training time when the GE run time is very small, but it is conceivable if the difference in individual size is sufficiently significant. Second, we are less inclined to look at instances where GE run time behaves as expected, that is, when it becomes shorter as training data size decreases, but we are more interested in situations where GE run time increases significantly as training data size reduces.

\begin{figure}[!h]
    \centering
    \subfigure[Airfoil\label{fig:EffSize_Airfoil}]{\includegraphics[width=0.32\textwidth]{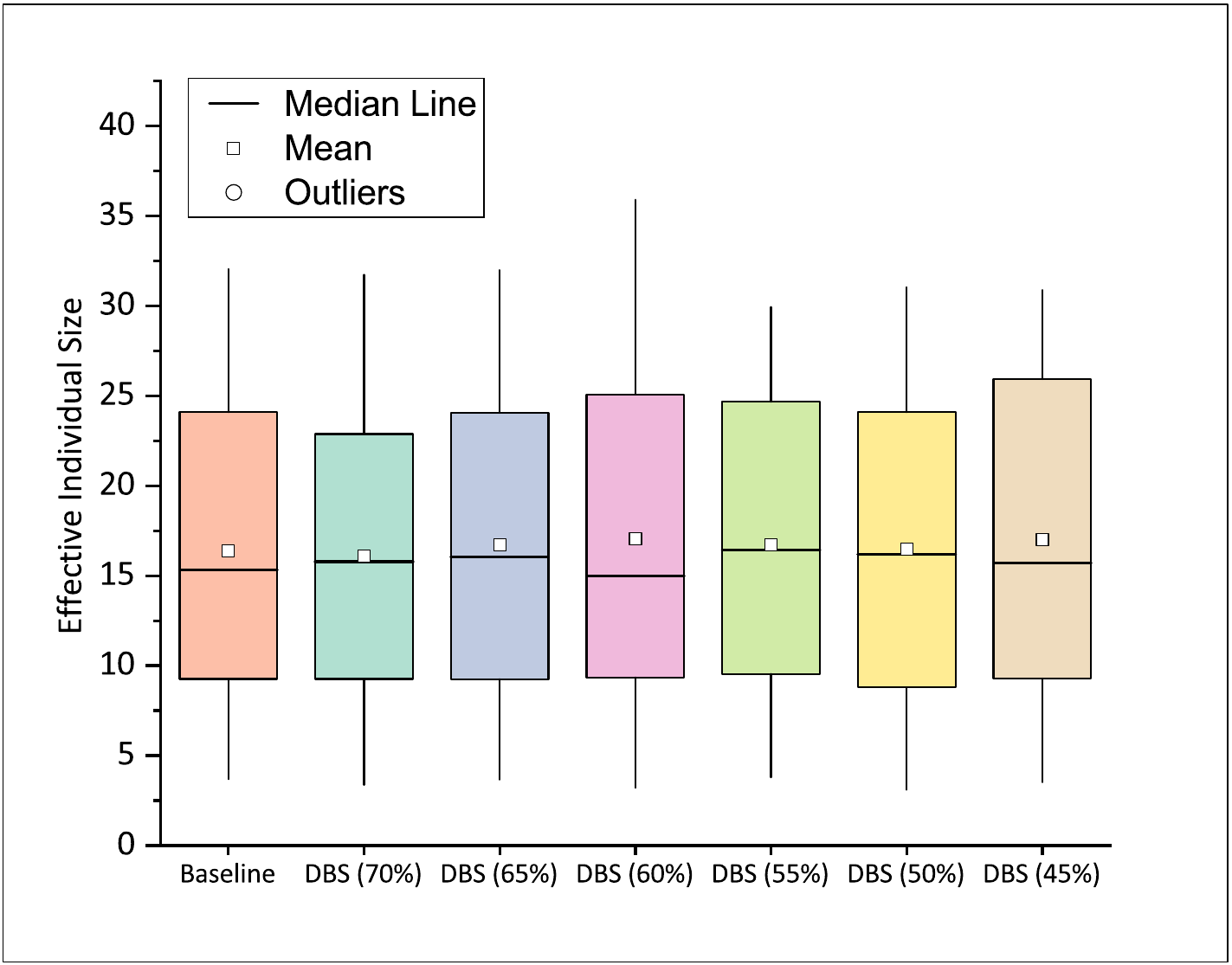}}
    \subfigure[Heating\label{fig:EffSize_Heating}]{\includegraphics[width=0.32\textwidth]{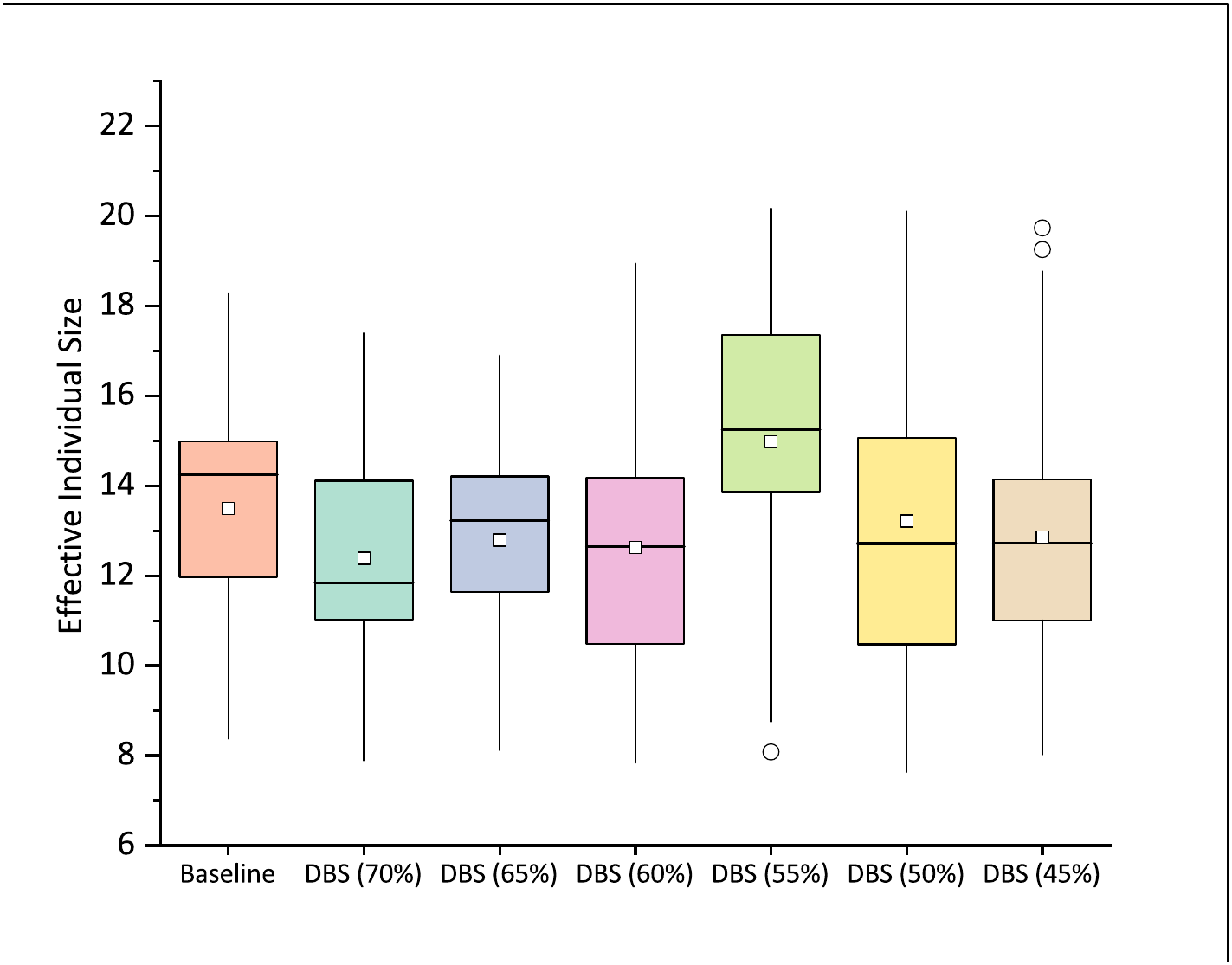}}
   \subfigure[Cooling\label{fig:EffSize_Cooling}]{\includegraphics[width=0.32\textwidth]{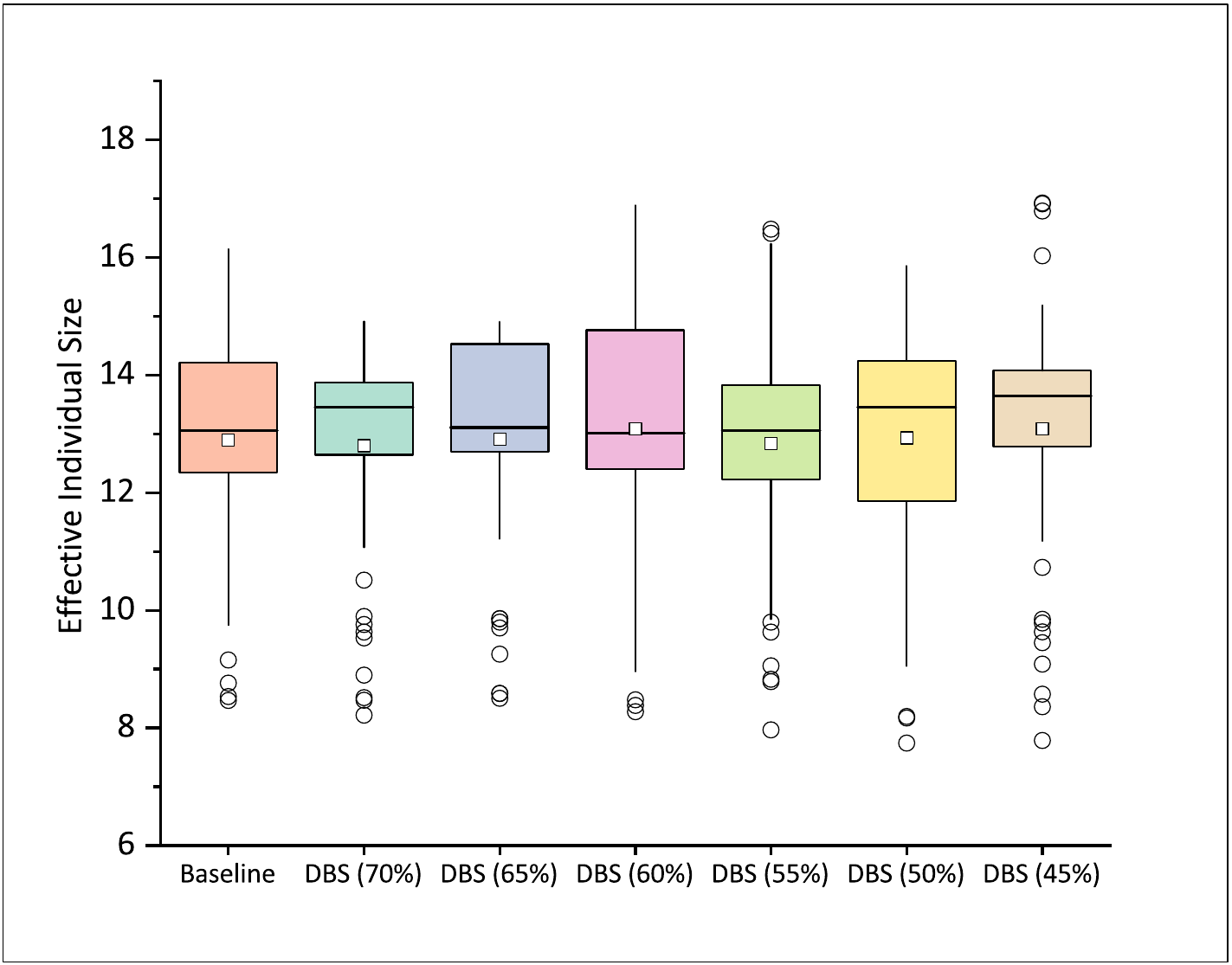}}\\
    \subfigure[Concrete\label{fig:EffSize_Concrete}]{\includegraphics[width=0.32\textwidth]{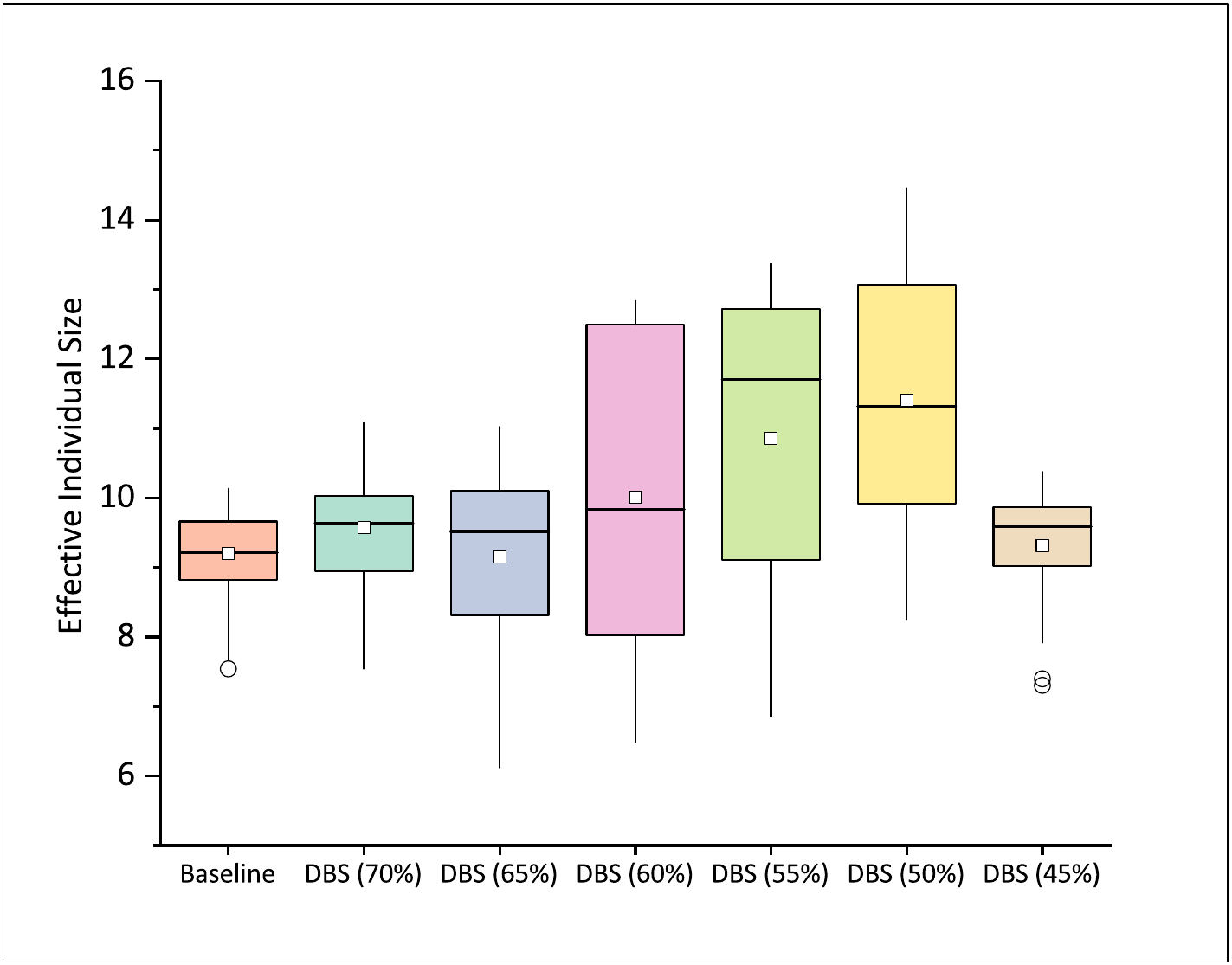}}
    \subfigure[Redwine\label{fig:EffSize_Redwine}]{\includegraphics[width=0.32\textwidth]{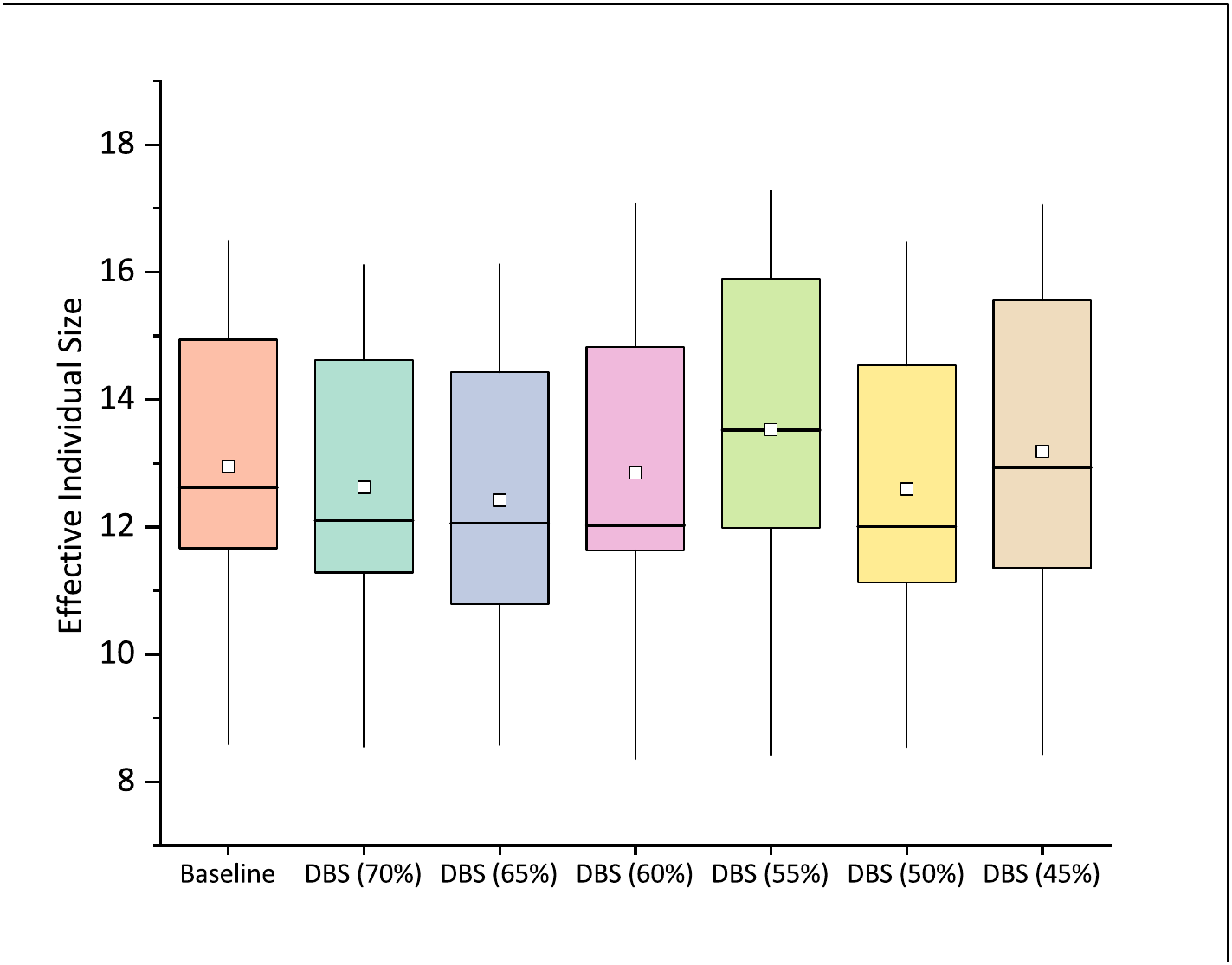}}
    \subfigure[Whitewine\label{fig:EffSize_Whitewine}]{\includegraphics[width=0.32\textwidth]{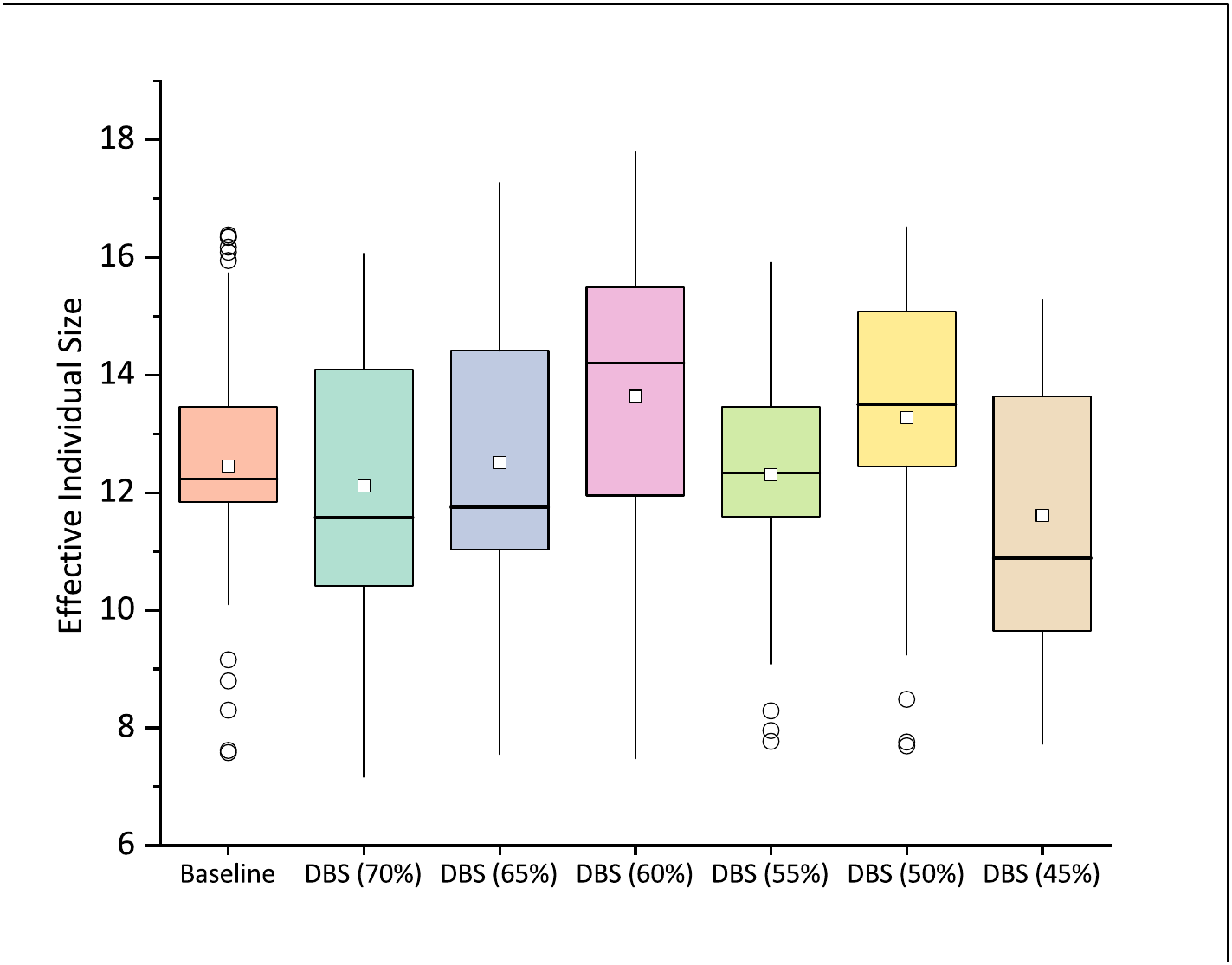}}\\
     \subfigure[Housing\label{fig:EffSize_Housing}]{\includegraphics[width=0.32\textwidth]{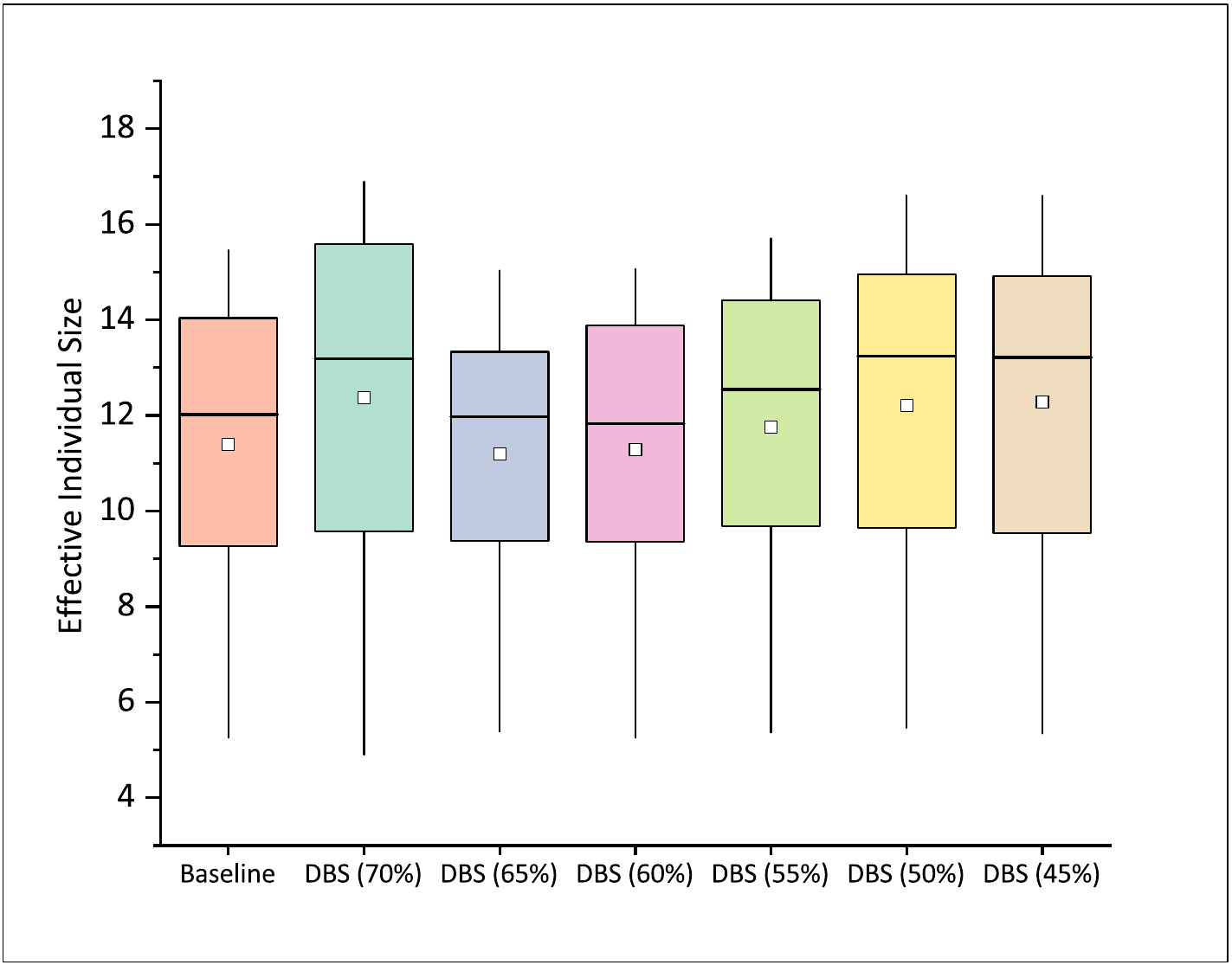}}
    \subfigure[Pollution\label{fig:EffSize_Pollution}]{\includegraphics[width=0.32\textwidth]{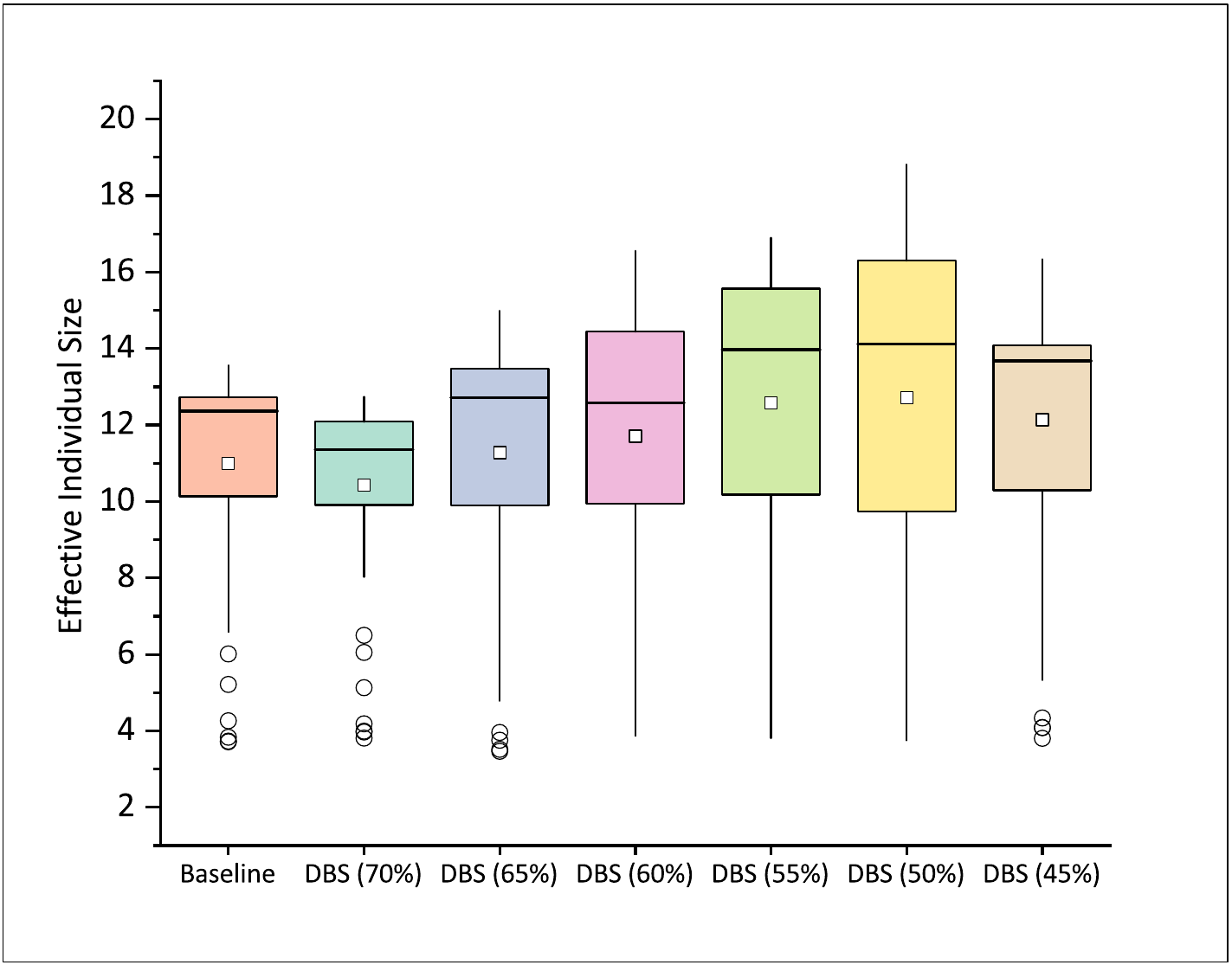}}
    \subfigure[Dowchem\label{fig:EffSize_Dowchem}]{\includegraphics[width=0.32\textwidth]{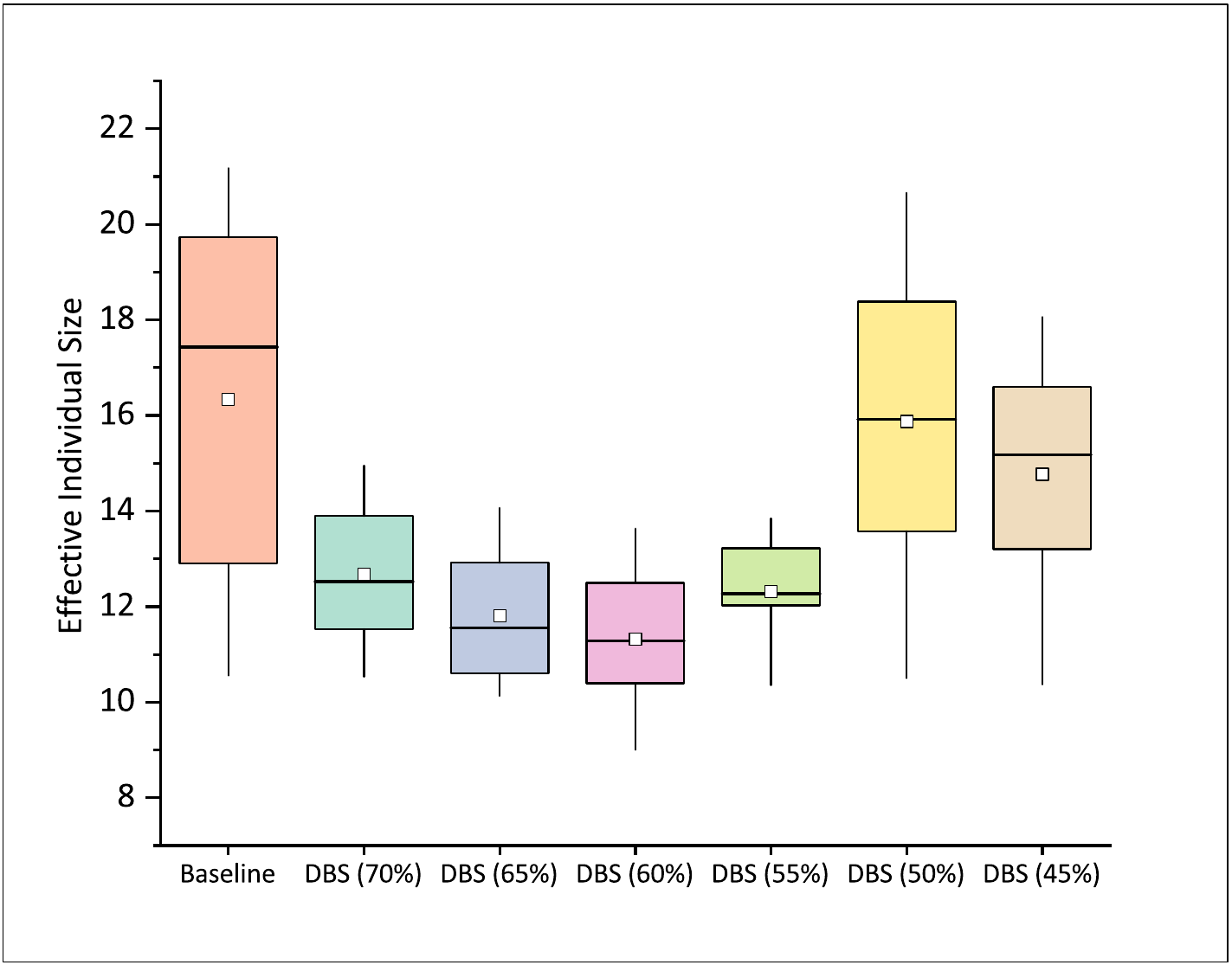}}\\
    \subfigure[Crime\label{fig:EffSize_Crime}]{\includegraphics[width=0.32\textwidth]{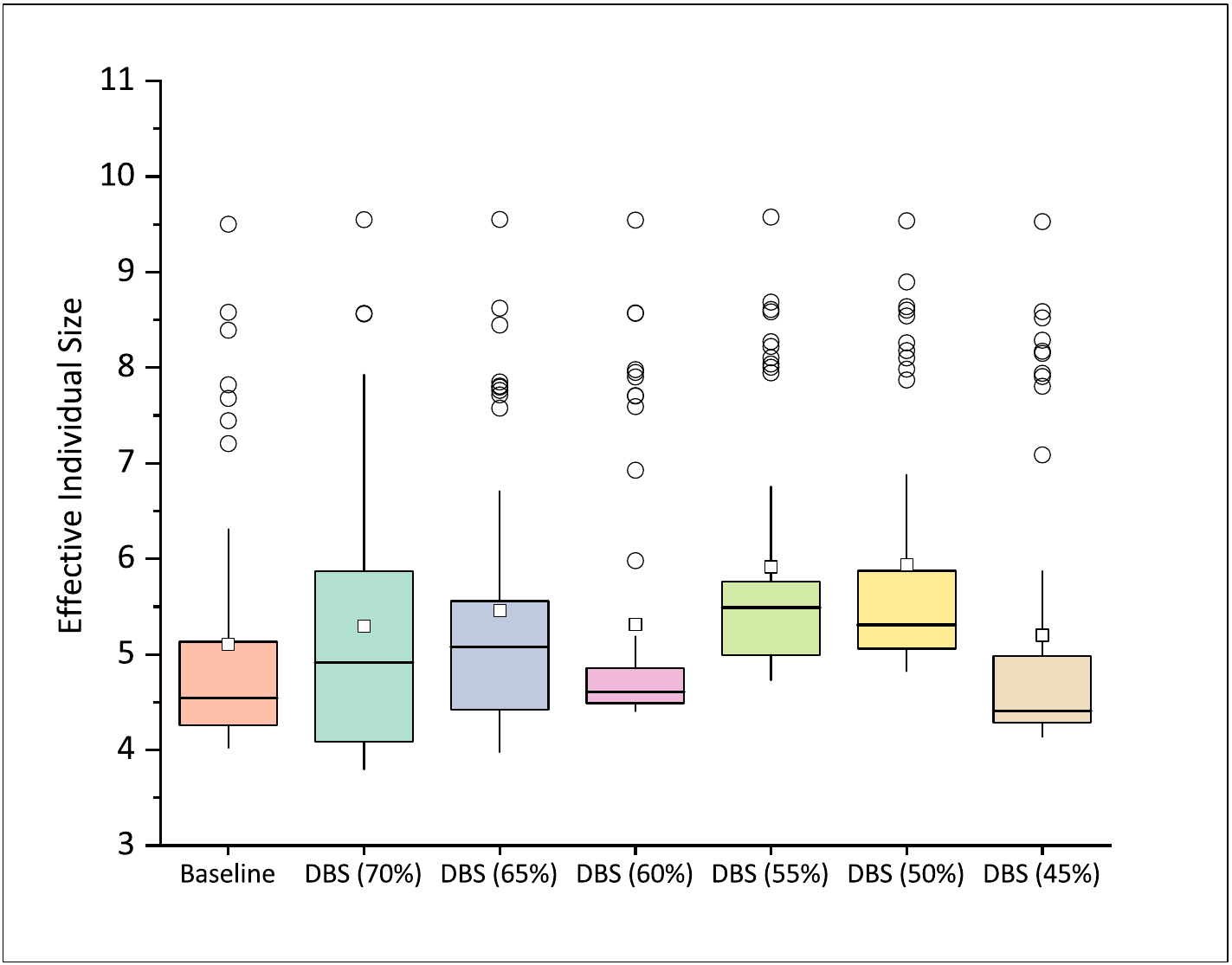}}
    \caption{Mean effective individual size of best solutions obtained on real-world SR data set.}
    \label{fig:RealEffSize}
\end{figure}

We perform a similar analysis for real-world SR problems, which is shown in Figure \ref{fig:RealEffSize}. In the case of \texttt{airfoil}, the mean effective individual size is similar, and consequently, we observe the GE run time in Figure \ref{fig:Time_Airfoil} is somewhat as expected or the run time difference is small in opposite cases. For instance, DBS (50\%) and DBS (45\%) took slightly more time than DBS (55\%) to train the model; the time difference is low. In the case of \texttt{heating}, the mean effective size at DBS (55\%) is slightly higher, but a direct impact is not observed in the GE time graph present in Figure \ref{fig:Time_Heating}. A possible explanation is that as the complexity of the problem increases, a small difference in the size of individuals does not have much impact on training time, provided the training time is sufficiently large. The mean effective individual size in the case of \texttt{cooling} experiment is almost comparable to each other. In the case of \texttt{concrete} dataset, the effective individual size obtained in the experiment using DBS (60\%), DBS (55\%), and DBS (50\%) are in increasing order as shown in Figure \ref{fig:EffSize_Concrete}. A direct impact of this is observed in the GE run time. Remember, the training time in Figure \ref{fig:Time_Concrete} is small, and the difference among individual sizes in Figure \ref{fig:EffSize_Concrete} is large. Conversely, we are less likely to see a direct correlation between individual size and training duration in complex problems.

The mean effective individual size of \texttt{redwine} (in Figure \ref{fig:EffSize_Redwine}) is comparable to that of their neighbors on the left. However, the effective individual size at DBS (55\%) is slightly higher, but as mentioned earlier, the impact of a small difference in individual size is hard to identify on complex problems having small run time differences unless thousands of independent runs are performed. A similar observation is made in \texttt{whitewine} and \texttt{housing} experiments as shown in Figure \ref{fig:EffSize_Whitewine}-\ref{fig:EffSize_Housing}. The effective individual size between two consecutive training data in \texttt{pollution} experiments is high, as reported in Figure \ref{fig:EffSize_Pollution}. The impact of large individual size on GE run time is noticeable in Figure \ref{fig:Time_Pollution}. This is possible due to the short GE run time and smaller training data size (see table \ref{table:trainSizes}) that causes the impact of individual size to be clearly noticeable. However, this is less possible if experiments using two distinct and sufficiently large training sets produce small size differences among individuals. The effective individual size for \texttt{dowchem} experiments is shown in Figure \ref{fig:EffSize_Dowchem}. DBS training data consistently produced individuals that were smaller than the baseline in terms of effective size. Figure \ref{fig:EffSize_Crime} shows the individuals’ sizes recorded in experiments using different training datasets from \texttt{crime} dataset.

\begin{figure}[!h]
    \centering
    \subfigure[Comparator\label{fig:EffSize_Comp}]{\includegraphics[width=0.4\textwidth]{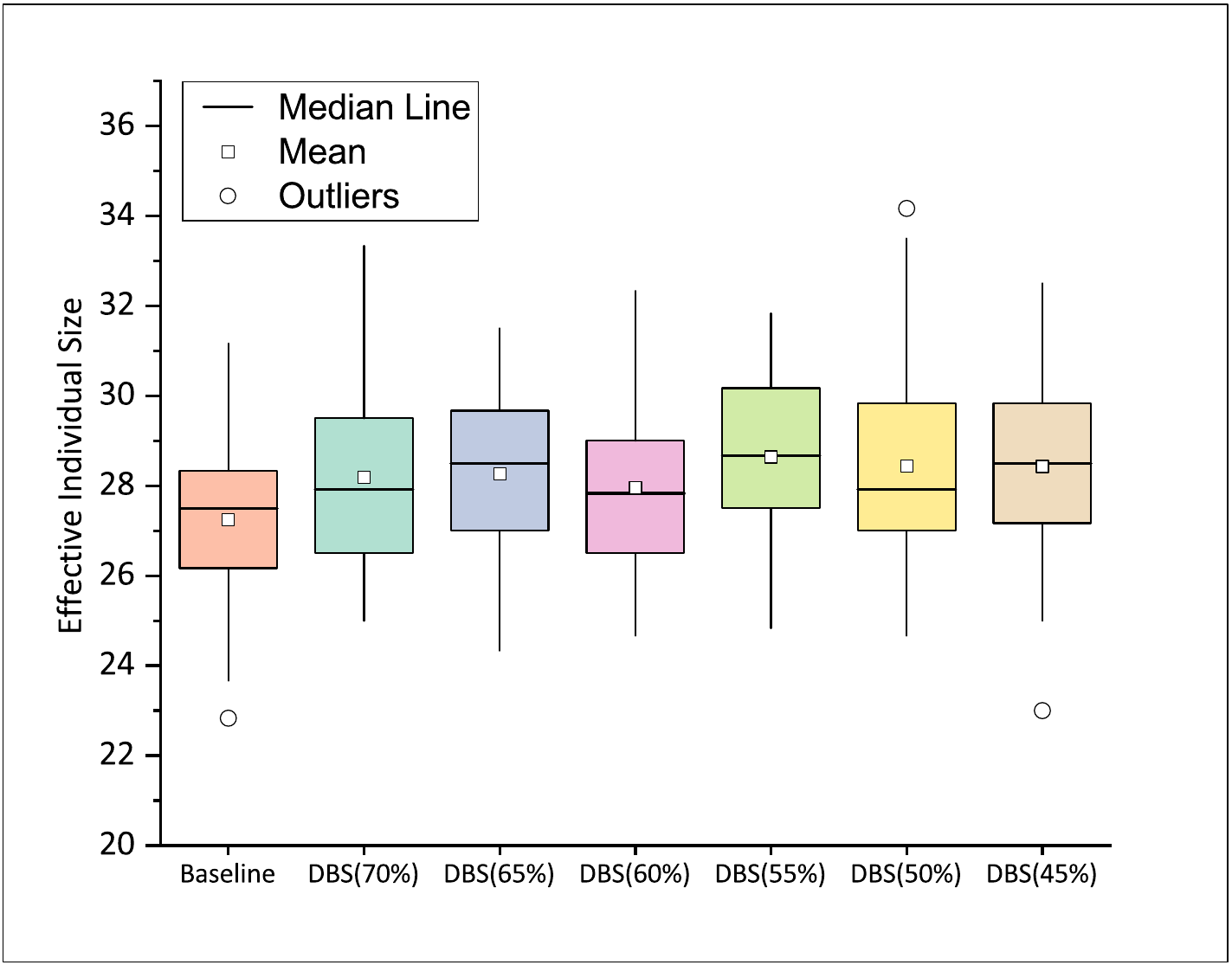}}
    \subfigure[Parity\label{fig:EffSize_Parity}]{\includegraphics[width=0.4\textwidth]{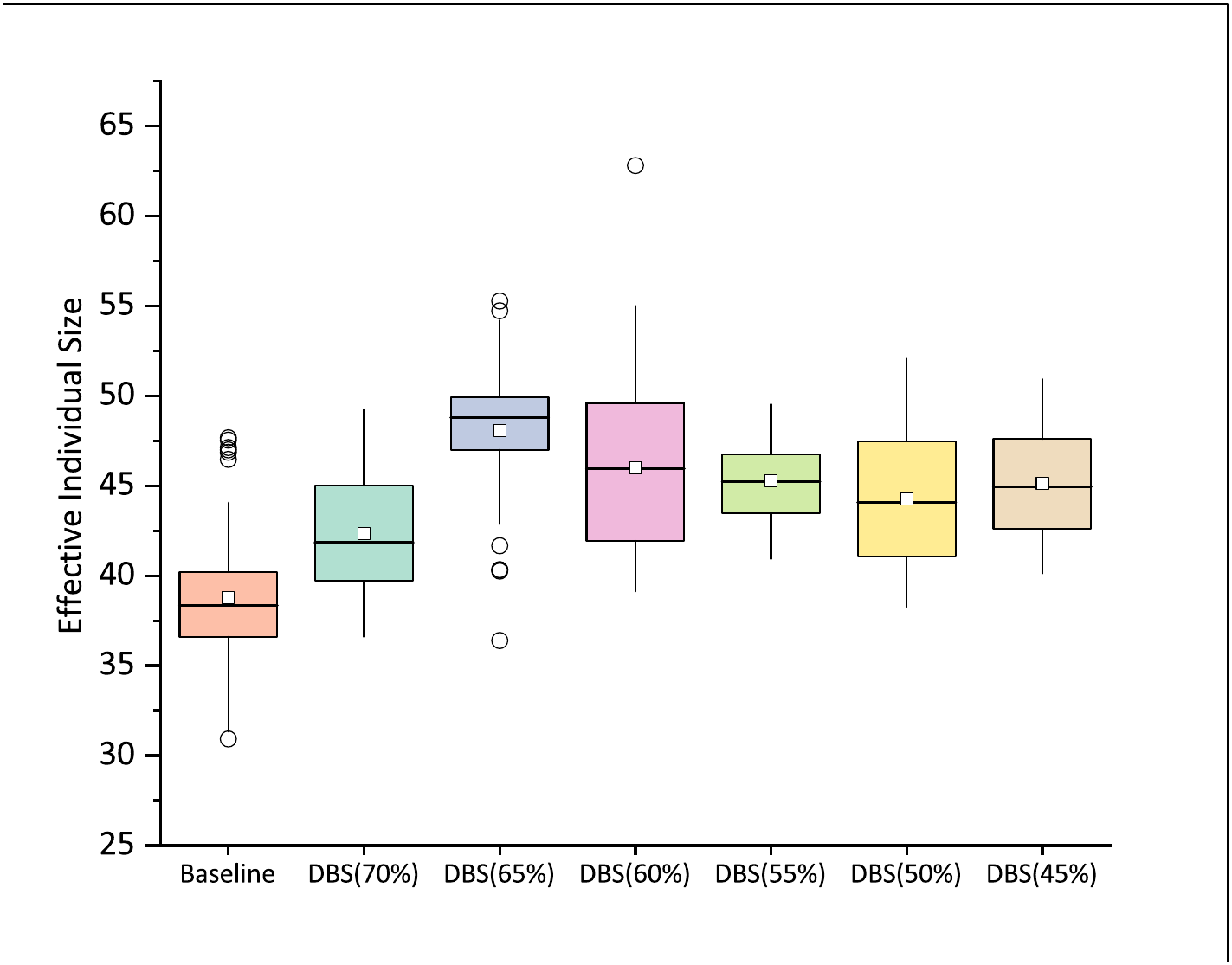}}\\
   \subfigure[Multiplexer\label{fig:EffSize_Mux}]{\includegraphics[width=0.4\textwidth]{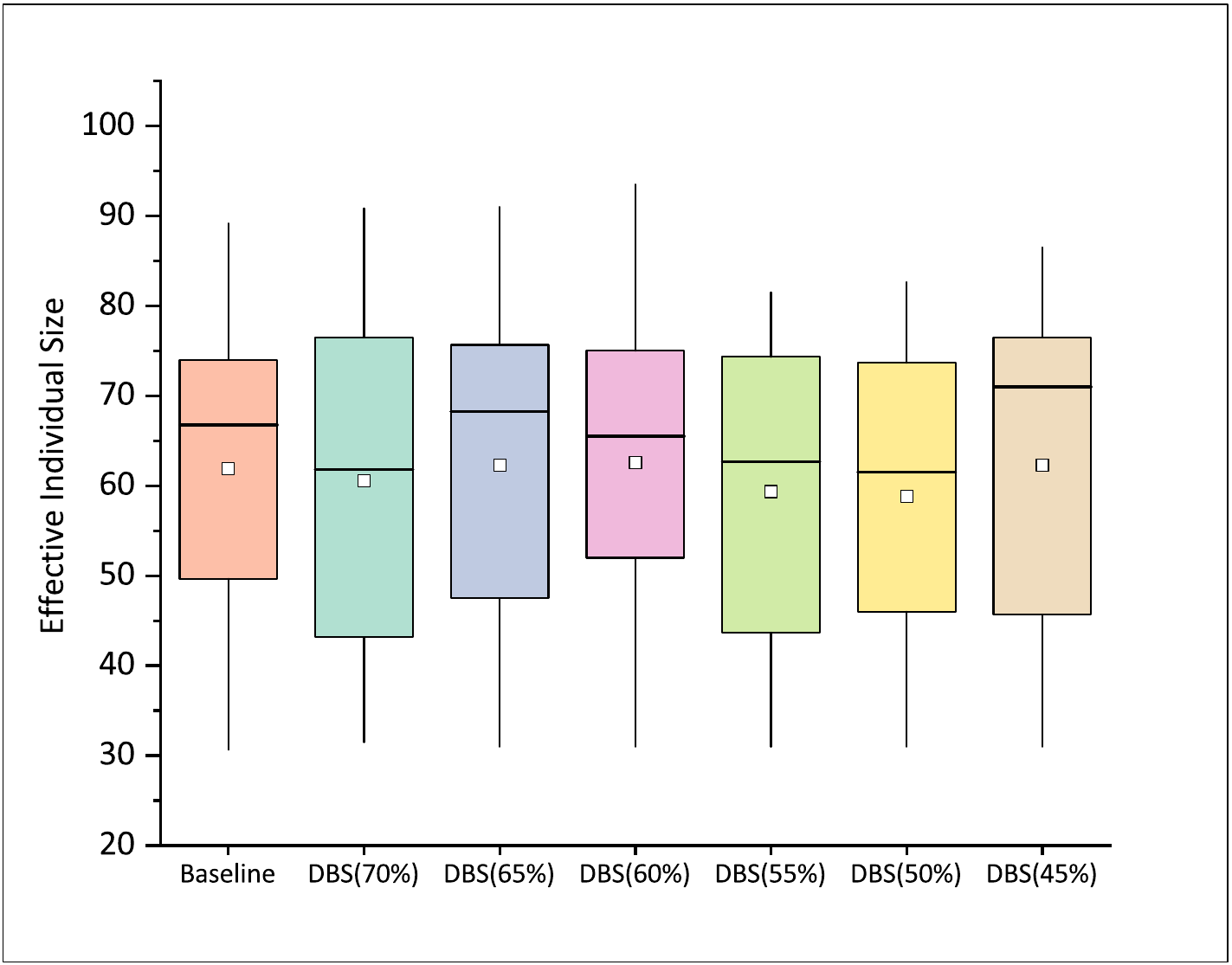}}
    \subfigure[ALU\label{fig:EffSize_ALU}]{\includegraphics[width=0.4\textwidth]{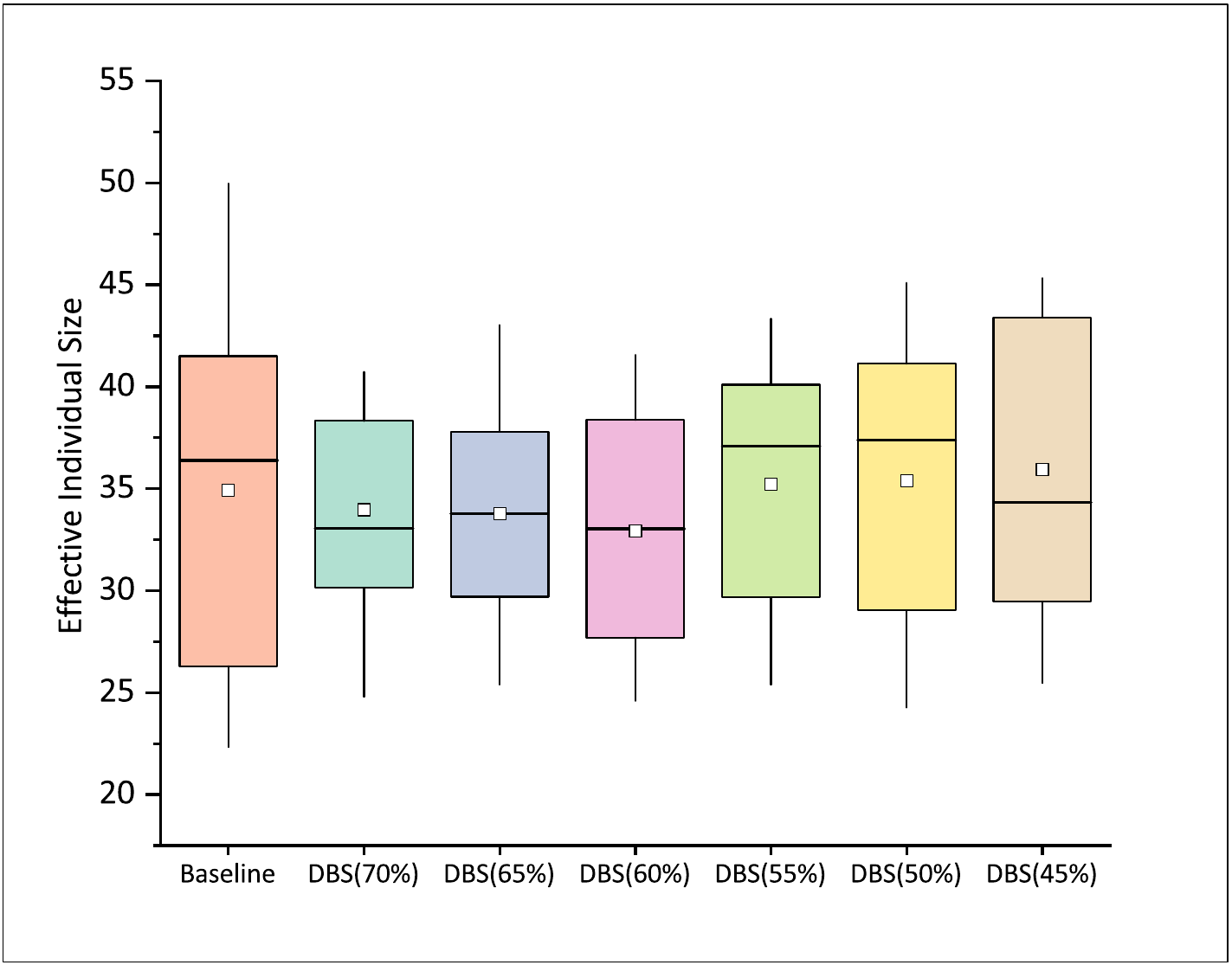}}
    \caption{Mean effective individual size of best solutions obtained on digital circuit data set.}
    \label{fig:CircuitEffSize}
\end{figure}

In a similar approach, we analyze the mean effective individual size of the best solutions obtained on circuit problems as shown in Figure \ref{fig:CircuitEffSize}. Since the GE run time in Figure \ref{fig:CircuitTime} is as expected, the variation in effective individual size has less of an influence if the training size and GE run-time are large, as mentioned previously.

Analyzing Figures \ref{fig:SynEff} to \ref{fig:CircuitEffSize}, the training data selected using the DBS approach is good at producing individuals having an effective size comparable to or smaller than the baseline approach in one or more instances of experiments, and thus have fewer chances of producing bloats.

\section{Conclusion}
\label{sec:conclusions}
We have introduced \gls{dbs} algorithm to obtain a reduced set of training data for \gls{ge} that generally produces similar or better results in terms of test case performance but with a substantially lower training cost. The algorithm offers flexibility to define a test case selection budget, which benefits the users in using the training data size of their choice. We tested the DBS approach on six different budgets. The test results of the solutions achieved using these training data were examined and compared to the corresponding baseline methods. Standard training techniques are used as a baseline in this work.

We evaluated the DBS algorithm on 24 benchmarks from the domain of \gls{sr} and digital circuits. The best solutions of a benchmark trained using different sets of training data in independent instances were tested against a common set of testing data. A graphical visualization of the obtained test scores is presented along with their discussion and statistical tests. In most of the cases, the solutions obtained using DBS training data were similar to or better than the models trained using baseline training data. The statistical test of the results indicates the training data selected using the proposed algorithm covers a wide region of the test case space and, hence, has less scope to compromise the quality of the solution. Time analysis is also provided, and it is observed that, in most instances, the training times for the models employing the DBS training data were comparable to or shorter than the baseline. Moreover, the effective individual size of the best solution achieved is analyzed. In most of the cases, the mean effective individual size of the solutions was found to be comparable to their corresponding baseline.

Nonetheless, our results suggest many novel directions for methodological research in test case selection to train models using \glspl{ea} like \gls{ge}. Moreover, the experiments using DBS in this paper serve as both an early case study and a platform to facilitate test case selection in other problem domains.

\section*{{Acknowledgements}}
This publication results from research conducted with the financial support of Science Foundation Ireland under Grant \# 16/IA/4605. The 3\textsuperscript{rd} author is also financed by the Coordenação de Aperfeiçoamento de Pessoal de Nível Superior - Brazil (CAPES), Finance Code 001, and the Fundação de Amparo à Pesquisa do Estado do Rio de Janeiro (FAPERJ).


\begin{thebibliography}{10}

\bibitem{AdvancesInGP:kinnear:1994}
Kenneth~E Kinnear, William~B Langdon, Lee Spector, Peter~J Angeline, and Una-May O'Reilly.
\newblock {\em Advances in genetic programming}, volume~3.
\newblock MIT press, 1994.

\bibitem{fitnessTimeComsum:yang:2003}
T~Yang, JM~McDonough, and JD~Jacob.
\newblock Parallelization of a genetic algorithm for curve fitting chaotic dynamical systems.
\newblock In {\em Parallel Computational Fluid Dynamics 2002}, pages 563--570. Elsevier, 2003.

\bibitem{CurseofBigData:duffy-deno}
Kevin Duffy-Deno.
\newblock {T}he {C}urse of {B}ig {D}ata --- bintel.io.
\newblock {https://www.bintel.io/blog/the-curse-of-big-data}, May 2021.
\newblock [Accessed 30-Oct-2022].

\bibitem{ali:automatedFeatureSel:2022}
Muhammad~Sarmad Ali, Meghana Kshirsagar, Enrique Naredo, and Conor Ryan.
\newblock Automated grammar-based feature selection in symbolic regression.
\newblock In {\em Proceedings of the Genetic and Evolutionary Computation Conference}, pages 902--910, 2022.

\bibitem{icsoft21}
Conor Ryan., Meghana Kshirsagar., Krishn~Kumar Gupt., Lukas Rosenbauer., and Joseph Sullivan.
\newblock {Hierarchical Clustering Driven Test Case Selection in Digital Circuits}.
\newblock In {\em Proceedings of the 16th International Conference on Software Technologies - ICSOFT}, pages 589--596. SciTePress, 2021.

\bibitem{gupt:predive:2022}
Krishn~Kumar Gupt, Meghana Kshirsagar, Lukas Rosenbauer, Joseph~P Sullivan, Douglas~Mota Dias, and Conor Ryan.
\newblock Predive: preserving diversity in test cases for evolving digital circuits using grammatical evolution.
\newblock In {\em Proceedings of the Genetic and Evolutionary Computation Conference Companion}, pages 719--722, 2022.

\bibitem{kshirsagar:igi:2022}
Meghana Kshirsagar, Krishn~Kumar Gupt, Gauri Vaidya, Conor Ryan, Joseph~P Sullivan, and Vivek Kshirsagar.
\newblock Insights into incorporating trustworthiness and ethics in ai systems with explainable ai.
\newblock {\em International Journal of Natural Computing Research (IJNCR)}, 11(1):1--23, 2022.

\bibitem{Bindra2021InsightsRevolution}
Prabhleen Bindra, Meghana Kshirsagar, Conor Ryan, Gauri Vaidya, Krishn~Kumar Gupt, and Vivek Kshirsagar.
\newblock {Insights into the Advancements of Artificial Intelligence and Machine Learning, the Present State of Art, and Future Prospects: Seven Decades of Digital Revolution}.
\newblock In Suresh~Chandra Satapathy, Vikrant Bhateja, Margarita~N Favorskaya, and T~Adilakshmi, editors, {\em Smart Computing Techniques and Applications}, pages 609--621, Singapore, 2021. Springer Singapore.

\bibitem{kubalik2020symbolic}
Ji{\v{r}}{\'\i} Kubal{\'\i}k, Erik Derner, and Robert Babu{\v{s}}ka.
\newblock Symbolic regression driven by training data and prior knowledge.
\newblock In {\em Proceedings of the 2020 Genetic and Evolutionary Computation Conference}, pages 958--966, 2020.

\bibitem{instanceSel:arnaiz:2016}
{\'A}lvar Arnaiz-Gonz{\'a}lez, Jos{\'e}-Francisco D{\'\i}ez-Pastor, Juan~J Rodr{\'\i}guez, and C{\'e}sar Garc{\'\i}a-Osorio.
\newblock Instance selection of linear complexity for big data.
\newblock {\em Knowledge-Based Systems}, 107:83--95, 2016.

\bibitem{instanceSelectionRegression:kordos:2012}
Miros{\l}aw Kordos and Marcin Blachnik.
\newblock Instance selection with neural networks for regression problems.
\newblock In {\em International Conference on Artificial Neural Networks}, pages 263--270. Springer, 2012.

\bibitem{instanceSelectionRegression:kordos:2018}
Miros{\l}aw Kordos and Krystian {\L}apa.
\newblock Multi-objective evolutionary instance selection for regression tasks.
\newblock {\em Entropy}, 20(10):746, 2018.

\bibitem{instanceSelectionRegression:Son:2006}
Seung-Hyun Son and Jae-Yearn Kim.
\newblock Data reduction for instance-based learning using entropy-based partitioning.
\newblock In {\em International Conference on Computational Science and Its Applications}, pages 590--599. Springer, 2006.

\bibitem{trainingSetSelection:kajdanowicz:2011}
Tomasz Kajdanowicz, Slawomir Plamowski, and Przemyslaw Kazienko.
\newblock Training set selection using entropy based distance.
\newblock In {\em 2011 IEEE Jordan Conference on Applied Electrical Engineering and Computing Technologies (AEECT)}, pages 1--5. IEEE, 2011.

\bibitem{clusterSelection:czarnowski:2010}
Ireneusz Czarnowski and Piotr J{\k{e}}drzejowicz.
\newblock Cluster integration for the cluster-based instance selection.
\newblock In {\em International Conference on Computational Collective Intelligence}, pages 353--362. Springer, 2010.

\bibitem{clusterSelection:czarnowski:2012}
Ireneusz Czarnowski.
\newblock Cluster-based instance selection for machine classification.
\newblock {\em Knowledge and Information Systems}, 30:113--133, 2012.

\bibitem{ClusterApproach:czarnowski:2003}
Ireneusz Czarnowski and Piotr J{\k{e}}drzejowicz.
\newblock An approach to instance reduction in supervised learning.
\newblock In {\em International Conference on Innovative Techniques and Applications of Artificial Intelligence}, pages 267--280. Springer, 2003.

\bibitem{ClusterReduction:wilson:2000}
D~Randall Wilson and Tony~R Martinez.
\newblock Reduction techniques for instance-based learning algorithms.
\newblock {\em Machine learning}, 38:257--286, 2000.

\bibitem{featureSR:chen:2017}
Qi~Chen, Mengjie Zhang, and Bing Xue.
\newblock Feature selection to improve generalization of genetic programming for high-dimensional symbolic regression.
\newblock {\em IEEE Transactions on Evolutionary Computation}, 21(5):792--806, 2017.

\bibitem{instanceSelectionRegression:arnaiz:2016}
{\'A}lvar Arnaiz-Gonz{\'a}lez, Jos{\'e}~F D{\'\i}ez-Pastor, Juan~J Rodr{\'\i}guez, and C{\'e}sar Garc{\'\i}a-Osorio.
\newblock Instance selection for regression: Adapting {DROP}.
\newblock {\em Neurocomputing}, 201:66--81, 2016.

\bibitem{book:IntrotoDigSys}
Mohammed Ferdjallah.
\newblock {\em {Introduction to digital systems: modeling, synthesis, and simulation using VHDL}}.
\newblock John Wiley {\&} Sons, 2011.

\bibitem{tan2014verilog}
Tze~Sin Tan and Bakhtiar~Affendi Rosdi.
\newblock Verilog hdl simulator technology: a survey.
\newblock {\em Journal of Electronic Testing}, 30(3):255--269, 2014.

\bibitem{gupt2021PrimeField}
Krishn~Kumar Gupt, Meghana Kshirsagar, Joseph~P. Sullivan, and Conor Ryan.
\newblock {Automatic Test Case Generation for Prime Field Elliptic Curve Cryptographic Circuits}.
\newblock In {\em 2021 IEEE 17th International Colloquium on Signal Processing {\&} Its Applications (CSPA)}, pages 121--126. IEEE, 3 2021.

\bibitem{gupt2021GF}
Krishn~Kumar Gupt, Meghana Kshirsagar, Joseph~P. Sullivan, and Conor Ryan.
\newblock {Automatic test case generation for vulnerability analysis of galois field arithmetic circuits}.
\newblock In {\em 2021 IEEE 5th International Conference on Cryptography, Security and Privacy, CSP 2021}, pages 32--37, 2021.

\bibitem{hsiung2018reconfigurable}
Pao-Ann Hsiung, Marco~D Santambrogio, and Chun-Hsian Huang.
\newblock {\em Reconfigurable system design and verification}.
\newblock CRC Press, 2018.

\bibitem{tcg:YeildRampChallanges}
Ann~Steffora Mutschler.
\newblock {Yield Ramp Challenges Increase}, 12 2014.

\bibitem{orso2014software}
Alessandro Orso and Gregg Rothermel.
\newblock {Software Testing: A Research Travelogue (2000–2014)}.
\newblock In {\em Future of Software Engineering Proceedings}, FOSE 2014, pages 117--132, New York, NY, USA, 2014. Association for Computing Machinery.

\bibitem{mrozek2012antirandom}
Ireneusz Mrozek and Vyacheslav Yarmolik.
\newblock Antirandom test vectors for bist in hardware/software systems.
\newblock {\em Fundamenta Informaticae}, 119(2):163--185, 2012.

\bibitem{PseudoExhaustive:Kuhn:2006}
D.~Richard Kuhn and Vadim Okun.
\newblock {Pseudo-exhaustive testing for software}.
\newblock {\em Proceedings of the 30th Annual IEEE/NASA Software Engineering Workshop, SEW-30}, pages 153--158, 2006.

\bibitem{thamarai2010heuristic}
SM~Thamarai, K~Kuppusamy, and T~Meyyappan.
\newblock Heuristic approach to optimize the number of test cases for simple circuits.
\newblock {\em arXiv preprint arXiv:1009.6186}, 2010.

\bibitem{muselli2000training}
Marco Muselli and Diego Liberati.
\newblock Training digital circuits with hamming clustering.
\newblock {\em IEEE Transactions on Circuits and Systems I: Fundamental Theory and Applications}, 47(4):513--527, 2000.

\bibitem{testing:faultDefType:timeCal:Bushnell2002EssentialsCircuits}
Michael~L. Bushnell and Vishwani~D. Agrawal.
\newblock {\em {Essentials of Electronic Testing for Digital, Memory and Mixed-Signal VLSI Circuits}}, volume~17 of {\em Frontiers in Electronic Testing}.
\newblock Springer US, Boston, MA, 2002.

\bibitem{OptimalRandomTest:Mrozek:2017}
Ireneusz Mrozek and Vyacheslav Yarmolik.
\newblock {Optimal Controlled Random Tests}.
\newblock {\em Lecture Notes in Computer Science (including subseries Lecture Notes in Artificial Intelligence and Lecture Notes in Bioinformatics)}, 10244 LNCS:27--38, 2017.

\bibitem{ge:ryan1998}
Conor Ryan, John~James Collins, and Michael~O Neill.
\newblock {Grammatical evolution: Evolving programs for an arbitrary language}.
\newblock In {\em European Conference on Genetic Programming}, pages 83--96, 1998.

\bibitem{handbookofGE}
Conor Ryan, Michael O’Neill, and J.~J. Collins.
\newblock {Handbook of grammatical evolution}.
\newblock {\em Handbook of Grammatical Evolution}, pages 1--497, 1 2018.

\bibitem{anjum2021seeding}
Muhammad~Sheraz Anjum and Conor Ryan.
\newblock Seeding grammars in grammatical evolution to improve search-based software testing.
\newblock {\em SN Computer Science}, 2(4):1--19, 2021.

\bibitem{murphy2021time}
Aidan Murphy, Ayman Youssef, Krishn~Kumar Gupt, Muhammad~Adil Raja, and Conor Ryan.
\newblock Time is on the side of grammatical evolution.
\newblock In {\em 2021 International Conference on Computer Communication and Informatics (ICCCI)}, pages 1--7, 2021.

\bibitem{youssef2021evolutionary}
Ayman Youssef, Krishn~Kumar Gupt, Muhammad~Adil Raja, Aidan Murphy, and Conor Ryan.
\newblock Evolutionary computing based analysis of diversity in grammatical evolution.
\newblock In {\em 2021 International Conference on Artificial Intelligence and Smart Systems (ICAIS)}, pages 1688--1693, 2021.

\bibitem{ibmClusteringBinary}
{IBM}.
\newblock Clustering binary data with {K}-{M}eans (should be avoided) --- ibm.com.
\newblock {https://www.ibm.com/support/pages/clustering-binary-data-k-means-should-be-avoided}, Apr 2020.
\newblock [Accessed 11-Oct-2022].

\bibitem{AgHierarchClustring:Contreras:2015}
Pedro Contreras and Fionn Murtagh.
\newblock Hierarchical clustering.
\newblock In {\em Handbook of cluster analysis}, pages 124--145. Chapman and Hall/CRC, 2015.

\bibitem{tamasauskas2012evaluation}
Darius Tamasauskas, Virgilijus Sakalauskas, and Dalia Kriksciuniene.
\newblock Evaluation framework of hierarchical clustering methods for binary data.
\newblock In {\em 2012 12th International Conference on Hybrid Intelligent Systems (HIS)}, pages 421--426. IEEE, 2012.

\bibitem{hierClsu:linkage:manning:2008}
Christopher~D Manning, Prabhakar Raghavan, and Hinrich Sch{\"u}tze.
\newblock {\em Introduction to information retrieval}.
\newblock Cambridge university press, 2008.

\bibitem{gelab:krishn:2022}
Krishn~Kumar Gupt, Muhammad~Adil Raja, Aidan Murphy, Ayman Youssef, and Conor Ryan.
\newblock {GELAB} – the cutting edge of grammatical evolution.
\newblock {\em IEEE Access}, 10:38694--38708, 2022.

\bibitem{ali:2021:autoge}
Muhammad Ali., Meghana Kshirsagar., Enrique Naredo., and Conor Ryan.
\newblock Autoge: A tool for estimation of grammatical evolution models.
\newblock In {\em Proceedings of the 13th International Conference on Agents and Artificial Intelligence - Volume 2: ICAART,}, pages 1274--1281. INSTICC, SciTePress, 2021.

\bibitem{Allan:grape:2022}
Allan de~Lima, Samuel Carvalho, Douglas~Mota Dias, Enrique Naredo, Joseph~P Sullivan, and Conor Ryan.
\newblock Grape: Grammatical algorithms in python for evolution.
\newblock {\em Signals}, 3(3):642--663, 2022.

\bibitem{mcdermott:2012:gpNeedBench}
James McDermott, David~R White, Sean Luke, Luca Manzoni, Mauro Castelli, Leonardo Vanneschi, Wojciech Jaskowski, Krzysztof Krawiec, Robin Harper, Kenneth De~Jong, et~al.
\newblock Genetic programming needs better benchmarks.
\newblock In {\em Proceedings of the 14th annual conference on Genetic and evolutionary computation}, pages 791--798, 2012.

\bibitem{Oliveira:SRreal-word:2018}
Luiz Otavio V.~B. Oliveira, Joao Francisco B.~S. Martins, Luis~F. Miranda, and Gisele~L. Pappa.
\newblock Analysing symbolic regression benchmarks under a meta-learning approach.
\newblock In {\em Proceedings of the Genetic and Evolutionary Computation Conference Companion}, GECCO '18, page 1342–1349, New York, NY, USA, 2018. Association for Computing Machinery.

\bibitem{sarmaad:SRreal-world:ecta21}
Muhammad Ali, Meghana Kshirsagar, Enrique Naredo, and Conor Ryan.
\newblock Towards automatic grammatical evolution for real-world symbolic regression.
\newblock In {\em Proceedings of the 13th International Joint Conference on Computational Intelligence - Volume 1: ECTA,}, pages 68--78. INSTICC, SciTePress, 2021.

\bibitem{Nicolau:libGE}
Miguel Nicolau and Darwin Slattery.
\newblock {\em libGE}, 2006.
\newblock for version 0.27alpha1, 14 September 2006.

\end{thebibliography}
\end{document}